\documentclass{article}
\usepackage{iclr2027_conference,times}

\usepackage{amsmath,amsfonts,bm}

\def\eqref#1{equation~\ref{#1}}

\def\1{\bm{1}}

\DeclareMathAlphabet{\mathsfit}{\encodingdefault}{\sfdefault}{m}{sl}
\SetMathAlphabet{\mathsfit}{bold}{\encodingdefault}{\sfdefault}{bx}{n}

\usepackage{amsmath}
\usepackage{amssymb}
\usepackage{booktabs}
\usepackage{tabularx}
\usepackage{longtable}
\usepackage{array}
\usepackage{mathtools}
\usepackage{multirow}
\usepackage{titletoc}
\usepackage{graphicx}
\usepackage{bm}
\usepackage{subcaption}
\usepackage{microtype}
\usepackage{xcolor}
\usepackage{url}
\usepackage{hyperref}
\hypersetup{hidelinks}
\usepackage[nameinlink,capitalize,noabbrev]{cleveref}
\usepackage{float}
\usepackage{placeins}
\title{RiboUnmix: Learning Shared Translational Dynamics from Biased and Noisy Ribo-seq Measurements}

\author{Gabriele Martino
  \thanks{Equal contribution.}
  \thanks{Correspondence: \texttt{\{gabriele.martino,denis.skibinski\}@univie.ac.at}.}
  \\
  Doctoral School Computer Science\\
  University of Vienna
  \And
  Denis Skibinski\footnotemark[1]
  \\
  Doctoral School Computer Science\\
  University of Vienna
  \And
  Ivo L. Hofacker
  \\
  ds:UniVie\\
  University of Vienna
  \And
  Sebastian Tschiatschek
  \\
  ds:UniVie\\
  University of Vienna
}

\iclrfinalcopy

\begin{document}
\maketitle
\lhead{Preprint. Under review at ICLR 2027}
\begin{abstract}
Ribosome profiling (Ribo-seq) is used to study translation dynamics by measuring how ribosomes are distributed along mRNA sequences, but the resulting occupancy profiles also reflect experiment-specific distortions and stochastic measurement variability. Models trained to predict these profiles from mRNA sequences can also learn these distortions, so accurate profile prediction alone does not establish recovery of the underlying biological behavior. We ask whether combining Ribo-seq datasets affected by different experimental conditions can enable us to reveal shared sequence-dependent patterns in the underlying ribosome distribution. We introduce \emph{RiboUnmix}, a probabilistic multi-dataset framework that jointly learns from multiple Ribo-seq datasets collected in different experiments. RiboUnmix represents each expected measured profile as a shared sequence dependent signal modulated by a dataset-specific multiplicative factor, while a negative-binomial observation model captures variability across individual Ribo-seq replicates. To evaluate recovery of the shared profile and dataset-specific effects, we construct a controlled synthetic Ribo-seq benchmark combining programmed translation kinetics, ribosome traffic, stochastic count sampling, and multiple sequence-dependent experimental distortions. The programmed kinetics and injected distortions provide known targets for evaluating the inferred shared profile and dataset-specific effects separately. Both inferred components show high correlation with their synthetic targets, showing RiboUnmix's ability to separate shared kinetic patterns from dataset-specific effects under controlled conditions. Across four organism-specific real-data benchmarks, RiboUnmix outperforms the evaluated sequence-to-profile baselines in predicting measured Ribo-seq profiles. The model independently trained on separate subsets of 114 HEK-derived datasets recovers concordant shared profiles for holdout transcripts, remarking reproducibility of the shared signal. Complementary experiments varying the number and composition of training datasets further show that the inferred shared representation remains substantially stable as the experimental evidence base changes. RiboUnmix turns variation across experiments into evidence for reproducible sequence-dependent patterns of ribosome occupancy, providing a foundation for biological hypothesis generation from diverse Ribo-seq datasets.
\\
\end{abstract}

\section{Introduction}
\label{sec:introduction}
Protein production depends on ribosome recruitment to mRNAs and progression along their coding sequences. These dynamics depend strongly on the coding sequence and can influence the abundance and folding of the resulting proteins
\citep{stein2019stop,tahmasebi2018translation}.
Ribosome profiling (Ribo-seq) measures ribosome occupancy along coding
sequences at approximately codon-level resolution
\citep{ingolia2014ribosome}, providing a rich target for learning how coding sequence shapes translation dynamics.
However, measured profiles combine translation together with experiment-specific distortions and stochastic sampling variability \citep{Lecanda2016dual, mok2023choros}. Sequence-to-profile models have shown that substantial structure in Ribo-seq
measurements can be predicted from the corresponding coding sequence
\citep{tunney2018accurate,hu2021riboexp,tian2021full,
zeng2025swamamba,kaynar2026seq2ribo}.
Architectures for such models have progressed from local-context predictors to full-sequence
and long-range models, but most are optimized to reproduce
measured profiles.
This creates an ambiguity for interpretation: learning reproducible sequence-dependent experimental effects can improve reconstruction without improving recovery of the real underlined biological translational profile \citep{tunney2018accurate}. Sampling variability limits agreement with observed profiles, whereas reproducible sequence-dependent experimental distortions can be learned by the model. Replicate averaging reduces sampling variability but does not separate biological and experimental effects.
We therefore ask whether variation across datasets can be used to learn a
shared sequence-derived profile that generalizes to unseen coding sequences while separating dataset-specific effects.

We introduce \textbf{RiboUnmix}, a probabilistic multi-dataset model that
treats Ribo-seq datasets as different observations of a shared
sequence-dependent structure.
A dataset-independent encoder predicts a positive, mean-one positional
profile $\mathbf L_t$ from coding sequence, while a separate
dataset-conditioned encoder predicts multiplicative positional corrections
and per-position count dispersion.
Experimental replicates are modeled individually with a negative-binomial
likelihood that accounts for overdispersed counts
\citep{love2014moderated,mok2023choros}.
A fixed-reference panel of datasets makes the allocation of positional structure
between the shared and dataset-specific components well defined across the
training panel.
By requiring the same $\mathbf L_t$ to explain multiple datasets while
allowing their corrections to vary, RiboUnmix uses experimental
heterogeneity to constrain a sequence-derived representation that can be
applied to previously unseen coding sequences.
We evaluate this decomposition in three complementary ways.
First, we construct a controlled synthetic benchmark with known programmed
sequence-dependent kinetics, simulated ribosome profiles, and injected
distortions.
This allows us to evaluate recovery of the shared and dataset-specific
components separately from reconstruction of the final noisy observations.
RiboUnmix recovers both components with strong agreement to their known
synthetic targets. Second, across four organism-specific prediction benchmarks, RiboUnmix
achieves the highest median transcript-level Pearson correlation between
predicted and measured Ribo-seq profiles among the evaluated pipelines.
Finally, models trained on source-disjoint panels drawn from 114 HEK-derived Ribo-seq datasets recover concordant shared profiles for held-out transcripts. Additional experiments assess how the shared representation changes with the number and composition of the used training datasets showing high stability.

Our main contributions are:
\begin{itemize}
    \item \textbf{A formal analysis of predictability from noisy and biased
    measurements.}
    We formulate sequence-to-profile learning as prediction of an experimental
    measurement process and provide formal guarantees characterizing two distinct
    effects on reconstruction: replicate variability limits attainable
    predictability, whereas reproducible dataset-specific bias becomes part of the
    optimal prediction target and can therefore improve reconstruction
    performance.

    \item \textbf{RiboUnmix: separating shared and dataset-specific structure.}
    We introduce a probabilistic multi-dataset model that learns a
    sequence-derived shared positional profile together with dataset-specific
    multiplicative corrections from heterogeneous Ribo-seq experiments.

    \item \textbf{A controlled benchmark for disentanglement.}
    We construct synthetic Ribo-seq observations with known programmed
    kinetics and known sequence-dependent distortions, enabling separate evaluation of both inferred components against ground-truth targets.

    \item \textbf{Real-data evidence for a stable shared representation.}
        Across 114 real Ribo-seq datasets, RiboUnmix learns concordant shared
        profiles from source-disjoint experimental panels, showing that the model can
        extract a reproducible sequence-dependent representation despite substantial
        variation in the training data.
        RiboUnmix also achieves the highest median transcript-level Pearson
        correlation between predicted and measured profiles among the evaluated
        models across four organisms.
        These results support the use of the learned shared profiles as a basis for
        future biological analysis and hypothesis generation.
\end{itemize}

\begin{figure*}[t]
    \centering
    \includegraphics[
        width=\textwidth,
        trim={0.75cm 0.5cm 0.5cm 0.5cm},
        clip
    ]{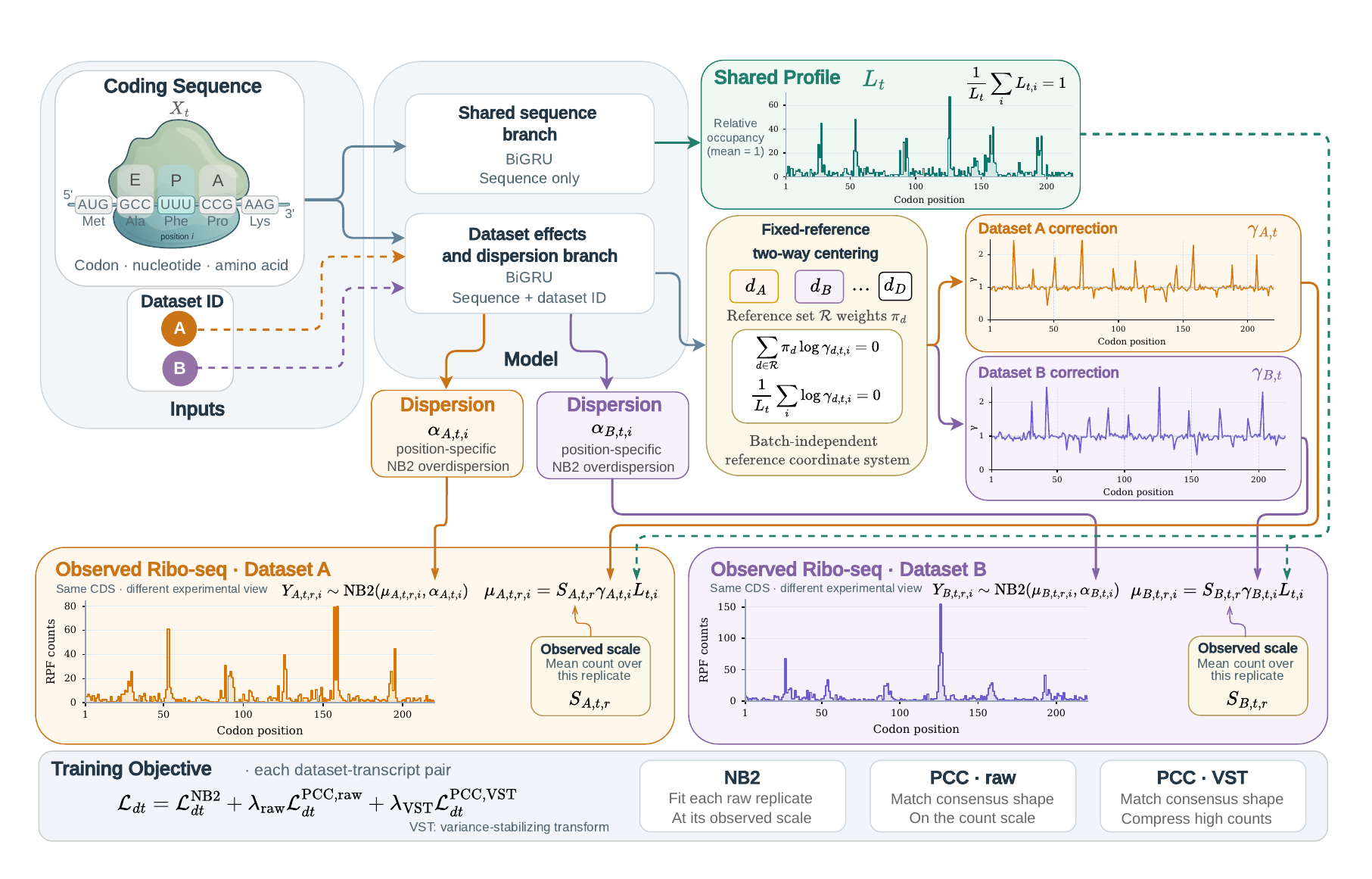}
    \caption{
    \textbf{Overview of RiboUnmix.}
    A dataset-independent sequence encoder produces the positive, mean-one
    shared profile $\mathbf L_t$.
    A separate dataset-conditioned encoder predicts positional corrections
    $\boldsymbol\gamma_{t,d}$ and NB2 dispersion parameters.
    Together with the observed replicate-specific scale
    $S_{t,d}^{(r)}$, these components define the expected count profile
    $\boldsymbol\mu_{t,d}^{(r)}
    =S_{t,d}^{(r)}
    (\mathbf L_t\odot\boldsymbol\gamma_{t,d})$.
    Raw replicates supervise the count likelihood, while their arithmetic
    consensus is used by auxiliary profile-agreement objectives.
    }
    \label{fig:model_architecture}
\end{figure*}

An implementation and scripts for reproducing the experiments are available at  \url{https://github.com/GabMartino/RiboUnmix}

\section{Related work}
\label{sec:related_work}

Probabilistic and deep generative models have been widely used to account for
technical variation in RNA-seq data. For example, scVI models batch effects
and measurement uncertainty in single-cell RNA-seq through a deep latent
variable model \citep{lopez2018deep}. Such approaches provide a useful
methodological precedent for separating biological and technical sources of
variation, but they operate on gene-level expression measurements rather than
sequence-conditioned, codon-resolution Ribo-seq profiles and are therefore
not directly comparable to our setting.

Within Ribo-seq, Choros separates biological and technical
sequence-associated effects using a structured negative-binomial regression
\citep{mok2023choros}.
Its technical component is defined through predefined bias-associated
features, whose fitted coefficients are used to derive corrected ribosome profiles. This requires the relevant bias patterns to be specified in advance:
technical effects not represented by the chosen features are not explicitly
recovered by the model.Riboformer instead uses deep learning to predict context-dependent Ribo-seq
profiles and can correct experimental distortions, but relies on a measured
reference profile as an additional input
\citep{shao2024riboformer}.
Riboclette conditions sequence-based prediction on biological perturbations
to model condition-specific changes
\citep{nallapareddy2025conditional}, rather than separating technical
measurement effects.These approaches address complementary aspects of Ribo-seq heterogeneity but
differ substantially in objective and supervision.
RiboUnmix does not require the form of the dataset-specific distortion to be
specified a priori and does not require a measured reference profile at
inference. Instead, it learns jointly across multiple datasets, using their
variation to separate a sequence-derived shared positional representation
from dataset-specific effects.

\section{Limits of Ribo-seq profile reconstruction}
\label{sec:problem}

Let $\mathbf X_t$ denote the coding sequence of transcript $t$ and
$\mathbf Y^{(r)}_{t,d}$ its measured Ribo-seq profile in replicate
$r$ of dataset $d$.
We consider each replicate as a realization of a dataset-specific
measurement process,
\begin{equation}
    \mathbf Y^{(r)}_{t,d}
    \sim p(\mathbf Y\mid\mathbf X_t,d),
    \qquad
    \mathbf m_d(\mathbf X_t)
    :=
    \mathbb E[
        \mathbf Y^{(r)}_{t,d}
        \mid \mathbf X_t,d
    ].
    \label{eq:experimental_observation_distribution}
\end{equation}
Here $\mathbf m_d(\mathbf X_t)$ is the expected measured profile
associated with dataset $d$, while
$\mathbf Y^{(r)}_{t,d}$ denotes an observed realization of this
measurement process. Conventional sequence-to-profile models are trained against experimentally
measured Ribo-seq profiles
\citep{hu2021riboexp,tian2021full,zeng2025swamamba,kaynar2026seq2ribo}.
We denote this supervision target by
$\mathbf Y^{\mathrm{target}}_{t,d}$, which may correspond to a single
replicate $\mathbf Y^{(r)}_{t,d}$ or to the replicate average $\overline{\mathbf Y}_{t,d}
=
\frac{1}{R_{t,d}}
\sum_{r=1}^{R_{t,d}}
\mathbf Y^{(r)}_{t,d}.
$ with $R_{t,d}$ the number of replicas present for the experiment $d$.
Such predictions approximate the outcome of experimental measurements but usually do not distinguish the effects of shared sequence-dependent structure and systematic differences between experimental settings. Conflating these effects, however, limits the insights that can be gained about the underlying translational processes. To formalize this distinction, we decompose the expected measurement profile
into a dataset-independent component
$\mathbf A_t=\mathbf A(\mathbf X_t)$ and a systematic
dataset-specific deviation $\mathbf B_{t,d}=\mathbf B(\mathbf X_t, d)$: 
\begin{equation}
    \begin{aligned}
        \mathbf m_d(\mathbf X_t)
        & =\mathbf A_t+\mathbf B_{t,d}, \quad
        \boldsymbol\varepsilon^{(r)}_{t,d}:=
            \mathbf Y^{(r)}_{t,d}
            -\mathbf m_d(\mathbf X_t),
        \quad
        \mathbb E[
        \boldsymbol\varepsilon^{(r)}_{t,d}
        \mid\mathbf X_t,d
        ]=\mathbf 0.
    \end{aligned}
    \label{eq:conceptual_observation}
\end{equation}
Here $\mathbf B_{t,d}$ may include predictable experimental artifacts as
well as condition-specific biology, whereas
$\boldsymbol\varepsilon^{(r)}_{t,d}$ represents realization-specific
variation around the expected measured profile.
Importantly, $\mathbf A_t$ is not identifiable from this decomposition alone. Conditional centering of
$\boldsymbol\varepsilon^{(r)}_{t,d}$ follows from the definition of the
expectation and requires no Gaussian assumption. However, this decomposition reveals two distinct ways in which reconstruction
performance can be misleading. First, if the objective is to predict the expected profile
$\mathbf m_d(\mathbf X_t)$, reconstruction performance against observed
replicates is constrained by their conditional variance: even exact prediction
of $\mathbf m_d(\mathbf X_t)$ leaves the realization-specific variability
unexplained. Second,
the systematic dataset-specific term
$\mathbf B_{t,d}$ is part of the optimal prediction target. A model can therefore improve its predictive performance of the measured profile by
learning a reproducible experimental artifact, without improving recovery of
the intended shared component $\mathbf A_t$. High reconstruction accuracy consequently
does not by itself demonstrate recovery of the intended shared signal, and
low accuracy may partly reflect measurement variance rather than a signal of poor model
capability. Such models may remain useful predictors, but without separating
dataset-specific effects their outputs have limited support as transferable
biological hypotheses. Appendix~\ref{app:predictability} formalizes both
effects.
RiboUnmix (Section~\ref{sec:model}) instead uses multiple heterogeneous
datasets to learn a positive sequence-derived profile $\mathbf L_t$ together
with dataset-specific effects $\boldsymbol\gamma_{t,d}$ in a multiplicative
count model.
It additionally incorporates a probabilistic observation model to represent
the count uncertainty inherent to Ribo-seq measurements
\citep{love2014moderated,mok2023choros}.

\section{Model}
\label{sec:model}

RiboUnmix models each Ribo-seq observation as a dataset-specific view
of a shared sequence-derived positional profile. For transcript $t$, let
$\mathbf X_t=(x_{t,1},\ldots,x_{t,n_t})$ denote its coding sequence where $x_{t,i} \in \mathbb{R}^k$ represents the codon, nucleotide, and amino-acid sequence
features in one-hot encoding, following \citet{tian2021full}, and
$\mathbf Y_{t,d}^{(r)}
=(Y_{t,d,1}^{(r)},\ldots,Y_{t,d,n_t}^{(r)})
\in\mathbb N_0^{n_t}$ the observed count profile in replicate $r$ of
dataset $d$. For any profile
$\mathbf Y_{t,d}$, we write
$\langle\mathbf Y_{t,d}\rangle
:=n_t^{-1}\sum_{i=1}^{n_t}Y_{t,d,i}$ for the mean along the transcript.

The main output of RiboUnmix is a positive, mean-one profile
\[
    \mathbf L_t\in\mathbb R_{>0}^{n_t},
    \qquad
    \langle\mathbf L_t\rangle=1,
\]
predicted from the coding sequence alone. Every dataset and replicate
containing transcript $t$ therefore uses the same $\mathbf L_t$.
For dataset $d$, the model predicts a positive positional distortion
$\boldsymbol\gamma_{t,d}$.
Together, $\mathbf L_t$ and $\boldsymbol\gamma_{t,d}$ define the
model-implied expected profile
$\mathbf L_t\odot\boldsymbol\gamma_{t,d}$ for transcript $t$ in
dataset $d$, where $\odot$ denotes position-wise multiplication.

RiboUnmix exploits the observed replicate-specific average count $S_{t,d}^{(r)}$ as scaling anchor, while $\mathbf L_t$ and $\boldsymbol\gamma_{t,d}$ shape the profile, i.e.,
\[
\boldsymbol\mu_{t,d}^{(r)}
=
S_{t,d}^{(r)}
\left(
\mathbf L_t\odot\boldsymbol\gamma_{t,d}
\right).
\]
Ribo-seq count levels depend on transcript abundance and sequencing depth, which are not determined by coding sequence alone \citep{Zhong2016RiboDiff}.
We therefore use the observed mean count of each replicate as its scale, allowing the model to focus on relative positional patterns. Under the centering convention defined in \cref{sec:gamma_gauge}, $\mathbf L_t$ represents the shared relative profile, while $\boldsymbol\gamma_{t,d}$ represents dataset-specific departures. Multiplicative corrections are motivated by sequence-dependent footprint recovery during library preparation \citep{mok2023choros}.

\subsection{Probabilistic observation model}
\label{sec:model_nb}

Following negative-binomial models for overdispersed sequencing counts
in RNA-seq and codon-resolution Ribo-seq
\citep{love2014moderated,mok2023choros}, we use an NB2 working
likelihood:
\begin{equation}
    \begin{gathered}
        Y_{t,d,i}^{(r)}
        \mid \mathbf X_t,d,S_{t,d}^{(r)}
        \sim
        \operatorname{NB2}\!\left(
            \mu_{t,d,i}^{(r)},\alpha_{t,d,i}
        \right),
        \quad
        \operatorname{Var}\!\left(
            Y_{t,d,i}^{(r)}
            \mid\mu_{t,d,i}^{(r)},\alpha_{t,d,i}
        \right)
        =
        \mu_{t,d,i}^{(r)}
        +
        \alpha_{t,d,i}
        \left(\mu_{t,d,i}^{(r)}\right)^2
        \\[0.75ex]
        \mu_{t,d,i}^{(r)}
        =
        \underbrace{S_{t,d}^{(r)}}_{
            \substack{\text{replicate}\\\text{scale anchor}}
        }
        \underbrace{L_{t,i}}_{
            \substack{\text{shared relative}\\\text{profile}}
        }
        \underbrace{\gamma_{t,d,i}}_{
            \substack{\text{dataset-specific}\\\text{correction}}
        }
    \end{gathered}
    \label{eq:nb_observation}
\end{equation}

Because the model targets the relative positional shape rather than the overall transcript count, as already shown we take the scale directly from the obeserved replicate $S_{t,d}^{(r)}
:=\langle\mathbf Y_{t,d}^{(r)}\rangle_t$. Hence, replicates of the same transcript--dataset pair share
$\mathbf L_t\odot\boldsymbol\gamma_{t,d}$ and the positional
dispersion $\boldsymbol\alpha_{t,d}$ also predicted by the model. Appendix~\ref{app:replicate_scale_rationale} discusses the interpretation
and limitations of this working likelihood.

\subsection{Sequence encoders for shared and dataset-specific structure}
\label{sec:profile_decomposition}
RiboUnmix is composed by two independent branches made up of bidirectional two-layered GRUs models \citep{cho2014learning}.
One branch predicts $\mathbf L_t$ from the coding sequence alone with a position-wise output head producing a positive profile:
\begin{equation}
    \widetilde{\mathbf L}_t=f_\theta(\mathbf X_t),\qquad
    L_{t,i}
    =
    \frac{\widetilde L_{t,i}}
         {\langle\widetilde{\mathbf L}_t\rangle}
    \label{eq:shared_profile_normalization}
\end{equation}
Here, the normalization ensures that $\langle\mathbf L_t\rangle=1$. It is important to highlight that neither dataset identity nor measured counts are supplied to this
branch, so $\mathbf L_t$ is identical for every Ribo-Seq observation of
transcript $t$.

The second branch predicts the positional correction
$\boldsymbol\gamma_{t,d}$ and dispersion
$\boldsymbol\alpha_{t,d}$.
Its inputs combine $\mathbf X_t$ with the dataset identity to produce a
dataset-conditioned sequence representation $\mathbf z_{t,d}$.
Two separate output heads then predict the raw log-correction $\mathbf{a_{t,d}^{\mathrm{raw}}}$ and
log-dispersion $\mathbf{\ell_{t,d}}$:
\begin{equation}
    \mathbf z_{t,d}
    =
    \psi_\phi(\mathbf X_t,d),
    \qquad
    \mathbf a_{t,d}^{\mathrm{raw}}
    =
    h_{\gamma}(\mathbf z_{t,d}),
    \qquad
    \boldsymbol\ell_{t,d}
    =
    h_{\alpha}(\mathbf z_{t,d}),
    \qquad
    \alpha_{t,d,i}
    =
    \exp(\ell_{t,d,i})
    \label{eq:dataset_branch}
\end{equation}

The dispersion is obtained directly as
$\alpha_{t,d,i}=\exp(\ell_{t,d,i})$.
The raw correction scores, instead, require an additional identification
step before defining $\boldsymbol\gamma_{t,d}$: because only the product
$\mathbf L_t\odot\boldsymbol\gamma_{t,d}$ enters the expected profile,
positional structure could otherwise be redistributed between the shared
and dataset-specific components. Figure~\ref{fig:model_architecture} summarizes the complete RiboUnmix architecture and the main operations linking the shared and dataset-specific branches to the observation model.

\subsection{Fixed-reference identification}
\label{sec:gamma_gauge}

We resolve this ambiguity by centering the correction scores across a fixed
collection of datasets. This makes positional shapes shared across the
datasets contribute to $\mathbf L_t$, while the centered corrections
represent dataset-specific departures from that reference, solving the identifiability issue.

To implement this, we define a \emph{reference panel} $\mathcal R$ as the set of datasets used
for this centering, with positive weights $\pi_d$ satisfying
$|\mathcal R|\geq2$ and
$\sum_{d\in\mathcal R}\pi_d=1$.
The panel and its weights are fixed before training.

For each transcript, the correction branch computes the raw score
$a_{t,e,i}^{\mathrm{raw}}$ for every reference dataset
$e\in\mathcal R$.
At each position, we subtract the reference-weighted mean across datasets and
then remove the positional mean of each resulting correction:
\begin{equation}
    c_{t,i}
    =
    \sum_{e\in\mathcal R}
    \pi_e a_{t,e,i}^{\mathrm{raw}},
    \qquad
    \widetilde g_{t,d,i}
    =
    a_{t,d,i}^{\mathrm{raw}}-c_{t,i},
    \qquad
    g_{t,d,i}
    =
    \widetilde g_{t,d,i}
    -
    \langle\widetilde{\mathbf g}_{t,d}\rangle.
    \label{eq:gamma_centering}
\end{equation}
The final positive correction is
$\gamma_{t,d,i}=\exp(g_{t,d,i})$.

This construction enforces
$\sum_{d\in\mathcal R}\pi_d g_{t,d,i}=0$ at every position to solve the identifiability issue, and
$\langle\mathbf g_{t,d}\rangle=0$ for every dataset removes a uniform log-amplitude shift from each correction, leaving
the replicate-specific scale $S_{t,d}^{(r)}$ to represent transcript-level
count magnitude as mentioned before.
Together, the two centering constraints yield a unique reference-defined
decomposition for profiles admitted by the model.
The formal identification results are given in Appendix~\ref{app:model_parameterization}.
The weights $\pi_d$ determine the reference used for this allocation.
Increasing $\pi_d$ gives dataset $d$ greater influence on which positional
structure is treated as shared rather than dataset-specific. In our experiments (Appendix~\ref{app:real_panel_reference_sensitivity}), we tested uniform, quality-informed, and reversed weighting to test how this choice would affect the final shared signal.

\begin{figure*}[!t]
\centering
\includegraphics[width=\textwidth]
{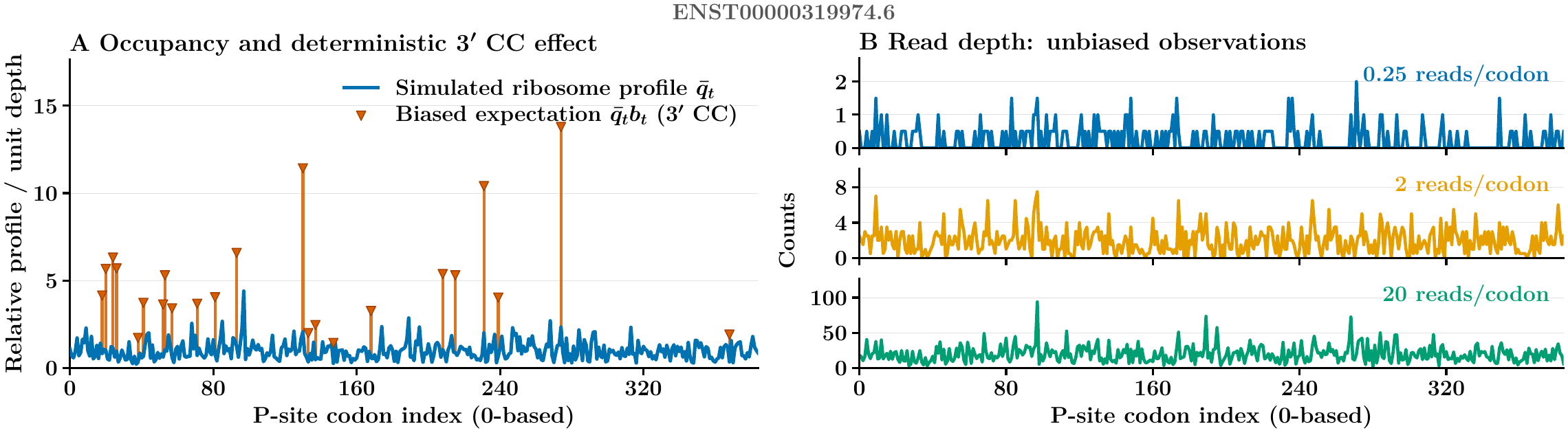}
\caption{
\textbf{Observation bias and sequencing depth.}
Both panels show transcript \texttt{ENST00000319974.6}.
\textbf{(A)} Mean of two simulated ribosome profiles,
$\overline{\mathbf q}_t$ (blue). At positions affected by the $3^\prime$-\texttt{CC} bias, stems and triangles show the expected profile $\bar q_{t,i}b_{t,f,i}$ per unit depth.
\textbf{(B)} Mean of two unbiased NB2 count replicates at nominal depths $C\in\{0.25,2,20\}$ reads per codon.}
\label{fig:synthetic_bias_depth_example}
\end{figure*}
\subsection{Composite training objective}
\label{sec:training_objective}
RiboUnmix is trained with a composite objective that combines
replicate-level count reconstruction with two measures of positional
agreement:
\begin{equation}
    \mathcal L_{t,d}
    =
    \mathcal L_{t,d}^{\mathrm{NB2}}
    +
    \lambda_{\mathrm{raw}}
    \mathcal L_{t,d}^{\mathrm{PCC,raw}}
    +
    \lambda_{\mathrm{VST}}
    \mathcal L_{t,d}^{\mathrm{PCC,VST}}.
    \label{eq:pair_objective}
\end{equation}

The NB2 term $\mathcal L_{t,d}^{\mathrm{NB2}}$ evaluates the raw replicate counts under the probabilistic
observation model.
The raw-PCC term $\mathcal L_{t,d}^{\mathrm{PCC,raw}}$ compares
$\boldsymbol\mu_{t,d}$ with
$\overline{\mathbf Y}_{t,d}$ on the original count scale using the Pearson Correlation Coefficient (PCC), while the
VST-PCC $\mathcal L_{t,d}^{\mathrm{PCC,VST}}$ term applies an NB2-motivated variance-stabilizing
transformation before computing the same positional agreement (Appendix~\ref{app:training_details}).
Together, the three terms encourage count-level fit and recovery of
profile shape across both high- and lower-count positions.

Within each transcript, the available transcript--dataset losses are
combined using positive reliability weights $w_{t,d}$ derived from read
support and positional coverage.
Their normalized weighted mean is then averaged across transcripts, so
each transcript contributes one outer loss term irrespective of the
number of available datasets. Appendix~\ref{sec:real_preprocessing} defines the reliability weights,
and Appendix~\ref{app:training_details} gives the complete loss
definitions and transcript-balanced reduction.
Experiment-specific settings are reported in
\cref{sec:synthetic_training_settings,sec:real_training_settings}.

\section{Controlled Validation on Synthetic Ribo-seq Observations}
\label{sec:synthetic}
\begin{figure}[!t]
    \centering
    \includegraphics[width=\textwidth]
        {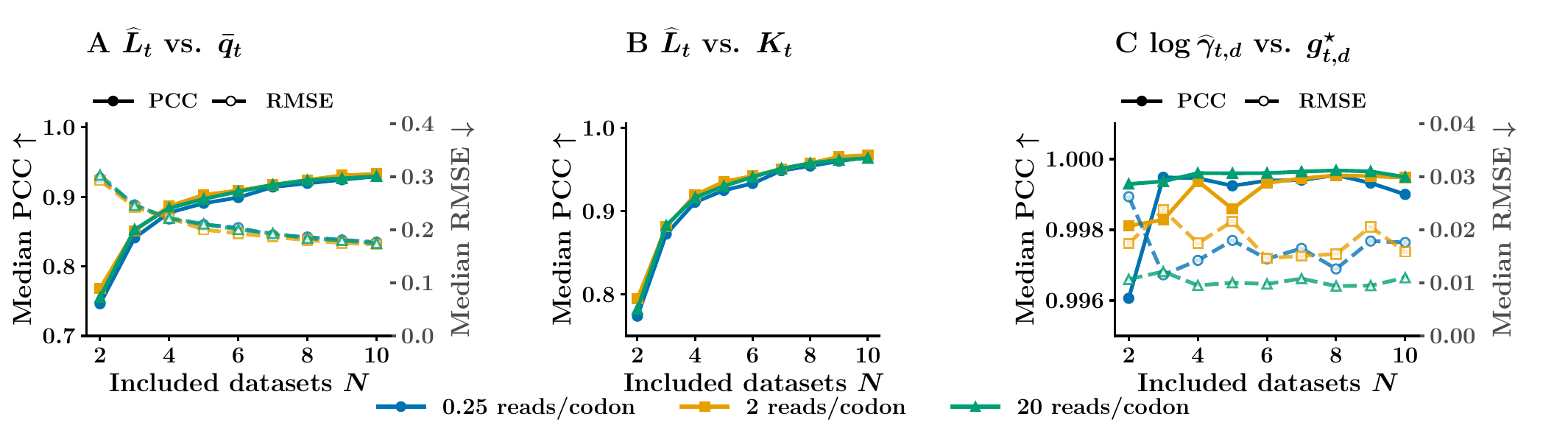}
    \caption{
    \textbf{Recovery of the simulated ribosome profile
    $\overline{\mathbf q}_t$ and centered observation effects
    $\mathbf g^\star_{t,d}$.}
    \textbf{(A)} Median transcript PCC (left axis) and RMSE
    (right axis) between $\widehat{\mathbf L}_t$ and
    $\overline{\mathbf q}_t$. Here RMSE is calculated after log transforming
    and position-centering each positive profile.
    \textbf{(B)} Median transcript
    $\operatorname{PCC}(\widehat{\mathbf L}_t,\mathbf K_t)$.
    \textbf{(C)} Mean dataset PCC (left axis) and joint
    dataset--position RMSE (right axis) between the fitted and exact centered
    log-corrections. Colors denote separate sequencing depths.}
    \label{fig:synthetic_overview}
\end{figure}
Experimental Ribo-seq measurements alone do not provide ground truth for the underlying ribosome profile or sequence-dependent measurement bias \citep{gerashchenko2017ribonuclease,mok2023choros}. We therefore construct a
controlled benchmark in which both quantities are known. For each of 19,290
human coding sequences \citep{morales2022mane}, we derive position-specific dwell time profiles $\mathbf K_t$ from codon identity and local sequence context.
These dwell times describe how long a ribosome is expected to remain at each position before accounting for interactions between ribosomes. We then simulate ribosomes moving along each transcript without overlapping, using an extended TASEP model
\citep{macdonald1968kinetics,shaw2003totally}. We normalize each simulated occupancy profile to mean one, obtaining $\mathbf q_t^{(r)}$, which describes the relative time ribosomes spend at each position.
We run two independent simulations for each transcript, producing
$\mathbf q_t^{(1)}$ and $\mathbf q_t^{(2)}$. Their average
$\overline{\mathbf q}_t=
(\mathbf q_t^{(1)}+\mathbf q_t^{(2)})/2$
is the simulated ribosome profile used to evaluate shared-profile recovery.
The two individual profiles are retained to preserve variation between
independent simulations. For each synthetic bias condition and depth, they are used to generate two
corresponding count replicates.
We introduce ten known sequence-dependent observation biases, which differ in the fragment-end sequences or nucleotide-composition features they target and in their multiplier values. For each bias condition $f$, the multiplier $b_{t,f,i}$ changes the expected count at every position $i$ in transcript $t$ whose corresponding footprint matches that condition, and equals one elsewhere. For bias condition $f$, simulation $r$, and nominal sequencing
depth $C$, counts are sampled as
\begin{equation}
Y_{t,f,i}^{(r)}
\sim
\operatorname{NB2}\left(
Cq_{t,i}^{(r)}b_{t,f,i},
\alpha_{\mathrm{sim}}=0.1
\right),
\qquad
C\in\{0.25,2,20\}.
\label{eq:main_synthetic_sampling}
\end{equation}
We generate each bias condition at all three fixed nominal sequencing depths, $C\in{0.25,2,20}$ reads per codon. Because each simulated profile $q_t^{(r)}$ is normalized to mean one, multiplying it by $C$ sets the mean expected count across positions to $C$ reads per codon before bias is applied. Because the biased profiles are not
renormalized, the multiplier can change both their positional shape and their
total expected count.
The complete generation process is therefore
$\mathbf K_t\rightarrow\mathbf q_t^{(r)}\rightarrow
\mathbf q_t^{(r)}\odot\mathbf b_{t,f}\rightarrow
\mathbf Y_{t,f}^{(r)}$. The first step converts sequence-derived dwell times into a simulated ribosome profile, the second applies one of the ten observation biases, and the third samples noisy counts. This construction provides separate
targets for evaluating recovery of the shared ribosome profile and the
dataset-specific biases. Figure~\ref{fig:synthetic_bias_depth_example}
illustrates how an observation bias changes the profile and how sequencing depth changes the sampled counts. Generation details are provided in Appendix \ref{app:synthetic_creation} and training and evaluation details in Appendix \ref{sec:synthetic_training_results}.

\subsection{Recovery of the shared profile and observation effects}
\label{sec:synthetic_results}

We compare the learned shared profile
$\widehat{\mathbf L}_t$ against the mean simulated ribosome profile $\overline{\mathbf q}_t$. We also compare the learned
dataset-specific corrections with the injected observation biases,
after applying the model's centering.
We train multiple subsets of datasets separately at each sequencing depth,
adding one dataset at a time to obtain $N=2,\ldots,10$
datasets. Each dataset contains two raw count replicates generated
from the same underlying pair of simulated trajectories.
All fits use uniform reference weights $\pi_d=1/N$. This tests how recovery changes as differently biased datasets are added.
We calculate the Pearson correlation between $\widehat{\mathbf L}_t$ and $\overline{\mathbf q}_t$ for each transcript and report the median across transcripts.
To obtain the comparison target $\mathbf g^\star_{t,d}$ we apply the centering operation in Equation \ref{eq:gamma_centering} to the known log-multipliers $\log\mathbf b_{t,d}$. This makes the injected and learned log-corrections directly comparable.
We calculate their positional Pearson correlation for each dataset, average these correlations within each transcript, and report the median across transcripts.
The centering formula and its interpretation are given in
\cref{sec:synthetic_gamma_recovery}.
Agreement between the learned shared profile and the simulated
ribosome profile increases along the cumulative series
(Figure~\ref{fig:synthetic_overview}A). Its alignment with $\mathbf K_t$ also rises (panel B). The learned log-corrections also closely match the centered injected biases, with high correlations and small errors (Figure~\ref{fig:synthetic_overview} C). 
Together, these results support recovery of the simulated ribosome profile and centered observation effects under the tested conditions.
The shared profile can still contain bias common across datasets. Along the tested order of dataset additions, this remaining bias becomes more uniform across positions as $N$ increases, improving agreement with the simulated ribosome profile
(\cref{sec:synthetic_identifiable_target,sec:synthetic_component_recovery}).
The improvement depends on which observation biases are added:
additional datasets need not improve recovery if they retain the
same systematic distortions. Increasing sequencing depth reduces
relative count-sampling noise but does not remove these distortions.
The same broad improvement is observed when all three depths are trained jointly for each bias condition (\cref{fig:synthetic_mixed_depth_reference}).
Recovery is evaluated on the same validation transcripts used for checkpoint selection; further evaluation limits, including dependence in count sampling across conditions, are reported in
Appendices \ref{sec:synthetic_cross_condition_diagnostics} and \ref{sec:synthetic_training_settings}.

\section{Multi-dataset experiments on real Ribo-seq observations}
\label{sec:real_multi_dataset}

\begin{figure}[!t]
    \centering
    \includegraphics[width=\textwidth]
        {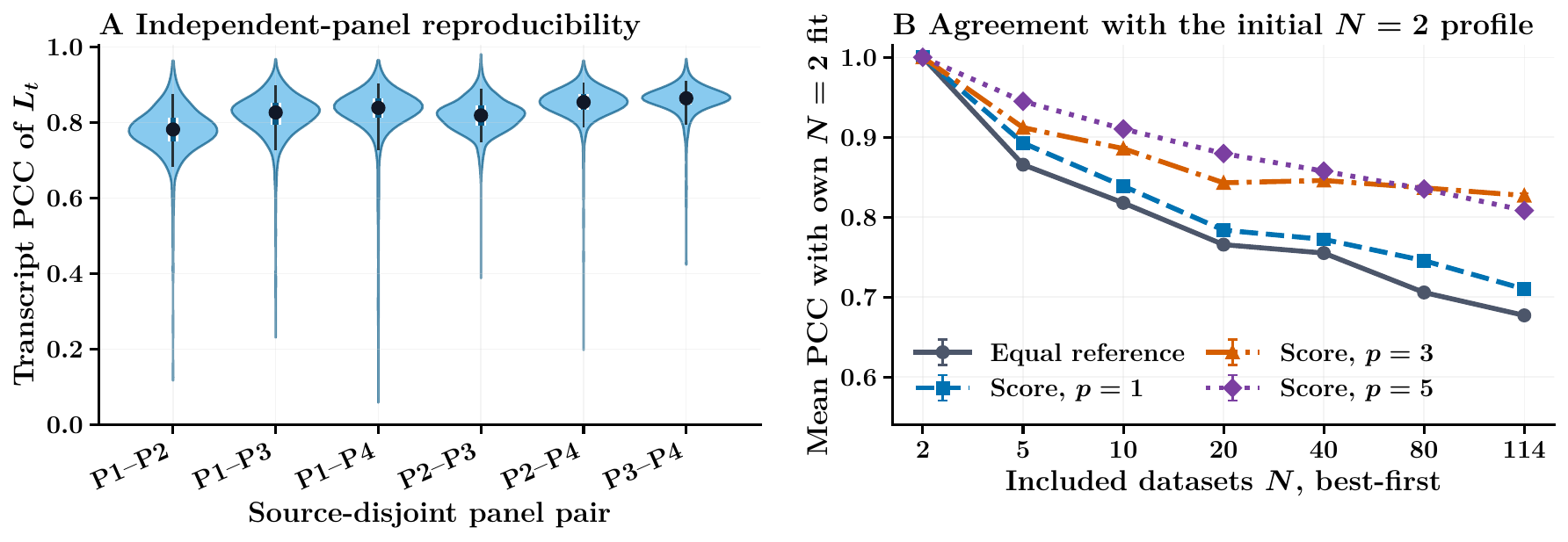}
    \caption{
    \textbf{Shared-profile reproducibility across heterogeneous real-data
    collections.}
    \textbf{(A)} Transcript-level shared-profile PCC for all six balanced,
    source-disjoint panel pairs.  Dots mark medians; bars show IQR and
    5th--95th percentiles.
    \textbf{(B)} Mean PCC with each reference policy's own $N=2$ shared
    profile along the nested best-first path. Bars are 95\%
    transcript-bootstrap intervals.}
    \label{fig:real_data_stability_main}
\end{figure}

Real data provide no true shared profile or dataset correction.  We therefore
test reproducibility across independent collections and stability along a
quality-ordered path.  Across 114 HEK-derived controls, raw replicates enter
the NB2 likelihood separately and reliability references use training
transcripts only (\cref{app:hek293_ribo,sec:real_preprocessing}).

\paragraph{Independent panels.}
Four balanced panels of 29, 29, 28, and 28 datasets keep source families
intact.  Uniform-reference models share 1,593 held-out transcripts, with no
dataset or source family shared between any panel pair.
\paragraph{Cumulative quality ordering.}
The prespecified quality ordering defines nested best-first collections at
$N\in\{2,5,10,20,40,80,114\}$ on one fixed transcript test split.  We fit uniform
and three score-weighted gamma references. Each policy's own $N=2$ fit is
its fixed anchor, so the comparison asks how much the initial shared profile
changes as datasets are added. Opposite quality paths and reference geometry
are deferred to the appendix. Source-disjoint panels concentrate at high agreement
(Figure~\ref{fig:real_data_stability_main}A), supporting a reproducible signal. Along the nested path, agreement with
the initial profile decreases as the collection expands (panel B) adding worse quality datasets as expected, with more retention the more quality scores are accentuated.
Opposite quality paths, exact estimates, and audits are in \cref{sec:real_stability_experiments}.


\section{Benchmark performance}
\label{sec:4organism_main}

\begin{table}[!t]
\centering
\begingroup
\small
\setlength{\tabcolsep}{4pt}
\renewcommand{\arraystretch}{1.08}
\setlength{\aboverulesep}{0.3ex}
\setlength{\belowrulesep}{0.3ex}
\newcommand{\mci}[2]{#1\,{\(\pm\)}\,#2}
\newcommand{\bmci}[2]{\textbf{#1\,{\(\pm\)}\,#2}}
\newcommand{\bestmci}[2]{\underline{\textbf{#1\,{\(\pm\)}\,#2}}}

\begin{tabular*}{\textwidth}{@{\extracolsep{\fill}}lcccc@{}}
\toprule
Model
& \textit{C. elegans}
& \textit{E. coli}
& Human (HEK293T)
& \textit{S. cerevisiae} \\
\midrule
iXnos
& \mci{0.271}{0.016}
& \mci{0.350}{0.028}
& \mci{0.472}{0.014}
& \mci{0.444}{0.019} \\
RiboExp
& \mci{0.253}{0.018}
& \mci{0.374}{0.024}
& \mci{0.418}{0.013}
& \mci{0.456}{0.020} \\
Riboformer
& \mci{0.257}{0.021}
& \mci{0.466}{0.032}
& \mci{0.465}{0.017}
& \mci{0.435}{0.016} \\
Seq2Ribo
& \mci{0.201}{0.017}
& \mci{0.236}{0.028}
& \mci{0.222}{0.012}
& \mci{0.347}{0.014} \\
RiboMIMO
& \mci{0.325}{0.024}
& \mci{0.516}{0.028}
& \mci{0.461}{0.016}
& \mci{0.502}{0.019} \\
\textbf{RiboUnmix}
& \bmci{0.335}{0.026}
& \bmci{0.556}{0.027}
& \bmci{0.539}{0.019}
& \bmci{0.517}{0.024} \\
\bottomrule
\end{tabular*}
\endgroup

\caption{\textbf{Pearson correlation benchmark across four organism.}
Entries report the median transcript-level Pearson correlation
\(\pm\) half the width of their 95\% percentile bootstrap intervals.  Metrics use positive-count positions. All baseline rows use the full RiboUnmix training cohort and the same transcript reliability weights. \cref{tab:four_model_nonzero_test_performance} reports native and matched training settings, together with Spearman correlation and normalized RMSE. The largest estimate
per organism is shown in bold.}
\label{tab:four_model_nonzero_pearson}
\end{table}

We evaluate RiboUnmix’s single-dataset prediction performance on four organism-specific benchmarks, training separate models for
\textit{C. elegans}, \textit{E. coli}, human \textit{HEK293T}, and
\textit{S. cerevisiae} \citep{Stein2022wurmyeast,Burkhardt2017ecoli,Iwasaki2016human}.
We compare with iXnos, RiboExp, sequence-only Riboformer, RiboMIMO, and Seq2Ribo on common held-out transcripts, excluding the first and last five codons of each coding sequence.
\citep{tunney2018accurate,hu2021riboexp,shao2024riboformer,tian2021full,kaynar2026seq2ribo}.
\cref{tab:four_model_nonzero_pearson} reports baselines trained on the full
RiboUnmix training split, target construction, and transcript-reliability
weights; matched unweighted controls and native settings are reported in the
Appendix \ref{app:training_set_checkpoint}.
We report the median transcript-level Pearson correlation between predicted and observed profiles at positions with positive observed counts $Y>0$. Additional metrics and evaluation details are provided in Appendix \ref{app:four_organism_benchmark}. RiboUnmix achieves the highest median transcript-level Pearson correlation estimate in all four benchmarks (\cref{tab:four_model_nonzero_pearson}).
These results support competitive prediction of measured profiles, complementing the multi-dataset analyses in \cref{sec:synthetic} and \ref{sec:real_multi_dataset}.  Matching training cohorts and reliability weights reduces differences between pipelines, although losses and checkpoint-selection criteria remain model-specific (Appendix \ref{app:4organism_results}).

\section{Conclusion}
\label{sec:conclusion}
RiboUnmix uses variation across Ribo-seq experiments to learn shared sequence-dependent ribosome profiles. Its fixed-reference decomposition jointly models shared profiles, dataset-specific effects, and count variability without requiring ground-truth shared profiles for training.

In controlled synthetic experiments, RiboUnmix recovers the simulated
ribosome profile and the centered injected observation biases, with
shared-profile recovery improving along the tested order of dataset additions.
Models trained on separate groups of the 114 HEK-derived datasets, with no studies shared between groups, produce similar shared profiles for held-out transcripts.
Across four organism-specific benchmarks, RiboUnmix also achieves the highest median transcript-level Pearson correlation estimates between predicted and measured profiles at positive-count positions among the evaluated pipelines.

The shared profile depends on the chosen reference and can retain biases common across datasets. Independent biological validation is needed to establish how faithfully it reflects ribosome occupancy in living cells. RiboUnmix turns variation across experiments into information for learning reproducible sequence-dependent structure, providing a basis for testable hypotheses about translational regulation.

\section*{Acknowledgments}
This work has been funded in parts by the Vienna Science and Technology Fund (WWTF) [10.47379/LS23053].

\section*{Use of Generative AI}
Large language models were used to assist with debugging and with the
implementation of analysis and visualization scripts. They were also used to
review mathematical exposition for clarity and consistency and to assist with
grammar correction during manuscript preparation.

All AI-assisted code and analyses were inspected and tested by the authors, and all reported results were computed from the saved experimental outputs. The authors reviewed the final text, mathematical statements, figures, and scientific claims and take responsibility for the complete content of this work.

\bibliography{iclr2027_conference}
\bibliographystyle{iclr2027_conference}

\clearpage
\appendix
\section*{ \LARGE Appendix Table of Contents}

\startcontents[appendixtoc]
\printcontents[appendixtoc]{}{1}{} 

\newpage

\section{Reconstruction limits under experimental variability}
\label{app:predictability}
We use the scalar counterpart of the observation framework in
Sec.~\ref{sec:problem}. Let $X$ denote a coding sequence together with
an evaluated codon position, and let $Y^{(r)}$ denote the measured count
at that position in replicate $r$. Thus, $Y^{(r)}$ corresponds to
$Y_{t,d,i}^{(r)}$ in the model notation.

Fix a dataset $d$, a replicate index $r$, and a distribution of
transcript--position examples. Assume
$\mathbb E[(Y^{(r)})^2\mid d]<\infty$ and
$\operatorname{Var}(Y^{(r)}\mid d)>0$, and consider predictors $f(X)$
satisfying $\mathbb E[f(X)^2\mid d]<\infty$. Define
$m_d(X):=\mathbb E[Y^{(r)}\mid X,d]$ and
$\varepsilon_d^{(r)}:=Y^{(r)}-m_d(X)$.
Conditioning on $X,d$ holds the sequence and position fixed;
conditioning only on $d$ also averages over the selected distribution
of transcript--position examples.

\subsection{Stochastic variability limits attainable reconstruction}
\label{app:noise_predictability}

For the squared-error population risk
$\mathcal R_d(f):=\mathbb E[(Y^{(r)}-f(X))^2\mid d]$,
substituting $Y^{(r)}=m_d(X)+\varepsilon_d^{(r)}$ gives
\begin{equation}
    \mathcal R_d(f)
    =
    \underbrace{
        \mathbb E[
            \operatorname{Var}(Y^{(r)}\mid X,d)
            \mid d
        ]
    }_{\text{irreducible conditional variability}}
    +
    \underbrace{
        \mathbb E[
            (m_d(X)-f(X))^2
            \mid d
        ]
    }_{\text{conditional-mean prediction error}}.
    \label{eq:appendix_risk_decomposition}
\end{equation}
The cross term vanishes because
$\mathbb E[\varepsilon_d^{(r)}\mid X,d]=0$; no Gaussian assumption
is needed. Consequently, $f_d^*(X)=m_d(X)$ minimizes the risk uniquely
up to sets of probability zero under the chosen input distribution.
The remaining error is irreducible for predictors using only $X$.
It may include unobserved biological variation as well as sampling
and experimental noise. Increasing model capacity cannot eliminate
this term without additional predictive information.

Define population $R^2$ relative to the best constant predictor as
$R_d^2(f):=1-\mathcal R_d(f)/
\operatorname{Var}(Y^{(r)}\mid d)$.
Combining the minimum risk with the law of total variance gives
\begin{equation}
    R_{\max,d}^2
    =
    \frac{
        \operatorname{Var}(m_d(X)\mid d)
    }{
        \operatorname{Var}(m_d(X)\mid d)
        +
        \mathbb E[
            \operatorname{Var}(Y^{(r)}\mid X,d)
            \mid d
        ]
    }.
    \label{eq:appendix_rmax}
\end{equation}
This maximum ranges over all square-integrable predictors using $X$;
a restricted model class need not attain it.
For a fixed input distribution and conditional mean, greater expected
conditional variance lowers the ceiling whenever
$\operatorname{Var}(m_d(X)\mid d)>0$.

The same variance ratio bounds population Pearson correlation.
Conditional centering gives
$\operatorname{Cov}(f(X),Y^{(r)}\mid d)
=
\operatorname{Cov}(f(X),m_d(X)\mid d)$.
Applying the Cauchy--Schwarz inequality therefore yields, for every
predictor with $\operatorname{Var}(f(X)\mid d)>0$,
\begin{equation}
    \operatorname{Corr}(f(X),Y^{(r)}\mid d)^2
    \leq
    \frac{
        \operatorname{Var}(m_d(X)\mid d)
    }{
        \operatorname{Var}(Y^{(r)}\mid d)
    }
    =
    R_{\max,d}^2.
    \label{eq:appendix_pearson_ceiling}
\end{equation}
When $\operatorname{Var}(m_d(X)\mid d)>0$, the predictor $f=m_d$
attains the maximum positive correlation,
$\sqrt{R_{\max,d}^2}$.
For a general predictor, however, $R_d^2(f)$ need not equal its
squared Pearson correlation with the observations.

To illustrate the effect of replication, now consider $R$ replicates
of the same transcript--position example. Assume that, conditional
on $X,d$, they are independent and identically distributed, with
conditional mean $m_d(X)$. Their average
$\overline Y_R:=R^{-1}\sum_{r=1}^{R}Y^{(r)}$
then satisfies
$\mathbb E[\overline Y_R\mid X,d]=m_d(X)$ and
$\operatorname{Var}(\overline Y_R\mid X,d)
=\operatorname{Var}(Y^{(r)}\mid X,d)/R$.
The attainable population $R^2$ becomes
\begin{equation}
    R_{\max,d}^2(\overline Y_R)
    =
    \frac{
        \operatorname{Var}(m_d(X)\mid d)
    }{
        \operatorname{Var}(m_d(X)\mid d)
        +
        R^{-1}
        \mathbb E[
            \operatorname{Var}(Y^{(r)}\mid X,d)
            \mid d
        ]
    }.
    \label{eq:appendix_replica_predictability}
\end{equation}
Averaging can therefore improve attainable reconstruction while
preserving all systematic effects contained in $m_d(X)$.
The $1/R$ variance reduction requires the stated assumptions.
Unequal replicate scales can change the conditional means and
variances, while conditional dependence introduces covariance terms.
This calculation illustrates the effect of replication; it does not
assume that the raw replicates used by RiboUnmix have identical
conditional count distributions.

\subsection{Predictable experimental effects are rewarded by reconstruction}
\label{app:bias_reward}

Let $A(X)$ be a dataset-independent reference satisfying
$\mathbb E[A(X)^2\mid d]<\infty$, expressed on the same count scale
as $m_d(X)$, and define $B_d(X):=m_d(X)-A(X)$.
These are the scalar counterparts of $\mathbf A_t$ and
$\mathbf B_{t,d}$ in Equation~\ref{eq:conceptual_observation}.
The reference $A(X)$ is introduced for this analytical comparison
and is distinct from the model's mean-one output $\mathbf L_t$.

For the predictor $f_A(X):=A(X)$,
Equation~\ref{eq:appendix_risk_decomposition} gives
\begin{equation}
    \mathcal R_d(f_A)-\mathcal R_d(f_d^*)
    =
    \mathbb E[B_d(X)^2\mid d].
    \label{eq:appendix_bias_reward}
\end{equation}
If $B_d(X)$ is nonzero with positive probability under the fixed
dataset distribution, predicting the full conditional mean achieves
strictly lower expected squared error than predicting $A(X)$ alone.
The objective therefore rewards sequence-predictable deviations from
the reference, including experimental artifacts and condition-specific
biology. This identity describes an optimization incentive; it does
not identify $A$ or $B_d$ from the observations.

The argument also covers multiplicative observation effects.
For $A(X)>0$ and a positive multiplier $b_d(X)$ satisfying
$m_d(X)=A(X)b_d(X)$, the deviation is
$B_d(X)=A(X)(b_d(X)-1)$.
The excess risk is consequently
$\mathbb E[A(X)^2(b_d(X)-1)^2\mid d]$.
Thus, the additive decomposition does not require an additive
physical mechanism.
Equation~\ref{eq:appendix_bias_reward} compares predictions of the
same experimental target; it does not imply that stronger artifacts
necessarily increase Pearson correlation or population $R^2$ when
comparing different datasets.

Together, these results explain why reconstruction performance alone
cannot establish recovery of a biological profile.
Conditional variability limits agreement with measured counts,
whereas sequence-predictable experimental effects contribute to the
optimal reconstruction target. Replicate averaging reduces variability
under the stated assumptions but preserves those systematic effects.

The results concern population quantities under the specified input
distribution. They do not directly give numerical ceilings for
finite-sample correlations or summaries of transcript-level
correlations. They also do not directly apply to count predictors
that additionally use the observed scale anchors $S_{t,d}^{(r)}$.
The excess-risk identity is specific to squared error, rather than
the NB2 likelihood or composite objective used by RiboUnmix.
Its relevance is that fitting measured profiles rewards predictable
structure in those measurements, whose biological interpretation
requires separate evidence.

\section{Identification and scale constraints}
\label{app:model_parameterization}
Fix a transcript $t$ and use the same $n_t$ positions across datasets.
Write $\langle\cdot\rangle_t$ for the positional mean.
Let $h_{t,d,i}>0$ denote a supplied relative expected profile to be
represented as $h_{t,d,i}=L_{t,i}\gamma_{t,d,i}$.
For the model, this profile is
$\mu_{t,d,i}^{(r)}/S_{t,d}^{(r)}$ and is shared across replicates.
The reference panel $\mathcal R$ has positive weights $\pi_d$
satisfying $\sum_{d\in\mathcal R}\pi_d=1$.

\subsection{Unique fixed-reference decomposition}
\label{app:gamma_projection}

Under the cross-dataset constraint alone,
$\sum_{d\in\mathcal R}\pi_d\log\gamma_{t,d,i}=0$,
each supplied collection of positive profiles has the unique
factorization
\begin{equation}
    L_{t,i}
    =
    \prod_{d\in\mathcal R}h_{t,d,i}^{\pi_d},
    \qquad
    \gamma_{t,d,i}
    =
    \frac{h_{t,d,i}}{L_{t,i}}.
    \label{eq:app_unique_decomposition}
\end{equation}
Indeed, taking the weighted logarithmic mean across datasets at
the same transcript and position gives
\begin{equation}
    \sum_{d\in\mathcal R}\pi_d\log h_{t,d,i}
    =
    \log L_{t,i}
    +
    \sum_{d\in\mathcal R}\pi_d\log\gamma_{t,d,i}
    =
    \log L_{t,i}.
    \label{eq:app_identification_proof}
\end{equation}
This determines $L_{t,i}$ and then $\gamma_{t,d,i}$ uniquely;
substitution verifies the factorization and constraint.

The result identifies the factors for specified relative profiles.
It does not establish unique estimation from incomplete or noisy
observations, or identify the shared factor as biological.
The model's additional normalization constraints restrict which
profiles can be represented, as discussed below.

\subsection{Two-way centering and fixed-reference evaluation}
\label{app:gamma_centering_compatibility}

Equation~\ref{eq:gamma_centering} first centers the raw log-corrections
across datasets at each position of transcript $t$, giving
$\sum_{d\in\mathcal R}\pi_d\widetilde g_{t,d,i}=0$.
Subtracting the positional mean within each transcript--dataset pair
then gives $\langle\mathbf g_{t,d}\rangle_t=0$ while preserving the
first constraint:
\begin{equation}
    \sum_{d\in\mathcal R}\pi_d g_{t,d,i}
    =
    \sum_{d\in\mathcal R}\pi_d\widetilde g_{t,d,i}
    -
    \left\langle
        \sum_{d\in\mathcal R}\pi_d
        \widetilde{\mathbf g}_{t,d}
    \right\rangle_t
    =
    0.
    \label{eq:app_simultaneous_constraints}
\end{equation}
Repeating the centering leaves the result unchanged.
A positional term shared across datasets or a position-independent
offset for one dataset is removed by these operations, so the raw
scores themselves remain non-unique.

For each transcript, the correction branch evaluates all reference
dataset identities needed to calculate the center, including those
without an observed profile for that transcript.
These evaluations remain connected to differentiation during training
but introduce no additional observation losses.
At deterministic evaluation, the correction therefore uses the same
reference regardless of which datasets are requested together.

The algebra requires the requested corrections and reference center
to use the same realized raw scores.
Independent dropout draws for those evaluations can break exact
cross-dataset centering during a stochastic forward pass, although
positional centering still holds.

\subsection{Representable profiles and predicted total counts}
\label{app:gamma_existence}
\label{app:replicate_scale_rationale}

The factors in Equation~\ref{eq:app_unique_decomposition} satisfy
the model's additional constraints
$\langle\mathbf L_t\rangle_t=1$ and
$\langle\log\boldsymbol\gamma_{t,d}\rangle_t=0$
if and only if
\begin{equation}
    \langle\mathbf L_t\rangle_t=1,
    \qquad
    \langle\log\mathbf h_{t,d}\rangle_t
    =
    \langle\log\mathbf L_t\rangle_t
    \quad\text{for every }d\in\mathcal R.
    \label{eq:app_profile_compatibility}
\end{equation}
The second condition follows directly from
$\log\boldsymbol\gamma_{t,d}
=\log\mathbf h_{t,d}-\log\mathbf L_t$.
These conditions concern the factorization itself; a finite neural
network may impose further restrictions.

Centering log-corrections does not ensure that
$\mathbf L_t\odot\boldsymbol\gamma_{t,d}$ has arithmetic mean one.
When $S_{t,d}^{(r)}$ equals the positive observed mean and no
numerical floor is active, the predicted-to-observed total-count
ratio is
\begin{equation}
    m_{t,d}
    :=
    \left\langle
        \mathbf L_t\odot\boldsymbol\gamma_{t,d}
    \right\rangle_t
    =
    \frac{
        \sum_i\mu_{t,d,i}^{(r)}
    }{
        \sum_iY_{t,d,i}^{(r)}
    }.
    \label{eq:app_mass_ratio}
\end{equation}
This ratio is shared across replicates of the same
transcript--dataset pair.

The weighted arithmetic--geometric mean inequality gives
$\sum_{d\in\mathcal R}\pi_d\gamma_{t,d,i}
\geq\prod_{d\in\mathcal R}\gamma_{t,d,i}^{\pi_d}=1$.
Multiplying by $L_{t,i}$ and averaging over positions yields
\begin{equation}
    \sum_{d\in\mathcal R}\pi_d m_{t,d}
    \geq
    \langle\mathbf L_t\rangle_t
    =
    1.
    \label{eq:app_reference_mass_inequality}
\end{equation}
Because all $L_{t,i}$ and reference weights are positive, equality
requires $\gamma_{t,d,i}=1$ for every reference dataset and position.
Nontrivial corrections therefore produce a reference-weighted
average count ratio above one, although individual ratios may be
below one. Consequently, the decoder cannot exactly represent
distinct positive relative profiles that all have mean one.

The ratio $m_{t,d}$ measures total-count calibration.
It is not an independent estimate of transcript abundance or
translational activity, because the observed total already enters
the prediction through $S_{t,d}^{(r)}$.

\subsection{Biological interpretation of the shared profile}
\label{app:shared_profile_interpretation}

Suppose the expected observations for transcript $t$ are proportional
to an underlying positive profile $\mathbf q_t$ multiplied by
dataset-specific measurement effects $\mathbf b_{t,d}>0$.
For this transcript, define the weighted geometric mean effect
across reference datasets as
\begin{equation}
    \mathbf G_t
    :=
    \prod_{d\in\mathcal R}\mathbf b_{t,d}^{\pi_d},
    \qquad
    \mathbf L_t^{\mathrm{ref}}
    :=
    \frac{
        \mathbf q_t\odot\mathbf G_t
    }{
        \langle\mathbf q_t\odot\mathbf G_t\rangle_t
    }.
    \label{eq:app_reference_shared_shape}
\end{equation}
Products and powers are element-wise.
The profile $\mathbf L_t^{\mathrm{ref}}$ is the normalized geometric
mean of the expected dataset shapes for the same transcript.
It provides a comparison under the chosen reference convention;
the compatibility restrictions above prevent identifying it
unconditionally with the fitted $\mathbf L_t$.

The reference-defined shape equals the normalized underlying profile
precisely when the combined measurement effect is constant across
positions:
\begin{equation}
    \mathbf L_t^{\mathrm{ref}}
    =
    \frac{\mathbf q_t}{\langle\mathbf q_t\rangle_t}
    \quad\Longleftrightarrow\quad
    G_{t,i}=C_t
    \quad\text{for every }i,
    \qquad C_t>0.
    \label{eq:app_biological_recovery}
\end{equation}
If $G_{t,i}=C_t$, the constant cancels under normalization.
Conversely, equality of the normalized profiles and positivity of
$q_{t,i}$ imply
$G_{t,i}
=\langle\mathbf q_t\odot\mathbf G_t\rangle_t/
\langle\mathbf q_t\rangle_t$,
which is independent of position.

A positional measurement effect shared across datasets can therefore
remain in the shared profile. More generally, replacing
$\mathbf q_t$ by $\mathbf q_t\odot\mathbf c_t$ and every
$\mathbf b_{t,d}$ by $\mathbf b_{t,d}\oslash\mathbf c_t$,
for any positive $\mathbf c_t$, leaves their products unchanged.
The reference convention fixes a decomposition, but independent
biological evidence is needed to determine how faithfully the learned
shared profile represents ribosome occupancy.

\section{Training objective and data handling}
\label{app:training_details}

The synthetic and HEK multi-dataset experiments use the objective
defined in the main text. This section specifies how its terms are
calculated and combined. Experiment-specific settings are given in
\cref{sec:synthetic_training_settings,sec:real_training_settings}.

\subsection{Replicate handling}
\label{app:replica_handling}

For transcript $t$ in dataset $d$, the NB2 term evaluates each of the
$R_{t,d}$ raw count replicates separately, using its own observed
scale $S_{t,d}^{(r)}$. The two correlation terms instead compare
the predicted profile with the arithmetic mean
$\overline{\mathbf Y}_{t,d}
=R_{t,d}^{-1}\sum_{r=1}^{R_{t,d}}\mathbf Y_{t,d}^{(r)}$.
For these terms, the prediction is
$\boldsymbol\mu_{t,d}
=\langle\overline{\mathbf Y}_{t,d}\rangle_t
(\mathbf L_t\odot\boldsymbol\gamma_{t,d})$.
The replicate mean enters each correlation term once.
Synthetic datasets contain two replicates, and all positional
calculations use valid codons only.

\subsection{NB2 count reconstruction}
\label{app:length_tempering}

For each replicate, we average the NB2 negative log-likelihood
over valid positions. We then multiply this average by
$\operatorname{clip}[(n_t/1000)^{0.25},0.5,2]$ and average across
replicates to obtain $\mathcal L_{t,d}^{\mathrm{NB2}}$.
Because the positional loss is already averaged, this bounded factor
gives longer transcripts moderately greater weight in the NB2 term
without weighting them in direct proportion to their length.

\subsection{Profile agreement}
\label{app:vst_loss}

The raw correlation term measures positional agreement on the count
scale. The second correlation term applies an NB2-motivated
transformation that compresses high counts:
\begin{equation}
    T_\alpha(x)
    =
    \frac{
        2\operatorname{asinh}\!\left(\sqrt{\alpha x+\varepsilon}\right)
    }{
        \sqrt{\alpha+\varepsilon}
    },
    \label{eq:app_nb_vst}
\end{equation}
where $\varepsilon>0$ is a small numerical constant.
The transformation acts separately at each position using the
predicted dispersion $\alpha_{t,d,i}$.
The two losses are
\begin{equation}
\begin{aligned}
    \mathcal L_{t,d}^{\mathrm{PCC,raw}}
    &=
    1-\operatorname{PCC}\!\left(
        \boldsymbol\mu_{t,d},
        \overline{\mathbf Y}_{t,d}
    \right),\\
    \mathcal L_{t,d}^{\mathrm{PCC,VST}}
    &=
    1-\operatorname{PCC}\!\left(
        T_{\boldsymbol\alpha_{t,d}}(\boldsymbol\mu_{t,d}),
        T_{\boldsymbol\alpha_{t,d}}(\overline{\mathbf Y}_{t,d})
    \right).
\end{aligned}
\label{eq:app_shape_penalties}
\end{equation}
Both experiment series use
\begin{equation}
    \mathcal L_{t,d}
    =
    \mathcal L_{t,d}^{\mathrm{NB2}}
    +
    0.5\,\mathcal L_{t,d}^{\mathrm{PCC,raw}}
    +
    0.5\,\mathcal L_{t,d}^{\mathrm{PCC,VST}}.
    \label{eq:app_pair_loss}
\end{equation}

Gradients through the dispersion values are stopped inside $T$,
so the correlation terms do not train the dispersion head.
The input representation supplied to that head is also detached,
preventing gradients from its dispersion-prediction path from
updating the context encoder. The NB2 term trains the dispersion
head and the predicted mean.

\subsection{Transcript weighting and batching}
\label{app:transcript_reduction}
\label{app:reliability_weights}

Within each transcript, we combine the available dataset losses
$\mathcal L_{t,d}$ using the normalized reliability weights
$w_{t,d}/\sum_d w_{t,d}$, where the sum includes its retained
datasets. We then average these combined losses equally across
transcripts in the batch. Each transcript therefore contributes
one loss, irrespective of how many datasets contain it.
The reliability weights $w_{t,d}$ determine each observation's
contribution to training; the reference weights $\pi_d$ determine
the centering of dataset-specific corrections.
Eligibility and reliability are specified in
\cref{sec:real_preprocessing}, with synthetic input processing in
\cref{sec:synthetic_training_preprocessing}.

The multi-dataset sampler retains transcripts with usable
observations in at least two selected datasets and keeps all
retained observations of each transcript together.
When a logical batch is divided into execution microbatches,
transcript groups remain intact. Each microbatch contributes
in proportion to its number of transcripts, preserving the
equal average across the complete logical batch.
Gradient clipping and the optimizer update occur after gradient
accumulation. The single-dataset reconstruction control also
permits transcripts observed in only one dataset.

\section{Synthetic data generation}
\label{app:synthetic_creation}
Synthetic ground truth refers throughout to quantities defined by the
simulator, rather than to experimentally established biological truth.
The generation process links four quantities:
\begin{equation}
    \mathbf K_t
    \longrightarrow
    \mathbf q_t^{(r)}
    \longrightarrow
    \boldsymbol\mu_{t,f}^{(r)}
    \longrightarrow
    \mathbf Y_{t,f}^{(r)}.
    \label{eq:synthetic_profile_hierarchy}
\end{equation}
We describe synthetic data generation here, examine the generated profiles in Appendix \ref{sec:synthetic_data_diagnostics} and evaluate component recovery in
Appendix \ref{sec:synthetic_training_results}.
Here, $\mathbf K_t$ is the programmed kinetic profile and
$\mathbf q_t^{(r)}$ is the mean-one simulated ribosome profile from simulation $r$.
$\boldsymbol\mu_{t,f}^{(r)}$ and $\mathbf Y_{t,f}^{(r)}$ are the expected and sampled count profiles under bias condition $f$. Here ``occupancy'' means
the fraction of recording time for which a ribosome P-site is at a codon.
Normalizing this occupancy profile to mean one gives the simulated ribosome profile $\mathbf q_t^{(r)}$.
For $R$ trajectories the arithmetic mean is
$\overline{\mathbf q}_t=R^{-1}\sum_{r=1}^{R}\mathbf q_t^{(r)}$.
We use two simulations and two corresponding count replicates per dataset to retain variation between simulations and from count sampling. The same saved trajectories are reused across depths
and conditions, so those datasets do not supply additional independent trajectories.

\subsection{Programmed kinetics}
\label{app:synthetic_generation}

\begin{figure}[htbp]
    \centering
    \includegraphics[width=\linewidth]
        {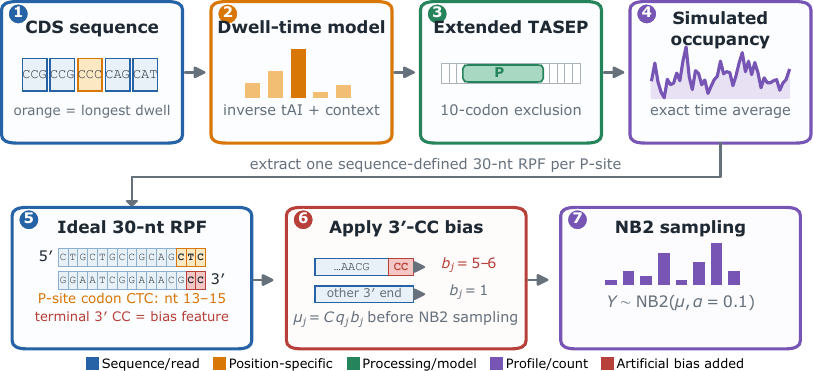}
    \caption{
    \textbf{Synthetic-data generation.}
    Sequence-dependent dwell times define the programmed kinetic
    profile $\mathbf K_t$ and parameterize an open extended TASEP with
    10-codon ribosome footprints. Two independent simulations produce the simulated ribosome profiles $\mathbf q_t^{(1)}$ and $\mathbf q_t^{(2)}$. Sequence features of idealized 30-nt ribosome-protected fragments determine the positions affected by each observation bias, illustrated here by a $3^\prime$-\texttt{CC} bias. These multipliers modify the
    expected counts before NB2 sampling.
    }
    \label{fig:synthetic_generation_pipeline}
\end{figure}

We generated synthetic Ribo-seq data for 19,290 human coding sequences from MANE Select \citep{morales2022mane} using the workflow summarized in
\cref{fig:synthetic_generation_pipeline}.

For each of four HEK-derived tRNA-abundance datasets
(GSE152621 \citep{Behrens2021tRNA2},
GSE66550 \citep{Clark2016tRNA1},
GSE141436 \citep{Pinkard2020tRNA3}, and
GSE95683 \citep{Gogakos2017tRNA4}),
we converted the measured abundances into codon-specific tRNA adaptation
index (tAI) weights following \citet{Reis2004tAI}. For each sense codon, we averaged the four weights and rescaled the resulting consensus weights so that their maximum was one.

Following previous uses of tRNA adaptation to parameterize translation
rates \citep{Tuller2011dwelltime,Gritsenko2015synth3}, the consensus
weight $w_c$ defines the baseline dwell time
\begin{equation}
    \tau_c^{\mathrm{base}}
    =
    \max\left(
        \frac{\lambda}{w_c},
        0.05~\mathrm{s}
    \right).
\end{equation}
We chose $\lambda$ so that the mean dwell time across the
61 sense codons is $0.25~\mathrm{s}$, setting a baseline elongation scale of $4~\mathrm{codons/s}$. This baseline is close to the mean elongation rate of $4.08~\mathrm{codons/s}$ measured in HEK293 T-REx cells by \citet{tomuro2024calibrated}.

The baseline dwell times are further modified by local sequence-context rules, including slow-codon combinations
\citep{Gamble2016slowcomb}, proline motifs
\citep{Gutierrez2013proline}, charged and bulky peptide contexts
\citep{Charneski2013positive}, and A/U-rich sequence contexts. 
After mapping the programmed dwell times to P-site positions, we denote the final dwell time at position $i$ of transcript $t$ by $\tau_{t,i}$.

The programmed kinetic profile is
\begin{equation}
    K_{t,i}
    =
    \frac{
        \tau_{t,i}
    }{
        \ell_t^{-1}
        \sum_{j=1}^{\ell_t}\tau_{t,j}
    },
    \label{eq:synthetic_K_target}
\end{equation}
where $\ell_t$ is the number of modeled sense-codon positions. Thus,
$\mathbf K_t$ has positional mean one, and larger $K_{t,i}$ denotes a
longer programmed dwell at position $i$. We define $\mathbf K_t$ before simulating ribosome traffic or applying observation biases and count sampling.

\subsection{Stochastic simulated ribosome profiles}
\label{sec:synthetic_simulated_profiles}
The programmed dwell times define the unblocked elongation rates
$\tau_{t,i}^{-1}$ of an open extended TASEP
\citep{rogers2017tasep},which we simulated using the direct Gillespie algorithm
\citep{gillespie1977exact}. Ribosomes occupy 10 codons
\citep{ingolia2014ribosome} and advance stochastically by one codon at the programmed rate whenever exclusion permits. At the final sense-codon P-site, ribosomes leave the transcript at rate $\bigl(0.30~\mathrm{s}\bigr)^{-1}$.

We obtained transcript-specific initiation intervals $I_t$ from
\citet{tomuro2024calibrated} for 3,998 exact transcript-version
matches. The remaining 15,292 transcripts received the matched-set
median of $20.61~\mathrm{s/event}$. For each simulation, we discarded an initial period equal to five transcript-specific collision-free translation times and then recorded for $T_t = \max(600~\mathrm{s},\,40I_t)$. The collision-free translation time is the sum of the programmed elongation dwell times and the terminal departure time, excluding initiation waiting and ribosome blocking.

For each transcript, we ran two independent simulations
$r\in\{1,2\}$. They share the same programmed rates, initiation
interval, termination rate, footprint length, and recording duration,
but differ in their realized sequence of stochastic initiation,
elongation, and termination events. Variation between
$\mathbf q_t^{(1)}$ and $\mathbf q_t^{(2)}$ therefore arises within the
traffic simulation, before observation effects or count-sampling noise
are introduced.

Let $Z_{t,i}^{(r)}(s)$ indicate whether a ribosome P-site occupies
position $i$ at time $s$ in trajectory $r$. The recorded time-averaged
occupancy is
\begin{equation}
    O_{t,i}^{(r)}
    =
    \frac{1}{T_t}
    \int_0^{T_t}
        Z_{t,i}^{(r)}(s)\,\mathrm ds.
\end{equation}
We normalize each simulated occupancy profile to mean one:
\begin{equation}
    q_{t,i}^{(r)}
    =
    \frac{
        O_{t,i}^{(r)}
    }{
        \ell_t^{-1}
        \sum_{j=1}^{\ell_t}O_{t,j}^{(r)}
    }.
    \label{eq:synthetic_latent_profile}
\end{equation}
Thus, $\mathbf q_t^{(r)}$ describes the relative time spent by
ribosome P-sites at different positions during one finite stochastic
trajectory. Two trajectories generated from the same programmed
kinetics generally produce different profiles because they experience
different event times, temporary queues, and numbers of ribosome visits
to each position.

The programmed kinetic and simulated ribosome profiles coincide in an ideal
collision-free steady state. If a common ribosome flux $J_t$ passes
every position without blocking, then
\begin{equation}
    O_{t,i}^{\mathrm{cf}}
    =
    J_t\tau_{t,i}.
\end{equation}
After normalization, the common flux cancels:
\begin{equation}
    q_{t,i}^{\mathrm{cf}}
    =
    \frac{
        J_t\tau_{t,i}
    }{
        \ell_t^{-1}
        \sum_{j=1}^{\ell_t}J_t\tau_{t,j}
    }
    =
    K_{t,i}.
    \label{eq:synthetic_K_q_relation}
\end{equation}

The finite TASEP simulations can differ from this ideal limit.
Ribosome blocking can create queues, while initiation and termination rates can alter the occupancy profile through their effects on ribosome traffic. Finite recording
introduces additional trajectory-specific variation. Consequently,
$\mathbf q_t^{(r)}$ can differ systematically from
$\mathbf K_t$, in addition to varying between simulations.

The mean simulated ribosome profile is
\begin{equation}
    \overline{\mathbf q}_t
    =
    \frac{
        \mathbf q_t^{(1)}
        +
        \mathbf q_t^{(2)}
    }{2}.
    \label{eq:synthetic_occupancy_consensus}
\end{equation}
Averaging reduces trajectory-specific variation but does not remove
systematic reshaping caused by ribosome traffic. Accordingly,
$\overline{\mathbf q}_t$ is the mean simulated ribosome profile over two
finite trajectories, not an exact stationary profile. Comparison with
$\overline{\mathbf q}_t$ assesses recovery of the simulated ribosome profile, whereas comparison with $\mathbf K_t$ assesses agreement with the programmed kinetics.

\subsection{Deterministic observation effects and count sampling}
\label{sec:synthetic_observation_generation}
Sequence-dependent biases in footprint recovery can arise during nuclease
digestion and library preparation, including preferences associated
with fragment ends and ligation
\citep{gerashchenko2017ribonuclease,mok2023choros,Lecanda2016dual}.
We construct idealized 30-nt ribosome-protected fragments in which the
P-site codon occupies nucleotides 13--15 in one-based fragment
coordinates. For fragments extending beyond the coding sequence, we use annotated UTR sequence. Observation biases are applied only where a complete fragment can be constructed.

The ten bias conditions target $5^\prime$- or
$3^\prime$-terminal AA, CC, GG, and UU dinucleotides, or on AU-rich and
GC-rich fragment composition. For bias condition $f$, we apply the multiplier $b_{t,f,i}$ specified in
\cref{tab:synthetic_biases} at matching positions with complete fragments and set $b_{t,f,i}=1$ elsewhere.

The same multipliers are used for both simulations and all sequencing depths.

\begin{table}[!htbp]
    \centering
    \small
    \caption{Artificial sequence-dependent bias conditions. Multiplier strengths are synthetic stress-test choices, not estimates of experimentally measured Ribo-seq bias. AU-rich fragments contain more than 70\% A and U nucleotides. GC-rich fragments contain more than 70\% G and C nucleotides.}
    \label{tab:synthetic_biases}
    \begin{tabular}{lrrr}
        \toprule
        Bias feature & Multiplier & P-sites & Transcripts \\
        \midrule
        3$'$ CC & 5.0--6.0$\times$ & 943,724 & 19,266 \\
        3$'$ GG & 4.0--4.4$\times$ & 577,402 & 19,240 \\
        3$'$ AA & 2.8--3.2$\times$ & 785,775 & 19,160 \\
        3$'$ UU & 5.8--6.2$\times$ & 646,563 & 19,068 \\
        5$'$ CC & 5.2--5.6$\times$ & 706,527 & 19,242 \\
        5$'$ GG & 4.2--4.6$\times$ & 738,405 & 19,252 \\
        5$'$ AA & 3.0--3.4$\times$ & 1,048,194 & 19,256 \\
        5$'$ UU & 5.6--6.0$\times$ & 641,947 & 19,188 \\
        AU-rich & 7.0--7.4$\times$ & 263,552 & 10,119 \\
        GC-rich & 5.0--5.4$\times$ & 801,495 & 14,095 \\
        \bottomrule
    \end{tabular}
\end{table}

\paragraph{Expected profiles and count sampling.}
For nominal sequencing depth $C\in\{0.25,2,20\}$ , defined as the expected number of reads per codon before bias is applied, the expected count at position $i$ is
\begin{equation}
    \mu_{t,f,i}^{(r)}
    =
    C
    q_{t,i}^{(r)}
    b_{t,f,i}.
    \label{eq:synthetic_expected_count_profile}
\end{equation}
The expected count profile therefore combines one simulated ribosome
profile with a deterministic sequence-dependent
observation bias. Counts are then sampled as
\begin{equation}
    Y_{t,f,i}^{(r)}
    \mid
    \mathbf q_t^{(r)},\mathbf b_{t,f},C
    \sim
    \operatorname{NB2}\!\left(
        \mu_{t,f,i}^{(r)},
        \alpha_{\mathrm{sim}}=0.1
    \right),
    \label{eq:synthetic_biased_nb2}
\end{equation}
where
\begin{equation}
    \operatorname{Var}\!\left(
        Y_{t,f,i}^{(r)}
        \mid
        \mu_{t,f,i}^{(r)}
    \right)
    =
    \mu_{t,f,i}^{(r)}
    +
    0.1
    \bigl(
        \mu_{t,f,i}^{(r)}
    \bigr)^2.
\end{equation}

Setting $b_{t,f,i}=1$ gives the unbiased condition. Observation biases
are applied before count sampling, without renormalizing the biased expected profiles. Since
$\mathbf q_t^{(r)}$ has positional mean one, the unbiased expected total
is $\ell_t C$, whereas under condition $f$,
\begin{equation}
    \sum_{i=1}^{\ell_t}
        \mu_{t,f,i}^{(r)}
    =
    \ell_t C
    \left\langle
        \mathbf q_t^{(r)}
        \odot
        \mathbf b_{t,f}
    \right\rangle_t.
    \label{eq:synthetic_expected_total}
\end{equation}
Equal nominal depth therefore does not imply equal expected total counts
across observation conditions.

\subsection{Reading the generation layers and the depth intervention}
\label{sec:synthetic_generation_examples}

\begin{figure*}[p]
    \centering
    \includegraphics[width=\textwidth,height=0.74\textheight,keepaspectratio]
        {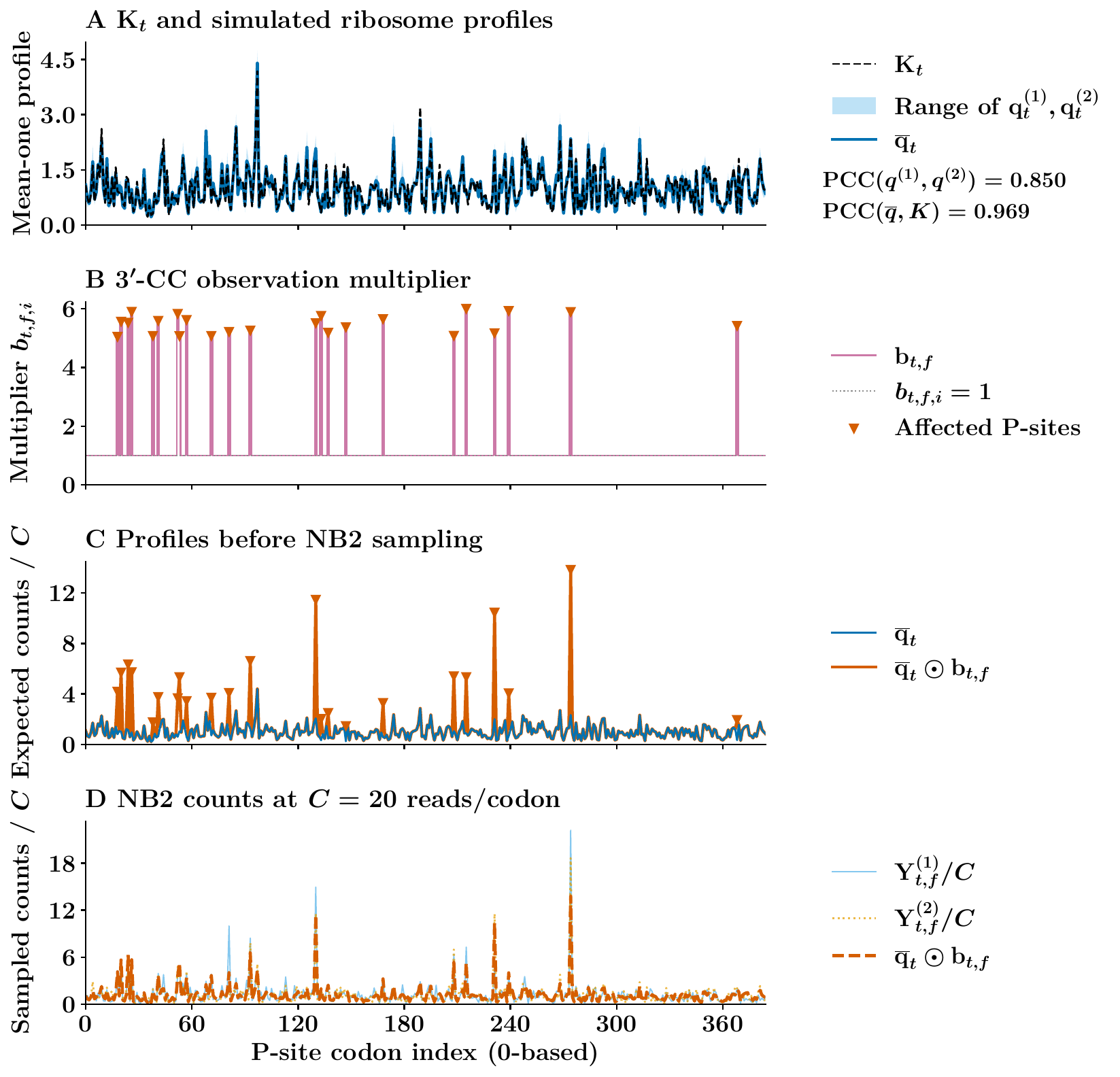}
    \caption{
    \textbf{Synthetic-generation layers for transcript
    \texttt{ENST00000319974.6}.}
    \textbf{(A)}
    The deterministic programmed kinetic profile $\mathbf K_t$
    (black dashed), the mean simulated ribosome profile
    $\overline{\mathbf q}_t$ (blue), and a translucent band spanning
    the two separately simulated ribosome profiles
    $\mathbf q_t^{(1)}$ and $\mathbf q_t^{(2)}$ at every position.
    \textbf{(B)}
    The exact $3^\prime$-\texttt{CC} bias multiplier
    $\mathbf b_{t,f}$; red triangles mark affected P-site coordinates and
    the horizontal reference denotes $b_{t,f,i}=1$.
    \textbf{(C)}
    Panel C shows the mean simulated ribosome profile
    $\overline{\mathbf q}_t$ and the biased expected profile per unit depth
    $\overline{\mathbf q}_t\odot\mathbf b_{t,f}$ before count sampling.
    \textbf{(D)}
    The biased expectation and the two saved NB2 count replicates divided
    only by the nominal depth $C=20$ for display. The biased
    expectations and depth-adjusted counts are not renormalized by their
    positional means. Zeros and peaks are retained without smoothing,
    clipping, interpolation, or pseudocounts. The same two simulations are reused across bias conditions and depths. $\overline{\mathbf q}_t$ is their mean is their mean simulated ribosome profile.
    }
    \label{fig:synthetic_profile_layers_example}
\end{figure*}

These examples distinguish changes in the expected profile from variation introduced by simulation and count sampling. \cref{fig:synthetic_profile_layers_example} illustrates the complete generation process for one transcript. The example shows variation between simulated ribosome profiles, changes caused by observation bias, and additional variation from NB2 count sampling.

Each sampled profile therefore contains two distinct stochastic
components: variation between finite TASEP trajectories and conditional
NB2 count-sampling noise. All bias conditions and sequencing depths use the same two simulated ribosome profiles.

\begin{figure*}[p]
    \centering
    \includegraphics[width=\textwidth,height=0.72\textheight,keepaspectratio]
        {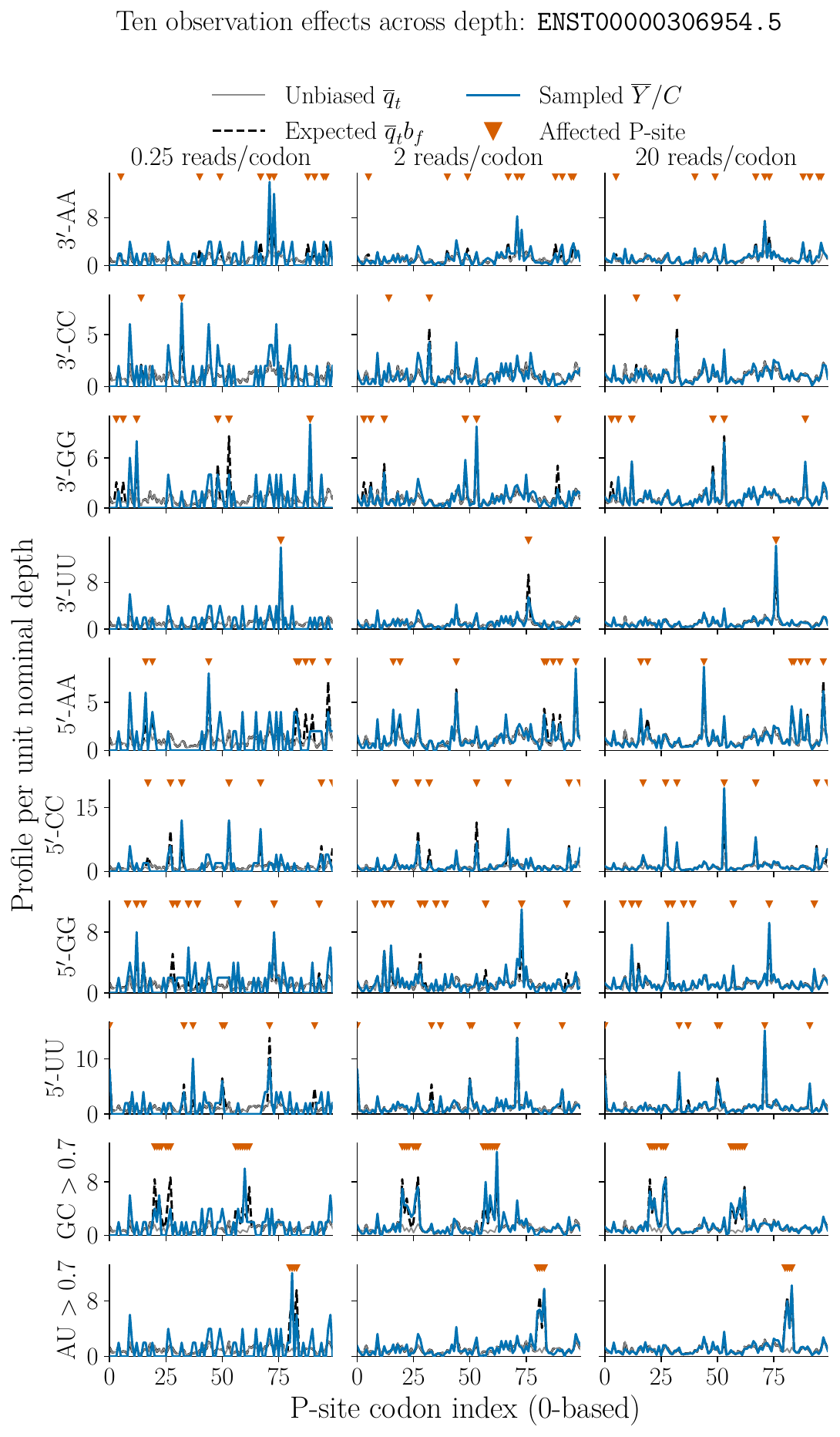}
    \caption{
    \textbf{Ten deterministic observation effects across sequencing
    depths.}
    Transcript \texttt{ENST00000306954.5} is the shortest transcript in
    the prespecified 100--149-sense-codon range with at least one saved
    affected P-site for every condition. Rows correspond to the ten bias conditions, and columns to nominal sequencing depths $C\in\{0.25,2,20\}$ expected reads per codon. Each cell shows the
    mean simulated ribosome profile
    $\overline{\mathbf q}_t$ (light gray), the exact biased expectation
    per unit depth
    $\overline{\mathbf q}_t\odot\mathbf b_{t,f}$ (black dashed), and the
    arithmetic mean of the two sampled NB2 count replicates divided only by
    $C$ (blue). Red triangles identify the exact affected coordinates.
    The three panels within a row share one y-axis range; different rows
    may differ. The biased expectation and sampled profiles are not
    normalized by their positional means. All zeros and peaks are
    retained without smoothing, clipping, interpolation, or
    pseudocounts. All conditions and depths reuse the same two finite simulations, whose mean simulated ribosome profile is $\overline{\mathbf q}_t$.
    }
    \label{fig:synthetic_bias_depth_examples}
\end{figure*}

\cref{fig:synthetic_bias_depth_examples} shows all ten observation biases at each sequencing depth. Dividing counts by nominal sequencing depth $C$ puts the three depths on the same expected scale while retaining changes in total expected count caused by the observation bias. As sequencing depth increases, depth-normalized counts fluctuate less around the biased expectation, while low-depth profiles show greater sampling variability and more zeros. With fixed NB2 overdispersion, residual sampling variability remains even at high depth.

\section{Synthetic-data diagnostics}
\label{sec:synthetic_data_diagnostics}
\label{sec:synthetic_diagnostic_rationale}

Before evaluating RiboUnmix, we examine how sequencing depth,
replicate averaging, and observation bias affect the synthetic
profiles themselves. Of the 19,290 transcripts in the synthetic cohort, seven were excluded because they lacked a canonical stop codon, leaving 19,283 transcripts for analysis. We analyze the ten bias conditions for these transcripts at nominal sequencing depths $C\in\{0.25,2,20\}$ reads per codon.
All comparisons use aligned sense-codon positions, and Pearson
correlation is calculated separately within each transcript over every position.

Each bias condition applies a different sequence-dependent multiplier $\mathbf b_{t,f}$ to the same simulated ribosome profiles.
In the figure labels, $5^\prime$ and $3^\prime$ indicate the ends of the idealized ribosome-protected fragment, and AA, CC, GG, and UU denote terminal dinucleotides. AU-rich and GC-rich denote fragments containing more than 70\% A and U, or G and C, respectively.

\subsection{Replicate agreement across sequencing depths}
\label{sec:synthetic_repeatability_similarity}

\begin{figure*}[!htbp]
    \centering
    \includegraphics[width=\textwidth]
        {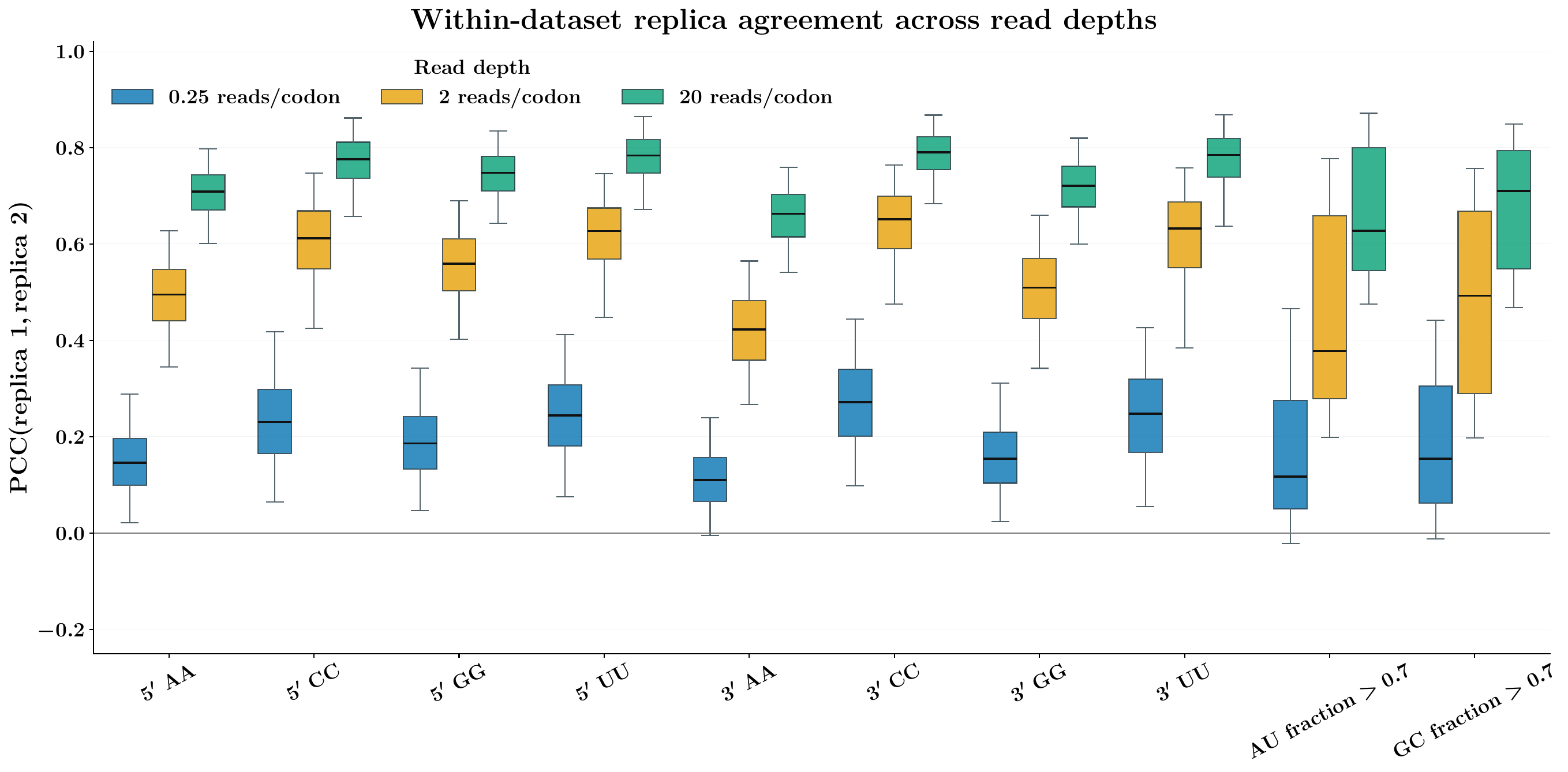}
    \caption{
    \textbf{Replicate agreement across sequencing depths.}
    Each box summarizes the within-transcript Pearson correlation
    between $\mathbf Y_{t,f}^{(1)}$ and $\mathbf Y_{t,f}^{(2)}$
    for one bias condition and depth.
    Boxes show medians and interquartile ranges; whiskers show the
    5th and 95th percentiles across 19,283 transcripts.}
    \label{fig:synthetic_replica_agreement}
\end{figure*}

We first compare the two count replicates within each bias
condition and sequencing depth (Figure~\ref{fig:synthetic_replica_agreement}).
Median transcript-level Pearson correlation increases with sequencing depth, consistent with reduced relative count-sampling noise. Differences remain because each count replicate is generated from a different simulated ribosome profile and a separate NB2 count sample. Replicate disagreement therefore reflects both sources of variation.

\subsection{Agreement with the expected and unbiased profiles}
\label{sec:synthetic_sampling_fidelity}

The simulator provides the expected profile before count sampling.
For count replicate $r$ under bias condition $f$, the expected profile is
$C\mathbf q_t^{(r)}\odot\mathbf b_{t,f}$.
Averaging the two expected profiles gives
$C\overline{\mathbf q}_t\odot\mathbf b_{t,f}$.
We use these known expectations to distinguish count-sampling
variation from the systematic change introduced by the bias.

\begin{figure*}[!htbp]
    \centering
    \includegraphics[width=\textwidth]
        {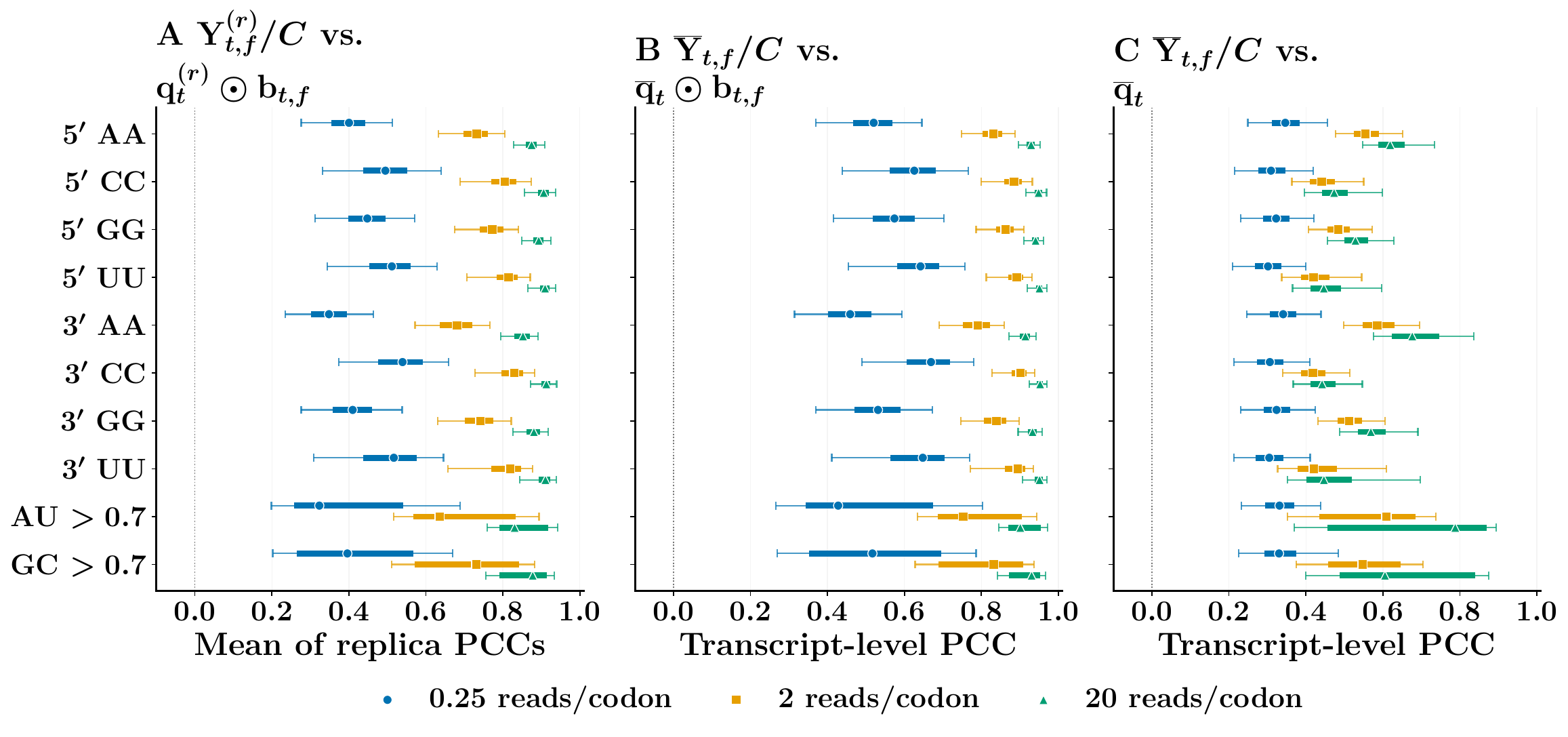}
    \caption{
    \textbf{Agreement with expected and unbiased profiles.}
    \textbf{(A)} Pearson correlation of each count replicate
    $\mathbf Y_{t,f}^{(r)}$ with its expected profile per unit nominal depth
    $\mathbf q_t^{(r)}\odot\mathbf b_{t,f}$, averaged over the
    two replicates within each transcript.
    \textbf{(B)} Pearson correlation of the mean count profile
    $\overline{\mathbf Y}_{t,f}$ with the expected profile per unit nominal depth,
    $\overline{\mathbf q}_t\odot\mathbf b_{t,f}$.
    \textbf{(C)} Pearson correlation of $\overline{\mathbf Y}_{t,f}$ with
    the mean simulated ribosome profile $\overline{\mathbf q}_t$, before observation bias is applied.
    Multiplication or division by nominal depth does not change
    Pearson correlation.
    Colors denote sequencing depth. Points show medians, thick
    intervals show interquartile ranges, and thin intervals show
    the 5th--95th percentiles across 19,283 transcripts.}
    \label{fig:synthetic_observation_layer_agreement}
\end{figure*}

Figure~\ref{fig:synthetic_observation_layer_agreement} compares each count replicate with its expected count profile (panel A), the mean of the two count replicates with its expectation (panel B), and that mean with the mean simulated ribosome profile
$\overline{\mathbf q}_t$ (panel C).
Agreement with the expected biased profile increases with depth
and improves further when the two replicates are averaged.
Agreement with the mean simulated ribosome profile $\overline{\mathbf q}_t$ remains lower than agreement with the biased expectation because the observation bias changes the positional profile.
Thus, deeper sequencing and replicate averaging improve measurement
of the biased profile without removing the bias.

\subsection{Agreement between bias conditions}
\label{sec:synthetic_cross_condition_diagnostics}

\begin{figure*}[!htbp]
    \centering
    \includegraphics[width=\textwidth]
        {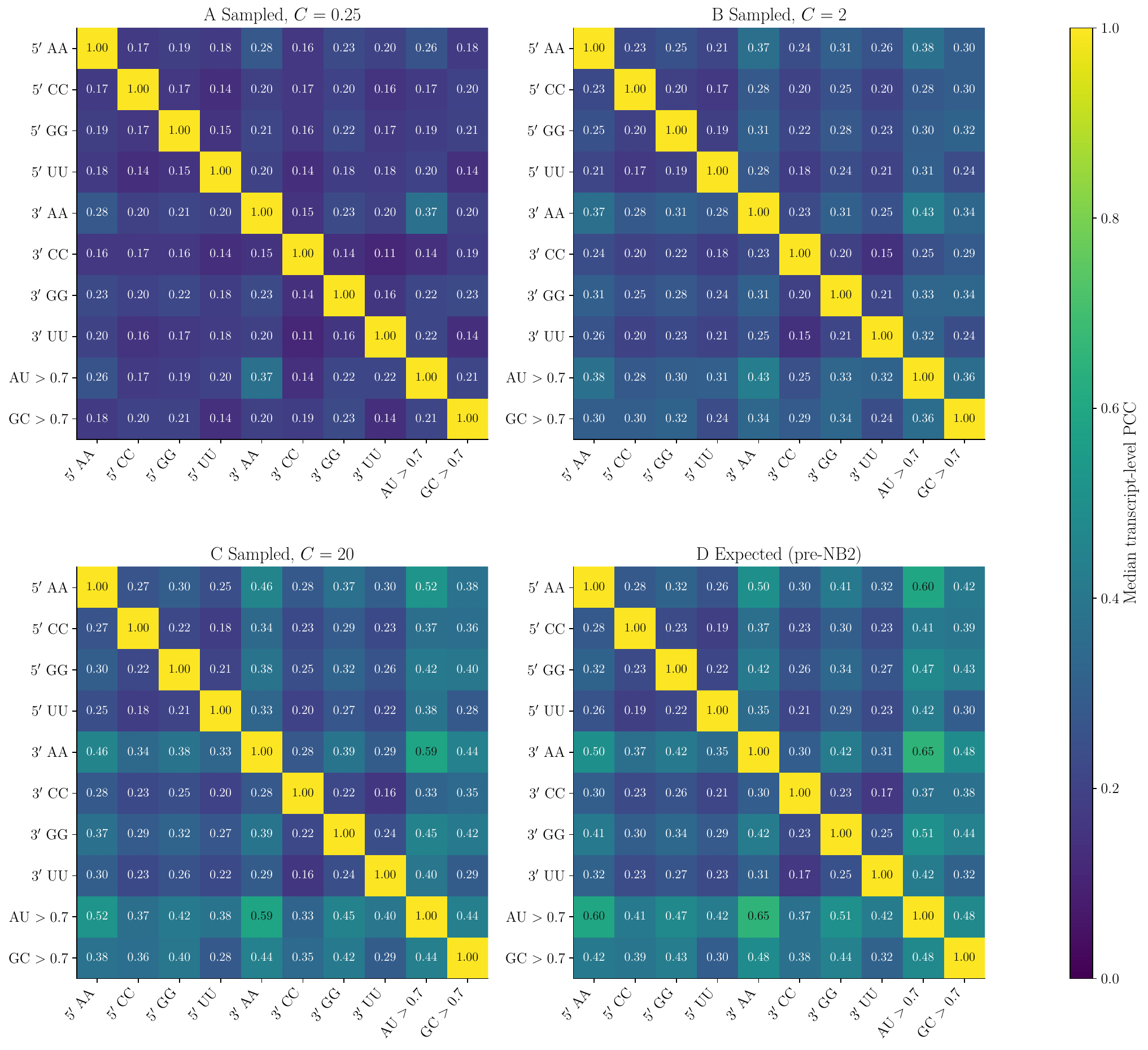}
    \caption{
    \textbf{Agreement between bias conditions before and after
    count sampling.}
    \textbf{(A--C)} Median within-transcript Pearson correlation
    between $\overline{\mathbf Y}_{t,f}$ and
    $\overline{\mathbf Y}_{t,g}$ for each pair of conditions
    at $C=0.25,2,20$ reads per codon.
    \textbf{(D)} The corresponding correlations between the
    expected profiles per nominal depth
    $\overline{\mathbf q}_t\odot\mathbf b_{t,f}$ and
    $\overline{\mathbf q}_t\odot\mathbf b_{t,g}$.
    All matrices use the same condition order and color scale.}
    \label{fig:synthetic_cross_dataset_agreement}
\end{figure*}

We next compare different bias conditions for the same transcript.
Figure~\ref{fig:synthetic_cross_dataset_agreement} shows pairwise Pearson correlations between bias conditions, calculated from the mean count profiles and from their expected profiles before count sampling.
Across the tested sequencing depths, correlations between the sampled count profiles move closer to those between the expected profiles. The expected profiles differ across bias conditions because their multipliers affect different positions with different strengths.

All conditions reuse the same two simulated ribosome profiles.
Checks of the generated counts suggest dependence in count sampling across bias conditions: positions unaffected by either bias have matching counts more often than expected under independent sampling. This dependence can increase agreement between sampled
conditions. Comparisons between the known expected profiles contain no NB2 count-sampling noise and are unaffected by this count-sampling dependence.

\subsection{Agreement with the programmed kinetic profile}
\label{sec:synthetic_diagnostic_kinetics}

\begin{figure*}[!htbp]
    \centering
    \includegraphics[width=\textwidth]
        {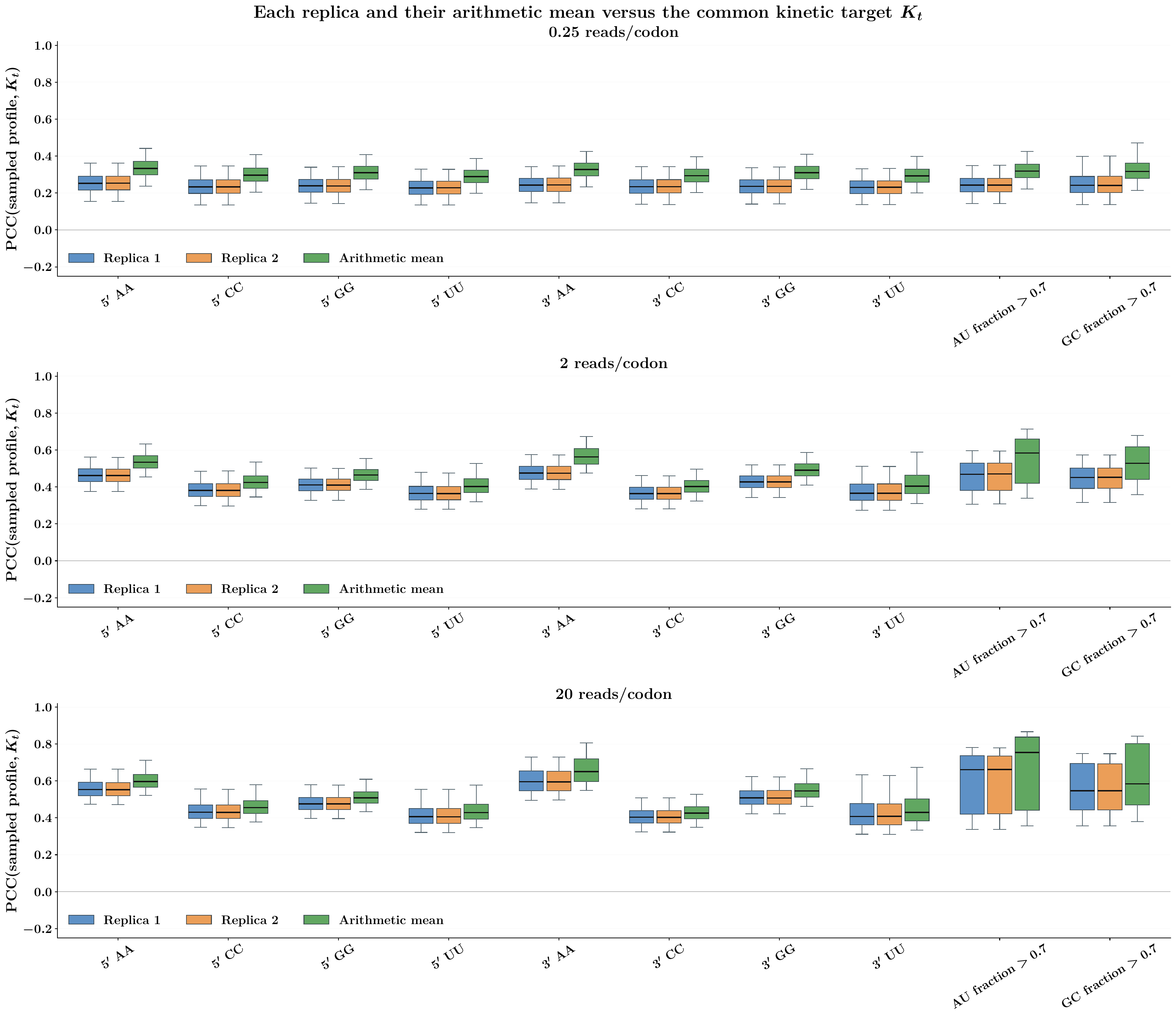}
    \caption{
    \textbf{Agreement with the programmed kinetic profile.}
    Blue and orange show within-transcript Pearson correlations
    of the two count replicates with $\mathbf K_t$.
    Green shows the correlation of their arithmetic mean with
    $\mathbf K_t$.
    Boxes show medians and interquartile ranges; whiskers show
    the 5th--95th percentiles.}
    \label{fig:synthetic_kinetic_agreement}
\end{figure*}

Finally, we compare each count replicate and their mean with the programmed kinetic profile
$\mathbf K_t$ used to parameterize ribosome movement
(Figure~\ref{fig:synthetic_kinetic_agreement}).
This comparison reflects ribosome traffic, variation between finite simulations, changes introduced by observation biases, and additional count-sampling noise.

Averaging the two count replicates generally increases median transcript-level PCC with
$\mathbf K_t$. However, high agreement
between replicates does not necessarily imply high agreement with
$\mathbf K_t$, because both replicates contain the same systematic
observation bias.

These diagnostics show why reconstruction and recovery require
separate evaluation. Higher sequencing depth and replicate averaging improve agreement with the expected count profile, which still contains the injected observation bias. Appendix \ref{sec:synthetic_evaluation_logic} therefore evaluates prediction of sampled counts, recovery of the simulated ribosome profile, and agreement between the learned dataset-specific corrections and the injected observation biases after applying the model’s centering.
\section{Synthetic training and component recovery}
\label{sec:synthetic_training_results}
\label{sec:synthetic_evaluation_logic}

The observation diagnostics establish that sequencing depth changes count
noise while the observation multipliers persist. We now test whether joint training recovers the simulated ribosome profile and the injected observation biases under the model’s centering convention. We evaluate observation reconstruction, shared-profile recovery, dataset-specific correction recovery, and total-count calibration.

\subsection{Experimental design and training settings}
\label{sec:synthetic_training_settings}
\label{sec:synthetic_training_preprocessing}
\label{app:training_configuration}

\paragraph{Collections and transcript splits.}
The processed count tables contain 19,283 exactly aligned transcripts.
The sequence and length criteria leave 19,204 transcripts: 17,284 for
training and 1,920 for validation. The maximum modeled CDS length is
4,000 codons, longer are excluded. Each dataset retains two raw count replicas; their mean
enters the correlation terms as defined in \cref{app:replica_handling}.
Input reliability weights are normalized within each dataset to median
one. We fit three complementary sets of models:
\begin{enumerate}
    \item \textbf{Single-dataset controls:} fit each observation condition
    separately at each depth to measure reconstruction of its count profiles.
    \item \textbf{Separate-depth series:} cumulatively add bias families at
    one fixed depth, giving $N=2,\ldots,10$ datasets per model.
    \item \textbf{Joint-depth series:} include all three depths for every
    added bias family, giving $N=3B$ datasets for $B=1,\ldots,10$ families.
\end{enumerate}
The cumulative bias order, saved TASEP trajectories, and training seed 42
are fixed. Equal reference weights $\pi_d=1/N$ define the primary series.
The joint-depth comparison also evaluates a depth-ranked reference.
Each panel is trained separately from initialization.
Here $d$ indexes one condition--depth dataset $(f,C)$; its multiplier
$\mathbf b_{t,d}=\mathbf b_{t,f}$ is shared across depths.

\paragraph{Architecture and optimization.}
The shared encoder is a two-layer bidirectional GRU with 256 hidden units
per direction and zero recurrent dropout. The dataset-conditioned branch
uses dataset, codon, nucleotide, and amino-acid embeddings of dimensions
32, 16, 4 per nucleotide, and 8, together with relative/start/stop position
features. Its two-layer bidirectional context GRU has hidden size 128;
the correction and log-dispersion heads each have a 128-unit hidden layer
and dropout 0.1. Log dispersion is learned and clamped to $[-5,1]$.
The decoder applies the centering in \cref{eq:gamma_centering} without normalizing \(L_t\odot\gamma_{t,d}\) by its positional mean. Training uses the composite objective and transcript-balanced reduction in \cref{app:training_details}.

AdamW uses learning rates $5\times10^{-4}$ for the shared encoder,
$10^{-3}$ for the remaining mean-model parameters, and $10^{-4}$ for the
dispersion head, with weight decay $10^{-2}$. A validation-loss
ReduceLROnPlateau schedule uses factor 0.99, patience 5, and minimum
learning rate $10^{-6}$, scaled by 0.1 for the dispersion group.
Training stops after 10 epochs without validation-loss improvement or
at 150 epochs. Every recovery analysis uses the minimum-validation-loss
checkpoint; simulator targets do not select checkpoints.

\paragraph{Execution and evaluation.}
Each fit uses one GPU with BF16 mixed precision. The logical per-dataset
quota and the target number of unique transcripts per optimizer update
are both 32. Execution limits are 1,024 pair rows, 600,000 padded codon
tokens, and 16 complete transcript groups per forward. Global-norm
clipping at 1 follows accumulation of the complete logical objective.
These limits partition execution and do not redefine the loss.

For shared-profile evaluation, we remove the appended terminal entry and ten sense codons from each end, retaining 1,919 validation transcripts at $C=0.25$ and $C=2$, and 1,920 at $C=20$. IDs are matched across panel
sizes within a depth; the depth-specific splits differ, so depth curves
are not paired transcript-level interventions. The joint-depth audit
uses one common 1,920-transcript validation cohort. Correction recovery and comparison with $K_t$ exclude five sense codons from each end. Intervals resample whole transcripts and are
conditional on these fitted models.

Validation transcripts receive no gradient updates, but they also select
checkpoints. These are therefore validation analyses, with no independent
synthetic test cohort. One seed and one cumulative bias ordering support
comparisons along the recorded training path; they do not establish
robustness to arbitrary panel composition. The evidence of count-sampling dependence reported in Appendix  \ref{sec:synthetic_cross_condition_diagnostics} motivates confirmation with independently resampled counts.

\subsection{Reconstructing the observed profiles}
\label{sec:synthetic_depth_reconstruction}

\begin{figure*}[!htbp]
\centering
\includegraphics[width=\textwidth]
{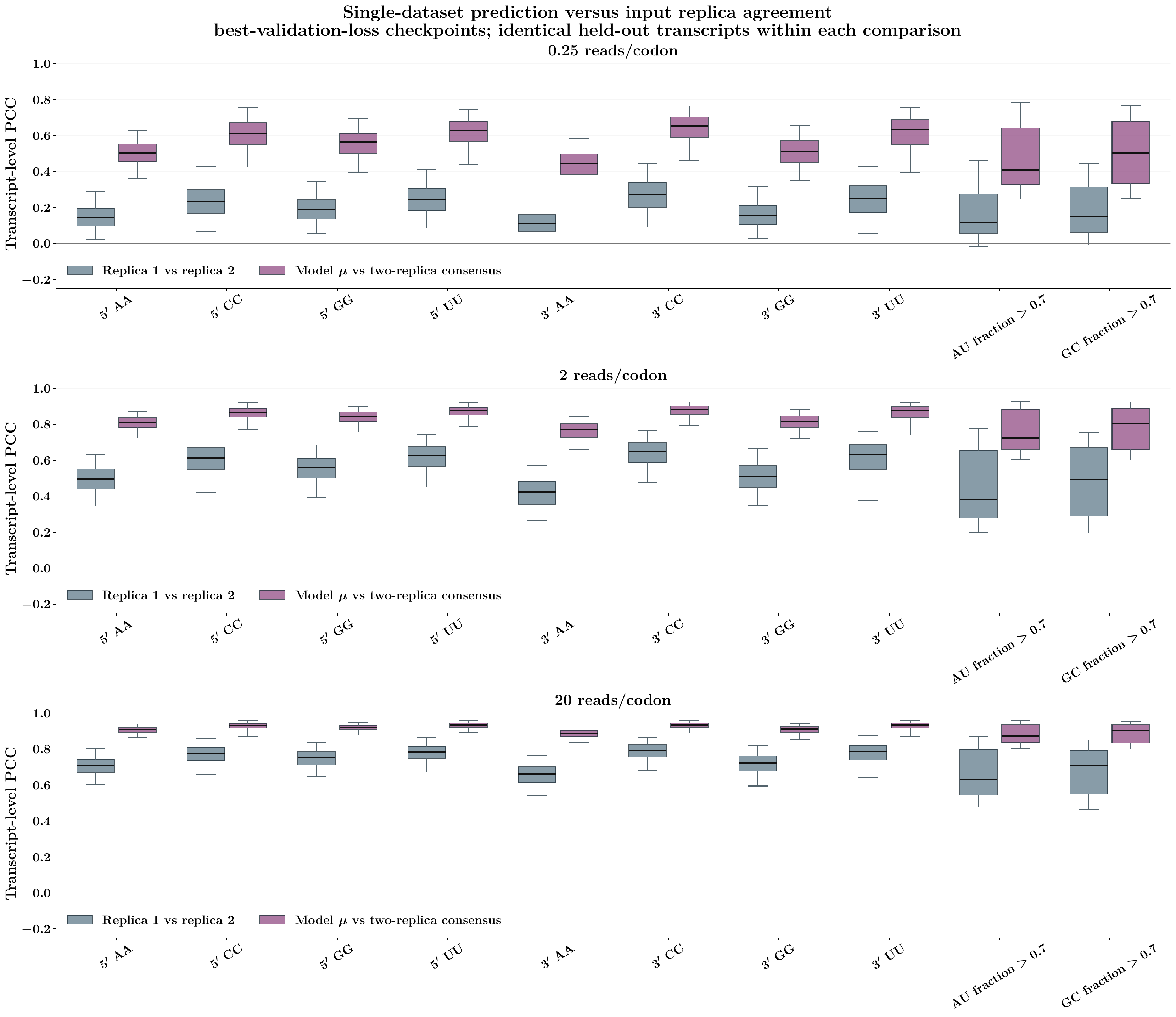}
\caption{\textbf{Single-dataset reconstruction and count-replicate agreement.}
For each observation condition and depth, gray boxes show
$\operatorname{PCC}(\mathbf Y_{t,d}^{(1)},\mathbf Y_{t,d}^{(2)})$ and
purple boxes show
$\operatorname{PCC}(\boldsymbol\mu_{t,d},\overline{\mathbf Y}_{t,d})$
from the minimum-validation-loss model. Boxes show medians and
interquartile ranges; whiskers show the 5th--95th percentiles.
The fitted mean is evaluated on the observed-count scale.}
\label{fig:synthetic_single_dataset_reconstruction}
\end{figure*}

If low depth mainly obscures the sampled positional pattern, reconstruction
should improve as depth increases. We first train each dataset separately and compare the fitted mean $\boldsymbol\mu_{t,d}$ with the replicate-average count profile $\overline{\mathbf Y}_{t,d}$. Figure~\ref{fig:synthetic_single_dataset_reconstruction}
shows improved reconstruction at greater depth, consistent with the
simulator-only comparison in \cref{sec:synthetic_sampling_fidelity}.
The two count replicas provide a repeatability reference.

Prediction--mean agreement can exceed replica--replica agreement because
averaging removes some replica-specific variation and the sequence model
shares information across training transcripts. Replica agreement is
therefore not a formal performance ceiling. The paired reconstruction audit
also finds little change when the ten conditions are trained jointly
instead of separately. Joint factorization preserves observation fit in
this comparison, but the following component-level tests are needed to
determine what the shared profile represents.

\subsection{Defining the shared-profile comparison}
\label{sec:synthetic_identifiable_target}

The programmed kinetic profile \(\mathbf K_t\), the mean simulated ribosome profile
\(\overline{\mathbf q}_t\), and the expected observed profiles represent different
levels of the simulator. The step
\(\mathbf K_t\rightarrow\overline{\mathbf q}_t\) captures ribosome traffic and finite simulation variation,
whereas
\(\overline{\mathbf q}_t\rightarrow\overline{\mathbf q}_t\odot\mathbf b_{t,d}\)
applies the observation bias. We therefore use \(\overline q_t\) to assess recovery of simulated ribosome occupancy and retain \(K_t\) as a secondary comparison with programmed kinetics.

Let \(\mathcal D_N\) denote the collection of \(N\) datasets used for
joint training, with reference weights \(\pi_d\). The model removes the
multiplicative ambiguity between its shared profile and dataset-specific
corrections by imposing
\begin{equation}
\prod_{d\in\mathcal D_N}
\gamma_{t,d,i}^{\pi_d}
=
1,
\qquad
\sum_{d\in\mathcal D_N}\pi_d=1.
\label{eq:synthetic_reference_constraint}
\end{equation}

The model additionally enforces positional log-centering,
$\langle \log \gamma_{t,d} \rangle_i = 0$, for each dataset
and normalizes the shared profile so that $\langle L_t \rangle_i = 1$.
For each dataset, define the noiseless expected relative profile
\(\mathbf h_{t,d}^{\star}\) by dividing
\(\overline{\mathbf q}_t\odot\mathbf b_{t,d}\) by its positional mean. We define the
reference-selected shape as the normalized reference-weighted
geometric mean of these profiles:
\begin{equation}
\mathbf H_t^{(N)}
=
\frac{
\displaystyle
\prod_{d\in\mathcal D_N}
\bigl(
\mathbf h_{t,d}^{\star}
\bigr)^{\pi_d}
}{
\displaystyle
\left\langle
\prod_{d\in\mathcal D_N}
\bigl(
\mathbf h_{t,d}^{\star}
\bigr)^{\pi_d}
\right\rangle_i
}
=
\frac{
\displaystyle
\overline{\mathbf q}_t\odot
\prod_{d\in\mathcal D_N}
\mathbf b_{t,d}^{\pi_d}
}{
\displaystyle
\left\langle
\overline{\mathbf q}_t\odot
\prod_{d\in\mathcal D_N}
\mathbf b_{t,d}^{\pi_d}
\right\rangle_i
}.
\label{eq:synthetic_reference_target}
\end{equation}
All products and powers are element-wise. With the uniform weights used
in the separate-depth experiments, \(\pi_d=1/N\).

The two expressions in
Equation~\ref{eq:synthetic_reference_target} are algebraically equivalent because the dataset-specific normalization constants cancel.
The first emphasizes that \(\mathbf H_t^{(N)}\) is determined from the
noiseless expected relative profiles. Thus, \(H_t^{(N)}\) is the mean simulated ribosome profile multiplied by the panel’s geometric-mean observation bias and renormalized to mean one.

If the noiseless expected relative profiles satisfy
Equation~\ref{eq:app_profile_compatibility} and the fitted model reconstructs
them exactly, then \(\widehat{\mathbf L}_t=\mathbf H_t^{(N)}\). In general,
however, the unnormalized decoder cannot exactly represent an arbitrary
collection of positive mean-one profiles under both gauges
(\cref{app:gamma_existence}). We therefore use \(\mathbf H_t^{(N)}\) as a
reference-selected comparison, not as an established population
optimum. The fitted \(\widehat L_t\) depends on the composite training objective, the model constraints and optimization. Positional distortion retained in the panel's
geometric mean remains inseparable from the simulated ribosome profile under this reference
convention.

\subsection{Recovery of the shared target and simulated ribosome profile}
\label{sec:synthetic_component_recovery}

We separate three comparisons: the learned shared profile
\(\widehat{\mathbf L}_t\) versus its reference-selected shape
\(\mathbf H_t^{(N)}\), the reference-selected shape \(\mathbf H_t^{(N)}\) versus the
mean simulated ribosome profile \(\overline{\mathbf q}_t\), and the end-to-end comparison of
\(\widehat{\mathbf L}_t\) with \(\overline{\mathbf q}_t\).

Across the separate-depth series, the fitted profile follows its
reference-selected target closely, including with few bias families.
The large change with panel size instead lies in the target itself:
as complementary multipliers are added, their geometric mean becomes
flatter and $\mathbf H_t^{(N)}$ moves toward $\overline{\mathbf q}_t$.
The fitted $\widehat{\mathbf L}_t$ consequently becomes more similar to
the mean simulated ribosome profile.

For a fixed transcript and collection of bias families, this target change
does not depend on nominal depth: the noiseless multipliers and trajectories
are the same. Depth changes the counts from which the network must learn.
The validation cohorts have similar sizes but different transcript identities, so comparisons across depths are unpaired.
This distinction prevents attributing all improvement with panel size to
easier optimization, or interpreting more reads as removal of systematic bias.

\paragraph{Secondary alignment with programmed kinetics.}
The same best-validation-loss profiles also become more similar to the
upstream kinetic target $\mathbf K_t$. This secondary audit uses the
saved model mask after removing the appended terminal entry and five sense
codons from each end; all 1,920 validation transcripts remain eligible.
This is mechanistically useful,
but it combines the programmed-kinetics-to-traffic transformation with
the observation-factorization problem.  We therefore use
$\widehat{\mathbf L}_t$ versus $\overline{\mathbf q}_t$---not
$\widehat{\mathbf L}_t$ versus $\mathbf K_t$---as the primary
end-to-end shared-profile comparison.

\begin{figure*}[t]
\centering
\includegraphics[width=\textwidth]
{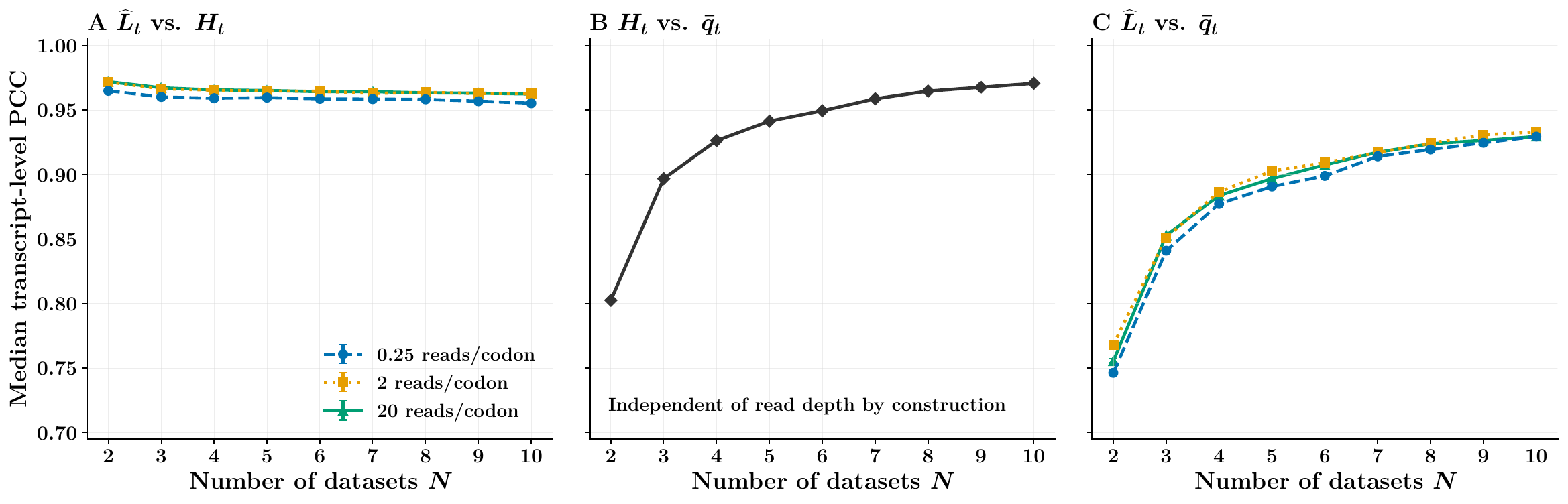}
\caption{\textbf{Agreement with the reference-selected shape and
mean simulated ribosome profile.}
For every panel size, connected points show median transcript-level Pearson
correlations and vertical intervals show 95\% transcript-bootstrap intervals for
\textbf{(A)} \(\widehat{\mathbf L}_t\) versus
\(\mathbf H_t^{(N)}\),
\textbf{(B)} \(\mathbf H_t^{(N)}\) versus \(\overline{\mathbf q}_t\), and
\textbf{(C)} \(\widehat{\mathbf L}_t\) versus \(\overline{\mathbf q}_t\).
Panels A and C use 1,919, 1,919, and 1,920 eligible validation
transcripts matched across all panel sizes at increasing depth. Panel B is depth-independent
by construction and uses the deduplicated union of 5,188 validation
identities. Intervals are conditional on one training seed and one
cumulative bias ordering; they quantify transcript heterogeneity, not
variation across independently sampled panels.}
\label{fig:synthetic_reference_target_recovery}
\end{figure*}

These correlations compare different stages of recovery and should not be
subtracted as if they formed an additive error decomposition.

\begin{figure*}[t]
\centering
\includegraphics[width=0.72\textwidth]
{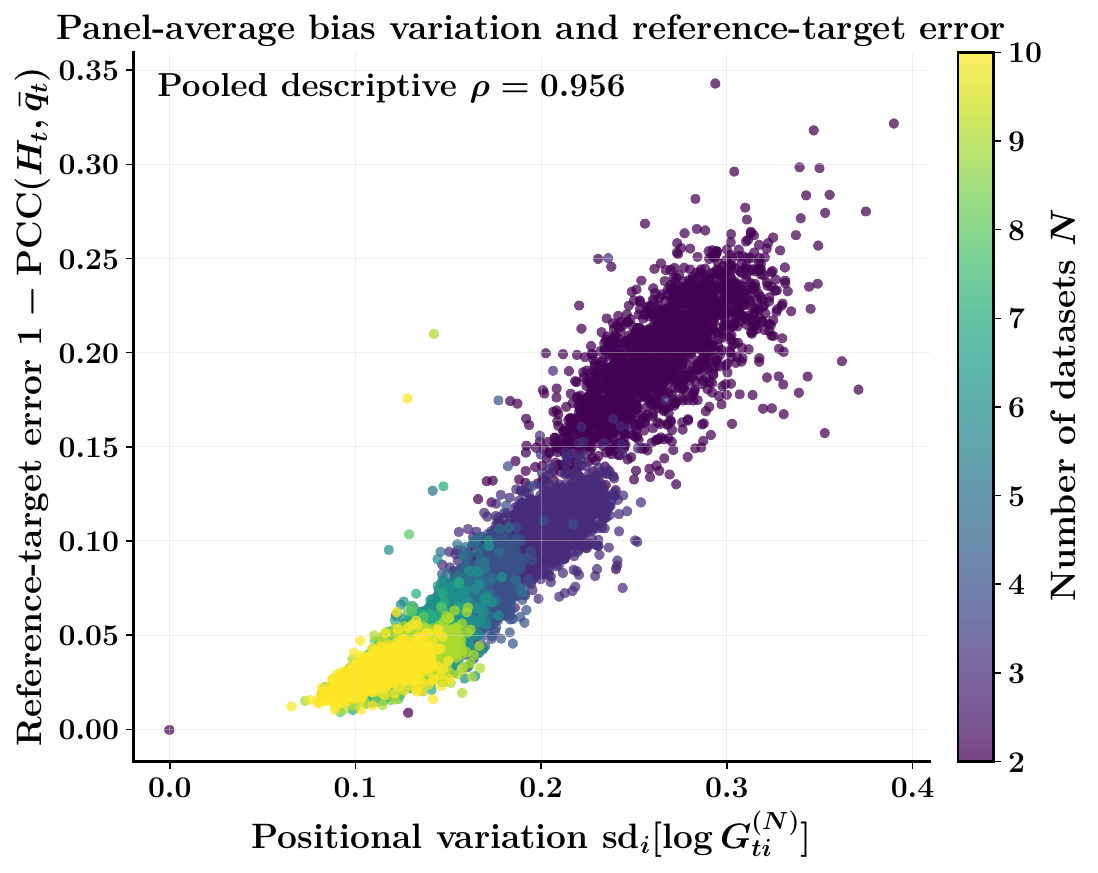}
\caption{\textbf{Panel-average bias variation and reference-target
error.}
Each point represents a transcript--panel pair. The horizontal axis
is the positional standard deviation of the log-geometric-mean
multiplier, and the vertical axis is the Pearson correlation error
between the reference-selected shape and the mean simulated ribosome profile.
Color denotes the number of datasets in the panel. The association is
descriptive because transcripts recur across panel sizes.}
\label{fig:synthetic_reference_error}
\end{figure*}

To check this explanation directly, define the panel's geometric-mean
multiplier as $G_{t,i}^{\mathrm{ref},N}=\prod_{d\in\mathcal D_N}b_{t,d,i}^{\pi_d}$.
Figure~\ref{fig:synthetic_reference_error} relates its positional log
variation to the discrepancy between $\mathbf H_t^{(N)}$ and
$\overline{\mathbf q}_t$. More variable multipliers accompany larger
discrepancies, as the explicit expression for $\mathbf H_t^{(N)}$ predicts.

Together, these comparisons support cancellation of the particular
observation effects along this cumulative path. Dataset count and
composition change together, so the result does not imply that adding
arbitrary datasets must improve recovery.

\subsection{Joint training across sequencing depths}
\label{sec:synthetic_mixed_depth_recovery}

The joint-depth experiments train on all three depths for every included
bias family. With $B=1,\ldots,10$ cumulative bias families, the panel
therefore contains $N=3B$ datasets. This design asks whether one shared
profile can be recovered when low-, medium-, and high-depth observations
are supplied simultaneously. It also compares equal reference weights
with depth-ranked weights in the ratio $1{:}2{:}3$ within each family.

\begin{figure*}[!htbp]
\centering
\includegraphics[width=\textwidth]
{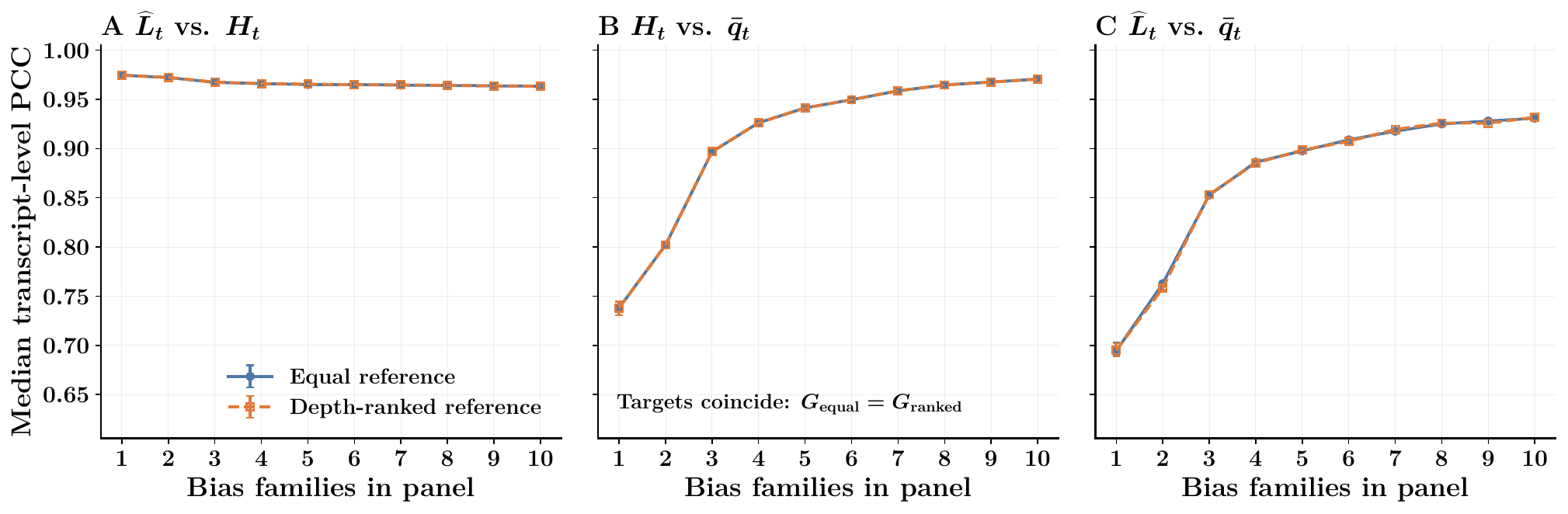}
\caption{\textbf{Shared-profile recovery when all read depths are
trained jointly.}
Each added bias family contributes one dataset at each of the three
nominal depths, so the horizontal axis corresponds to $N=3B$ datasets.
Points show median transcript-level PCC and error bars show 95\%
transcript-bootstrap intervals over the common 1,920-transcript
validation cohort. Equal and depth-ranked reference weights are shown
separately. The two schemes assign equal total weight to every bias
family and therefore define identical reference-selected shapes $\mathbf H_t^{(N)}$ in
this experiment; their comparison isolates only finite-training
sensitivity to the reference weighting.}
\label{fig:synthetic_mixed_depth_reference}
\end{figure*}

These experiments are a reference-weight sensitivity check, not an
independent panel-order experiment. Under both weighting schemes, every
bias family receives total reference mass $1/B$, and its programmed
multiplier is identical across depths. Consequently, the geometric
reference effect and the resulting \(\mathbf H_t^{(N)}\) are exactly the
same under the two conventions. Any separation between the curves in
Figure~\ref{fig:synthetic_mixed_depth_reference} is therefore a
finite-training difference in $\widehat{\mathbf L}_t$, not a change in
the reference-selected shape and not evidence that depth ranking is superior.

Both reference policies show the same broad improvement as bias families
are added. Their small differences change direction along the cumulative
series, supporting joint-depth recovery but not a systematic advantage of
depth-ranked centering.

\subsection{Recovery of the dataset-specific correction}
\label{sec:synthetic_gamma_recovery}

This comparison tests the other side of the factorization: whether
dataset-specific variation is assigned to the correction branch.
The raw programmed multiplier is not itself the gauge-fixed oracle comparison for
$\widehat{\boldsymbol\gamma}_{t,d}$.  Let
$a_{t,d,i}=\log b_{t,d,i}$,
$c_{t,i}=\sum_d\pi_d a_{t,d,i}$,
$m_{t,d}=\langle a_{t,d}\rangle_i$, and
$\overline m_t=\sum_d\pi_d m_{t,d}$.  The exact correction satisfying
the model's cross-dataset reference constraint and positional scale gauge is
\begin{equation}
g_{t,d,i}^{\star}
=
a_{t,d,i}-c_{t,i}-m_{t,d}+\overline m_t.
\label{eq:synthetic_gauge_fixed_gamma_target}
\end{equation}
We compare $\log\widehat\gamma_{t,d,i}$ with
$g_{t,d,i}^{\star}$ on the common 1,920-transcript validation cohort after
excluding five codons at each boundary.  The comparison uses every retained
transcript--dataset profile and the best-validation-loss checkpoint.

\begin{table}[tbp]
    \centering
    \caption{\textbf{Dataset-specific correction recovery for the
    30-dataset joint-depth panel.}
    Metrics compare learned and exact gauge-fixed log-corrections.  The
    calibration slope is from the pooled retained positions.}
    \label{tab:synthetic_gamma_recovery_mixed_depth}
    \small
    \begin{tabular}{@{}lccccc@{}}
        \toprule
        Reference
        & \shortstack{Mean pair\\PCC}
        & \shortstack{Pooled\\PCC}
        & \shortstack{Log-\\RMSE}
        & \shortstack{Calibration\\slope}
        & \shortstack{Affected sites\\within 10\%} \\
        \midrule
        Equal        & $0.9992$ & $0.9991$ & $0.0169$ & $0.9765$ & $94.79\%$ \\
        Depth-ranked & $0.9994$ & $0.9993$ & $0.0136$ & $0.9965$ & $97.83\%$ \\
        \bottomrule
    \end{tabular}
\end{table}

Both reference conventions therefore agree closely with the exact
gauge-fixed correction shapes in the largest joint-depth panel. At
$N=3$, all datasets share one programmed bias family and the gauge-fixed
target has zero positional variance, so its PCC is undefined rather than
zero. Equal and depth-ranked weighting assign the same total reference mass
to each bias family, and the multiplier for a family is identical across
depths. They therefore give the same $c_{t,i}$ and $\overline m_t$ in
Equation~\ref{eq:synthetic_gauge_fixed_gamma_target}, and hence the same exact
correction target for every corresponding dataset. Apparent differences
between learned corrections are finite-training differences: only one seed
and bias-family ordering are available, and the validation cohort also
selected the checkpoints.

\subsection{Aggregate profile mass}
\label{sec:synthetic_mass_analysis}
\label{sec:synthetic_deterministic_distortion}

The recovery comparisons above concern positional shape. Pearson
correlation is unchanged by a positive scalar multiplier, so those
comparisons cannot establish that total predicted counts are calibrated.
We therefore examine two different mass effects: the change deliberately
introduced by the simulator and the change permitted by the fitted decoder.

\subsubsection{Mass introduced by the observation multiplier}

\begin{figure*}[!htbp]
    \centering
    \includegraphics[width=\textwidth]
        {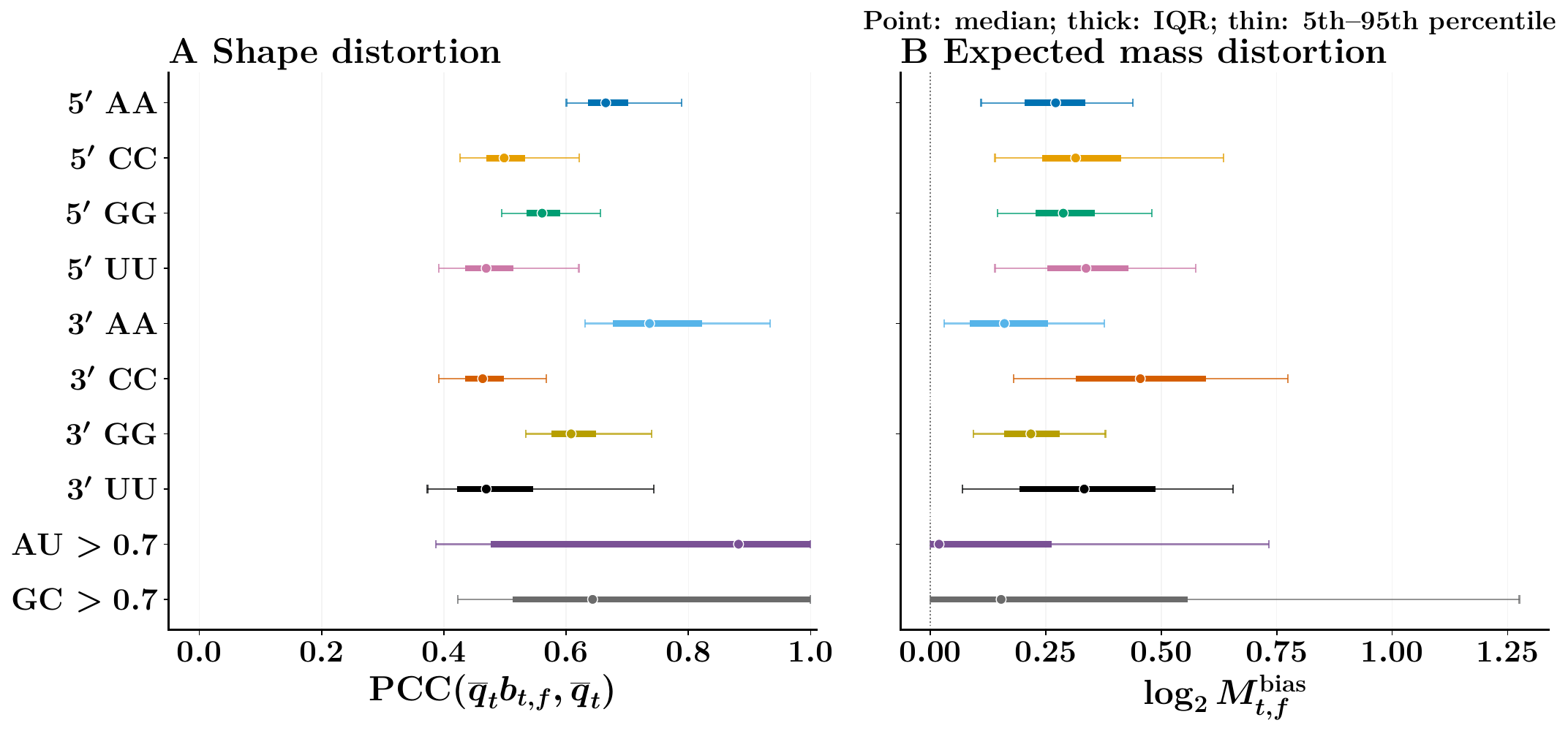}
    \caption{\textbf{Deterministic observation effects before count sampling.}
    \textbf{(A)} Transcript-level
    $\operatorname{PCC}(\overline{\mathbf q}_t\odot\mathbf b_{t,f},
    \overline{\mathbf q}_t)$.
    \textbf{(B)} $\log_2 M_{t,f}^{\mathrm{bias}}$, where zero denotes
    unchanged total expected counts. Points show medians, thick intervals
    interquartile ranges, and thin capped intervals the 5th--95th percentiles.
    Both panels use the 19,283-transcript simulator cohort before fitting.
    Applying the multiplier is not followed by positional-mean normalization.}
    \label{fig:synthetic_deterministic_bias_effects}
\end{figure*}

Before fitting any model, the expected total count under condition $f$ is
changed by
\begin{equation}
    M_{t,f}^{\mathrm{bias}}
    =\left\langle\overline{\mathbf q}_t\odot\mathbf b_{t,f}\right\rangle_i,
    \qquad
    \mathbb E\!\left[\sum_i\overline Y_{t,f,i}\,\middle|\,
      \mathbf q_t^{(1)},\mathbf q_t^{(2)}\right]
    =C\ell_t M_{t,f}^{\mathrm{bias}}.
    \label{eq:synthetic_bias_mass}
\end{equation}
Because $\langle\overline{\mathbf q}_t\rangle_i=1$, this is the expected
total-count multiplier relative to the unbiased condition.
Figure~\ref{fig:synthetic_deterministic_bias_effects} places this quantity
beside the corresponding shape agreement. A multiplier can alter a small
set of influential positions without producing the largest total-count
change. Shape distortion and mass distortion therefore require separate
measurements.

\subsubsection{Mass implied by the fitted factorization}

The decoder fixes $S_{t,d}$ to an observed arithmetic mean, while
fixed-reference gamma centering constrains geometric means.
It does not require
$\langle\widehat{\mathbf L}_t\odot\widehat{\boldsymbol\gamma}_{t,d}\rangle_i=1$.
Let $\boldsymbol\gamma_{t,d}^{\mathrm{ref},N}$ denote the oracle correction
paired with $\mathbf H_t^{(N)}$ under the same reference convention. The
oracle and learned masses are
\begin{equation}
    m_{t,d}^{\mathrm{ref}}
    =\left\langle\mathbf H_t^{(N)}
      \odot\boldsymbol\gamma_{t,d}^{\mathrm{ref},N}\right\rangle_i,
    \qquad
    \widehat m_{t,d}
    =\left\langle\widehat{\mathbf L}_t
      \odot\widehat{\boldsymbol\gamma}_{t,d}\right\rangle_i.
    \label{eq:synthetic_oracle_learned_mass}
\end{equation}
These quantities concern decoder calibration relative to $S_{t,d}$.
They are distinct from $M_{t,f}^{\mathrm{bias}}$, which compares simulated
expected totals before and after the observation effect.

Since $\mathbf H_t^{(N)}$ has positional mean one,
\begin{equation}
    m_{t,d}^{\mathrm{ref}}-1
    =\left(\left\langle\boldsymbol\gamma_{t,d}^{\mathrm{ref},N}
      \right\rangle_i-1\right)
     +\operatorname{Cov}_i\!\left(
       \mathbf H_t^{(N)},\boldsymbol\gamma_{t,d}^{\mathrm{ref},N}\right).
    \label{eq:synthetic_mass_decomposition}
\end{equation}
The first term is the arithmetic-mean excess of a geometrically centered
correction; the second describes its positional association with the
shared profile. The saved audit attributes most of the aggregate excess
to the first term.

\begin{figure*}[!htbp]
\centering
\includegraphics[width=0.82\textwidth]
{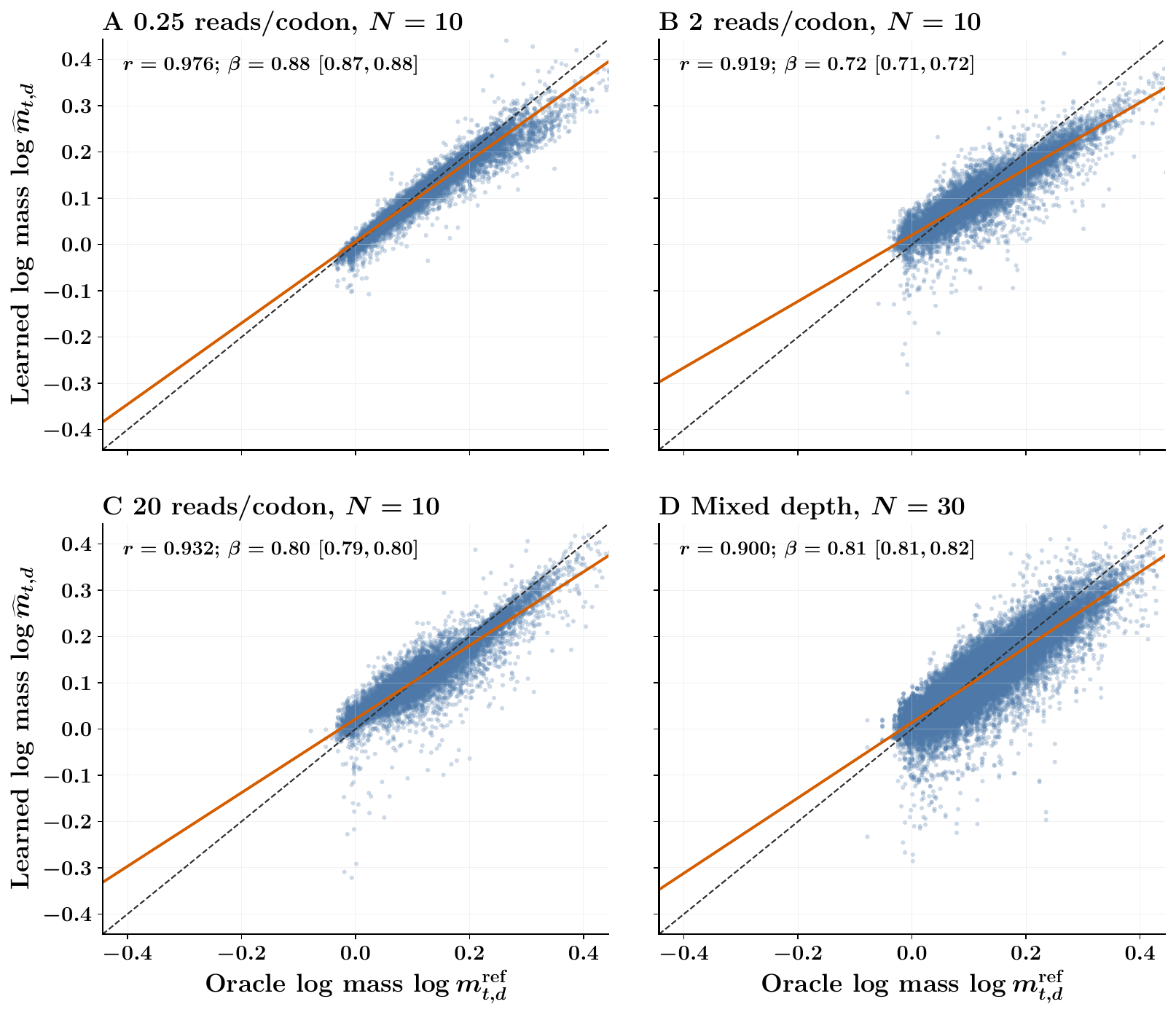}
\caption{\textbf{Learned versus oracle aggregate mass.}
Points compare learned and oracle log-mass values at representative
separate-depth endpoints and in the joint-depth experiment. Dashed lines
mark equality; orange lines are fitted calibration associations. Dense
point layers are rasterized inside the vector PDF; axes, labels, and
fitted lines remain vector graphics. Agreement with the oracle mass
does not imply mass one.}
\label{fig:synthetic_learned_oracle_mass}
\end{figure*}

Figure~\ref{fig:synthetic_learned_oracle_mass} shows that learned masses
track the oracle masses in both separate-depth and joint-depth fits.
This supports reproduction of the aggregate effect of the reference
parameterization. It does not establish equality between predicted and
observed totals: both decompositions can have mass above one, and the
fitted mean is $S_{t,d}\widehat m_{t,d}$. The discrepancy is a calibration
limitation that a shape correlation cannot detect.

\paragraph{What the synthetic experiments establish.}
For the recorded bias ordering, joint training preserves observation
reconstruction, follows the reference-selected shape $\mathbf H_t^{(N)}$,
and recovers the centered dataset corrections. Recovery of
$\overline{\mathbf q}_t$ improves as the included multipliers cancel in
their geometric mean. Greater depth reduces count noise, whereas adding
complementary conditions changes that reference shape. These are distinct
mechanisms. The validation-based evaluation, single seed, shared sampling
randomness, and aggregate-mass limitation delimit the claim; they leave
independent-test and broader panel-composition confirmation outstanding.

\section{HEK293 Ribosome Profile Datasets}
\label{app:hek293_ribo}
\subsection{Cohort and processing}
\label{app:hek293_cohort}
\label{app:hek293_processing}

The cohort comprises 114 HEK-derived control datasets from 86 GEO studies, containing 230 sample-level Ribo-seq profiles. 31 datasets contain one profile and 83 contain two to five profiles. Of these datasets, 81 derive from HEK293T, 27 from HEK293, and six from other HEK-derived cell lines. ``Control'' follows each source experiment's annotation (e.g., untreated, wild-type, vehicle, or mock), rather than implying identical experimental conditions. Dataset-specific differences may therefore include both technical effects and biological variation. \cref{tab:hek293_per_sample_metadata} lists the GEO study accessions, BioSample accessions, and number of sample profiles for each dataset..

\begingroup
\footnotesize
\setlength{\tabcolsep}{2.5pt}
\begin{longtable}{
    >{\raggedright\arraybackslash}p{0.28\linewidth}
    >{\raggedright\arraybackslash}p{0.11\linewidth}
    >{\raggedright\arraybackslash}p{0.43\linewidth}
    r}
\caption{GEO study and BioSample accessions, with sample-profile counts, for the 114 HEK-derived datasets.}
\label{tab:hek293_per_sample_metadata}\\
\toprule
Dataset name & GSE & Sample accession(s) & Replicates \\
\midrule
\endfirsthead
\multicolumn{4}{l}{\footnotesize Table~\thetable\ continued}\\
\toprule
Dataset name & GSE & Sample accession(s) & Replicates \\
\midrule
\endhead
\midrule
\multicolumn{4}{r}{\footnotesize Continued on next page}\\
\endfoot
\bottomrule
\endlastfoot
akichika\_2019 & GSE122071 & SAMN10359332 & 1 \\
andreev\_2015\_rep1 & GSE55195 & SAMN02646670 & 1 \\
andreev\_2015\_rep2 & GSE55195 & SAMN02646671 & 1 \\
apostolopoulos\_2024\_Cas13 & GSE232383 & SAMN35056827, SAMN35056828 & 2 \\
apostolopoulos\_2024\_dCas13 & GSE232383 & SAMN35056823, SAMN35056824 & 2 \\
barrington\_2023 & GSE202900 & SAMN28209264, SAMN28209265, SAMN28209266 & 3 \\
calviello\_2016 & GSE73136 & SAMN04093818 & 1 \\
chang\_2019 & GSE132725 & SAMN12056127, SAMN12056128 & 2 \\
cipullo\_2023 & GSE242965 & SAMN37386601 & 1 \\
cui\_2024 & GSE223418 & SAMN32851229, SAMN32851230, SAMN32851231 & 3 \\
dong\_2025 & GSE301911 & SAMN49830981, SAMN49830982 & 2 \\
eichhorn\_2014 & GSE52809 & SAMN02423228 & 1 \\
eliseeva\_2024 & GSE249895 & SAMN38765408, SAMN38765409 & 2 \\
gillen\_2021 & GSE158141 & SAMN16198221, SAMN16198222, SAMN16198223 & 3 \\
han\_2020 & GSE133393 & SAMN12145530, SAMN12145531 & 2 \\
han\_2021 & GSE166874 & SAMN17926858 & 1 \\
havkin-solomon\_2023 & GSE301928 & SAMN49826093, SAMN49826094, SAMN49826095 & 3 \\
hia\_2019 & GSE126298 & SAMN10889012, SAMN10889013 & 2 \\
huang\_2026 & GSE179871 & SAMN20166012, SAMN20166013 & 2 \\
ichihara\_2021 & GSE174329 & SAMN19116476, SAMN19116477 & 2 \\
ichihara\_2021\_chx & GSE174329 & SAMN19116480, SAMN19116481 & 2 \\
ichihara\_2021\_si & GSE174329 & SAMN19116432, SAMN19116433 & 2 \\
ichihara\_2025 & GSE288755 & SAMN46547843, SAMN46547844 & 2 \\
ichihara\_2025\_si & GSE288755 & SAMN46547831, SAMN46547832 & 2 \\
ingolia\_2012 & GSE37744 & SAMN00990790, SAMN00990791, SAMN00990792 & 3 \\
iwasaki\_2014 & GSE70211 & SAMN03788107, SAMN03788108, SAMN03788118, SAMN03788119 & 4 \\
iwasaki\_2018 & GSE102720 & SAMN07513440, SAMN07513444 & 2 \\
jia\_2025 & GSE277746 & SAMN43867648 & 1 \\
kashiwagi\_2021 & GSE174764 & SAMN19287229, SAMN19287234 & 2 \\
kito\_2023 & GSE186502 & SAMN22556762, SAMN22556763 & 2 \\
klauer\_2025 & GSE269144 & SAMN41693166, SAMN41693167, SAMN41693168 & 3 \\
koubek\_2025 & GSE294467 & SAMN49963852, SAMN49963853 & 2 \\
landthaler\_2020 & GSE93052 & SAMN06196954, SAMN06196955, SAMN06196956, SAMN06196957, SAMN06196959 & 5 \\
lee\_2025 & GSE282838 & SAMN45068961, SAMN45068962 & 2 \\
li\_2022\_1kcells & GSE151986 & SAMN15164538, SAMN15164539 & 2 \\
li\_2022\_50kcells & GSE151986 & SAMN15164540, SAMN15164541 & 2 \\
li\_2022\_high & GSE151986 & SAMN15164542, SAMN15164543 & 2 \\
li\_2022\_std & GSE151986 & SAMN15164536, SAMN15164537 & 2 \\
liu\_2019 & GSE107588 & SAMN08118010 & 1 \\
liu\_2025 & GSE269734 & SAMN41812977, SAMN41812981, SAMN41812985 & 3 \\
lyabin\_2020 & GSE130781 & SAMN11583004, SAMN11583008, SAMN11583009 & 3 \\
m6atrans\_2025 & GSE309588 & SAMN52066643, SAMN52066644, SAMN52066645 & 3 \\
mao\_2019 & GSE129194 & SAMN11316860, SAMN11316864 & 2 \\
mao\_2019\_aaa & GSE129194 & SAMN11316862 & 1 \\
mao\_2023 & GSE184825 & SAMN21855060, SAMN21855061, SAMN21855062, SAMN21855063 & 4 \\
mao\_2023\_chx & GSE184825 & SAMN21855045 & 1 \\
martinez\_2019 & GSE125218 & SAMN10760879, SAMN10760880 & 2 \\
matsuura\_suzuki\_2022 & GSE179854 & SAMN20164010, SAMN20164011 & 2 \\
mazor\_2018 & GSE112643 & SAMN08851427 & 1 \\
mazor\_2018\_chx & GSE112643 & SAMN08851428 & 1 \\
muller\_2023 & GSE214396 & SAMN31076922, SAMN31076923 & 2 \\
oconnell\_2025\_centri & GSE274847 & SAMN43193202, SAMN43193203, SAMN43193204 & 3 \\
oconnell\_2025\_nocentri & GSE274847 & SAMN43193197, SAMN43193199, SAMN43193201 & 3 \\
oconnell\_2025\_triton\_01 & GSE274847 & SAMN43193252, SAMN43193255, SAMN43193258 & 3 \\
oconnell\_2025\_triton\_05 & GSE274847 & SAMN43193243, SAMN43193246, SAMN43193249 & 3 \\
oconnell\_2025\_triton\_1 & GSE274847 & SAMN43193234, SAMN43193237, SAMN43193240 & 3 \\
oh\_2016 & GSE70802 & SAMN03855826, SAMN03855832 & 2 \\
patel\_2020 & GSE140366 & SAMN13281630, SAMN13281631 & 2 \\
philippe\_2020 & GSE132703 & SAMN12050111, SAMN12050112 & 2 \\
pkm\_2022 & GSE202881 & SAMN28207083, SAMN28207084, SAMN28207085 & 3 \\
rao\_2020 & GSE158374 & SAMN16238708, SAMN16238709, SAMN16238710 & 3 \\
rao\_2025 & GSE297442 & SAMN48541656 & 1 \\
remes\_2022 & GSE156937 & SAMN15915088, SAMN15915089 & 2 \\
riepe\_2018 & GSE123539 & SAMN10564027, SAMN10564032 & 2 \\
rozman\_2025 & GSE309271 & SAMN51886188, SAMN51886189 & 2 \\
rozman\_2025\_spike & GSE309271 & SAMN51886174, SAMN51886175 & 2 \\
saito\_2024\_15 & GSE243312 & SAMN37408066, SAMN37408067 & 2 \\
sako\_2020 & GSE160917 & SAMN16676319, SAMN16676320 & 2 \\
santos\_2026 & GSE290865 & SAMN47176317 & 1 \\
sauer\_2019 & GSE105172 & SAMN07814395, SAMN07814396 & 2 \\
schneider\_poetsch\_2024 & GSE233886 & SAMN35556658, SAMN35556659 & 2 \\
sehrawat\_2022 & GSE166742 & SAMN17916585 & 1 \\
sharma\_2020\_dmso & GSE144140 & SAMN13909719, SAMN13909720, SAMN13909721 & 3 \\
sharma\_2021 & GSE136940 & SAMN12700070, SAMN12700071, SAMN12700072 & 3 \\
sharma\_2021\_100ug\_2minchx & GSE136940 & SAMN12700069 & 1 \\
sharma\_2021\_100ug\_5minchx & GSE136940 & SAMN12700057 & 1 \\
sharma\_2021\_200\_1minugchx & GSE136940 & SAMN12700056 & 1 \\
sharma\_2021\_200\_2minugchx & GSE136940 & SAMN12700055 & 1 \\
sharma\_2021\_300U & GSE136940 & SAMN12700054 & 1 \\
sharma\_2021\_500U & GSE136940 & SAMN12700052 & 1 \\
sharma\_2021\_700U & GSE136940 & SAMN12700051 & 1 \\
sharma\_2021\_900U & GSE136940 & SAMN12700044 & 1 \\
shichino\_2024 & GSE184247 & SAMN21450306, SAMN21450307 & 2 \\
sidrauski\_2015 & GSE65778 & SAMN03334842, SAMN03334848 & 2 \\
song\_2019 & GSE124558 & SAMN12391238, SAMN12391240 & 2 \\
song\_2019\_MNase & GSE124558 & SAMN12391242, SAMN12391244 & 2 \\
soto\_2021 & GSE173283 & SAMN18870217, SAMN18870218 & 2 \\
spijker\_2022\_rnase & GSE178242 & SAMN24923268 & 1 \\
spijker\_2022\_rnase\_at & GSE178242 & SAMN24923287 & 1 \\
suzuki\_2020 & GSE150439 & SAMN14907346, SAMN14907348 & 2 \\
tang\_2017 & GSE102786 & SAMN07515392 & 1 \\
tebaldi\_2018 & GSE112353 & SAMN08797688, SAMN08797689 & 2 \\
thalalla\_2025 & GSE272399 & SAMN42537790, SAMN42537791, SAMN42537792, SAMN42537793 & 4 \\
tomuro\_2024 & GSE233555 & SAMN35438994, SAMN35438995 & 2 \\
tomuro\_2024\_dmso & GSE233555 & SAMN35438970, SAMN35438971 & 2 \\
tomuro\_2024\_hs & GSE233555 & SAMN35438958, SAMN35438959 & 2 \\
vaninsberghe\_2021 & GSE162060 & SAMN16881237 & 1 \\
volegova\_2018 & GSE118239 & SAMN09779021 & 1 \\
wan\_2016 & GSE80156 & SAMN04632613, SAMN04632617 & 2 \\
weber\_2020 & GSE144841 & SAMN14049755, SAMN14049756 & 2 \\
weber\_2020\_ctr & GSE149279 & SAMN14689053, SAMN14689056 & 2 \\
weber\_2022 & GSE155854 & SAMN15756319, SAMN15756320 & 2 \\
weber\_2024 & GSE231964 & SAMN35004565, SAMN35004566 & 2 \\
wei\_2022 & GSE153142 & SAMN15357005 & 1 \\
wilczynska\_2019 & GSE134517 & SAMN12365723, SAMN12365724, SAMN12365725 & 3 \\
woo\_2018 & GSE103719 & SAMN07629904, SAMN07629905, SAMN07629906, SAMN07629907 & 4 \\
wu\_2020\_hek & GSE124737 & SAMN10700556, SAMN10700557 & 2 \\
wu\_2021 & GSE173852 & SAMN19014971, SAMN19014972 & 2 \\
yoon\_2025 & GSE305005 & SAMN50534830, SAMN50534831, SAMN50534832, SAMN50534833, SAMN50534834 & 5 \\
zhang\_2018\_hek & GSE94460 & SAMN06293621, SAMN06293633 & 2 \\
zhou\_2024 & GSE248572 & SAMN38412698, SAMN38412699 & 2 \\
zhu\_2024 & GSE268324 & SAMN41527436, SAMN41527437 & 2 \\
zhu\_2025\_chx & GSE255657 & SAMN39932460, SAMN39932462 & 2 \\
zou\_2022 & GSE197265 & SAMN26199030 & 1 \\
\end{longtable}
\endgroup

\begin{figure}[!htbp]
    \centering
    \includegraphics[width=\linewidth]{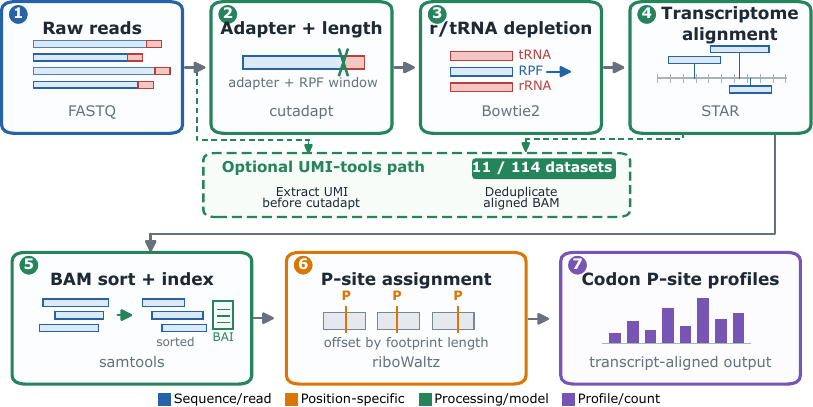}
    \caption{\textbf{Harmonized processing of HEK-derived Ribo-seq datasets.} Adapter trimming, read-length filtering, contaminant depletion, alignment, and P-site assignment produce codon-resolution profiles. Dataset-specific library settings are retained; 11 datasets follow the optional UMI extraction and deduplication path.}
    \label{fig:hek293_processing_pipeline}
\end{figure}

All datasets were processed with riboseq-flow \citep{Iosub2024riboseqflow}, the workflow can be seen in \cref{fig:hek293_processing_pipeline}, using GRCh38-based references and MANE Select transcripts from GENCODE v46 for transcript and CDS coordinates. \citep{Frankish2020gencode,morales2022mane}. Cutadapt \citep{Martin2011cutadapt} removed adapters and applied dataset-specific clipping and read-length filtering. Bowtie2 \citep{Langmead2012bowtie2} removed reads aligning to the specified human contaminant sequences and STAR \citep{Dobin2012star} generated genomic and transcriptome alignments. STAR used end-to-end alignment, allowed up to 20 mapping loci and capped the mismatch-to-read-length fraction at 0.08. FeatureCounts \citep{Liao2013feature}quantified reads overlapping annotated CDS regions, assigning fractional counts to multimapping reads. EFor 11 datasets, unique molecular identifiers (UMIs) were extracted before adapter trimming, and aligned reads were deduplicated with UMI-tools \citep{Smith2017umi}. We applied no additional duplicate removal to the remaining 103 datasets.

RiboWaltz \citep{Lauria2018waltz} inferred P-site offsets for each sample and read length from transcriptome alignments, after read-length and periodicity filtering. A 50\% dominant-frame threshold was used for 110 datasets; for four datasets with weaker periodicity, the threshold was relaxed to 30\% for one dataset and 10\% for three to retain read-length-specific P-site inference. Read lengths excluded by this filtering received no fallback P-site offset and contributed no reads to the resulting profiles. The workflow used riboseq-flow v1.1.1, Nextflow 25.10.4 \citep{DiTommaso2017nextflow}, and riboWaltz 1.2.0. Model-input filtering and reliability weighting are described in \cref{sec:real_preprocessing}.

\subsection{Quality control and ranking}
\label{app:hek293_quality}

\begin{figure}[!htbp]
    \centering
    \includegraphics[width=\linewidth]{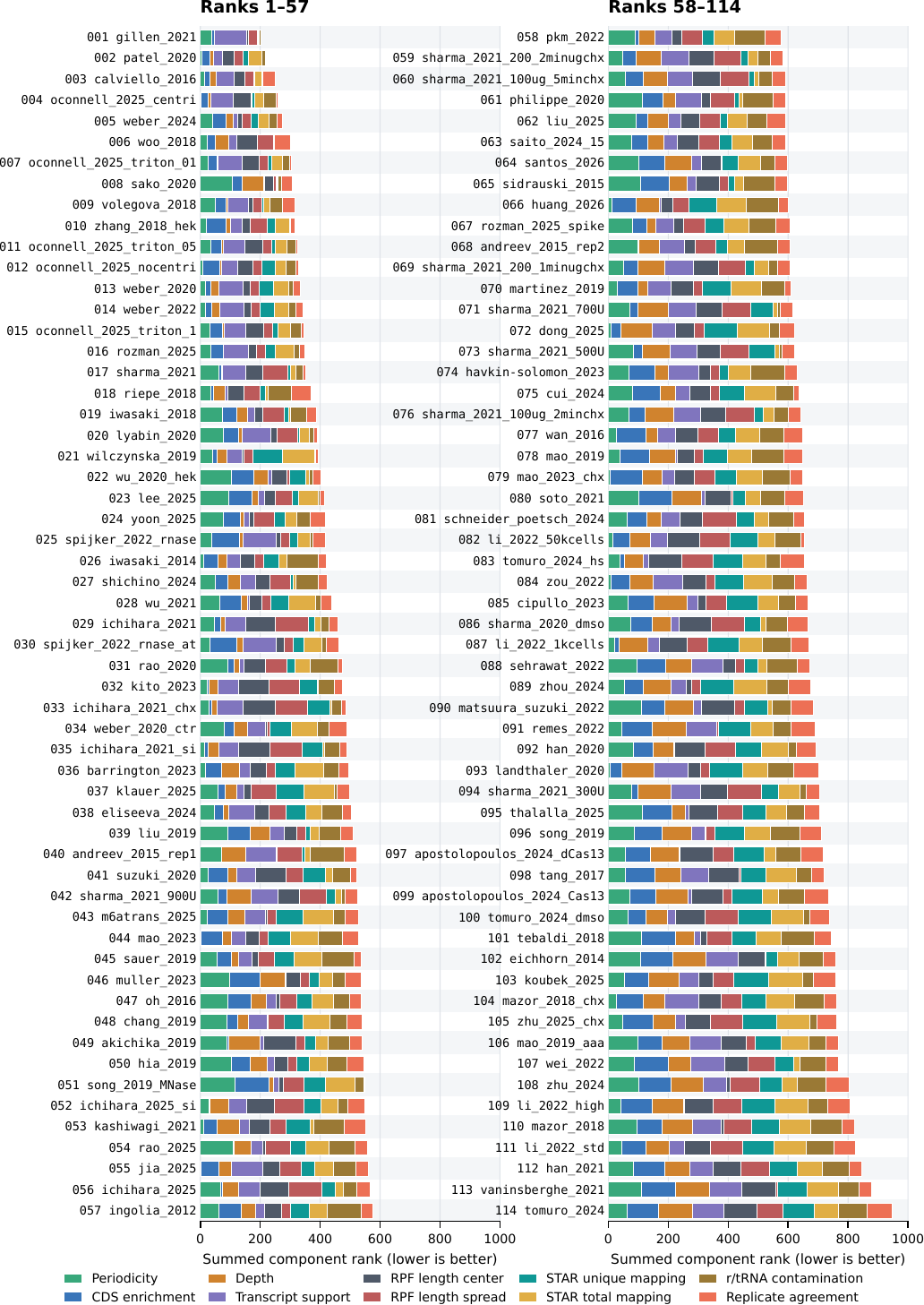}
    \caption{\textbf{QC ranking of all 114 HEK-derived datasets.} Each stacked bar shows the ten equally weighted component ranks that form the composite QC score. Shorter bars indicate lower composite QC scores.}
    \label{fig:hek293_quality_ranking_114}
\end{figure}

 \cref{tab:hek_qc_metrics} defines the ten QC components used to summarize periodicity, coding-region enrichment, sequencing support, footprint-length characteristics, mapping, contamination, and replicate agreement \citep{Iosub2024riboseqflow,Lauria2018waltz}. \cref{fig:hek293_quality_ranking_114} shows their contributions to the dataset ranking. For transcript support, we count transcripts with CDS-level Ribo-seq TPM greater than 1 in at least half of the sample profiles within a dataset.

\begin{table}[!htbp]
  \centering
  \caption{Ten equally weighted components of the composite QC score. Direction indicates whether higher or lower raw values yield
  a better component rank. Values are cohort medians, with first and third quartiles in brackets.}
  \label{tab:hek_qc_metrics}

  \begingroup
  \footnotesize
  \setlength{\tabcolsep}{3pt}
  \renewcommand{\arraystretch}{1.05}

  \begin{tabularx}{\linewidth}{@{}
    l
    >{\raggedright\arraybackslash}X
    c
    r
  @{}}
    \toprule
    Component & Implemented raw quantity & Direction
      & Median [$Q_1$--$Q_3$] \\
    \midrule
    Periodicity
      & Fraction of CDS P-sites in frame 0
      & Higher & 0.62 [0.56--0.72] \\

    CDS enrichment
      & Fraction of assigned P-sites falling in the CDS
      & Higher & 0.94 [0.93--0.95] \\

    Depth
      & Base-10 logarithm of the CDS P-site count plus one
      & Higher & 6.81 [6.25--7.23] \\

    Transcript support
      & Number of transcripts with CDS TPM $>1$
      & Higher & 11,896 [11,487--12,213] \\

    RPF-length center
      & Absolute deviation from 30 nt of the median
        sample-level mean RPF length (nt)
      & Lower & 1.78 [0.89--2.87] \\

    RPF-length spread
      & Median sample-level RPF-length interquartile range (nt)
      & Lower & 1.5 [1.0--3.0] \\

    STAR unique
      & Uniquely mapped STAR reads as a percentage of
        total STAR input reads (\%)
      & Higher & 45.51 [27.30--56.68] \\

    STAR total
      & Uniquely mapped plus multimapped STAR reads as a
        percentage of total STAR input reads (\%)
      & Higher & 84.09 [68.44--90.61] \\

    r/tRNA contamination
      & Recorded percentage of reads assigned to the local
        rRNA/tRNA contaminant category (\%)
      & Lower & 59.42 [42.61--72.18] \\

    Replicate agreement
      & Median transcript-level Pearson correlation of codon
        profiles across dataset replicate pairs
      & Higher & 0.19 [0.09--0.40] \\
    \bottomrule
  \end{tabularx}
  \endgroup
\end{table}

For each QC metric $m$, datasets are ranked from most to least favourable, giving ranks $r_{dm}$. The composite QC score of a dataset is $s_d=\sum_{m=1}^{10}r_{dm}$. Lower scores indicate more favourable measured data quality. The ten ranks have equal weight, and tied raw values receive average ranks. Missing replicate agreement is assigned the median rank among datasets with observed agreement. For each other metric $m$, missing values receive rank $n_m+1$, where $n_m$ is the number of datasets with an observed value for that metric. This ranking is a relative QC prioritization within the cohort, not validation of biological accuracy.

\section{Real-Data Evaluation across Independent Experimental Panels}
\label{app:real_four_panels}

\subsection{Profile eligibility and reliability weights}
\label{sec:real_preprocessing}

We retain finite, nonnegative profiles with positive total count.
Zero-count positions within retained profiles are kept. Raw replicates
enter the NB2 likelihood separately; their arithmetic mean supplies the
correlation targets and the reliability weights described below.

For transcript $t$ in dataset $d$, let $D_{t,d}$ be its mean count per
codon and $C_{t,d}$ the fraction of positions with positive counts,
both calculated from the replicate mean. Let $\tau_d$ be the median
$D_{t,d}$ over eligible training transcripts in dataset $d$. We calculate
\begin{equation}
    w_{t,d}^{\mathrm{raw}}
    =0.70\frac{\sqrt{D_{t,d}}}{\sqrt{D_{t,d}}+\sqrt{\tau_d}}
    +0.30C_{t,d},
    \qquad
    w_{t,d}
    =\frac{w_{t,d}^{\mathrm{raw}}}
    {\operatorname{median}_{u\in\mathcal T_d^{\mathrm{train}}}
    w_{u,d}^{\mathrm{raw}}},
    \label{eq:real_reliability}
\end{equation}
where $\mathcal T_d^{\mathrm{train}}$ contains the eligible training
transcripts. The first term increases with read support, with diminishing
returns, while the second rewards coverage across positions. These are
heuristic reliability weights. Both $\tau_d$ and the median normalization factor are calculated from training transcripts and held fixed when calculating validation and test weights.

\begin{figure}[!htbp]
    \centering
    \includegraphics[width=\linewidth]
        {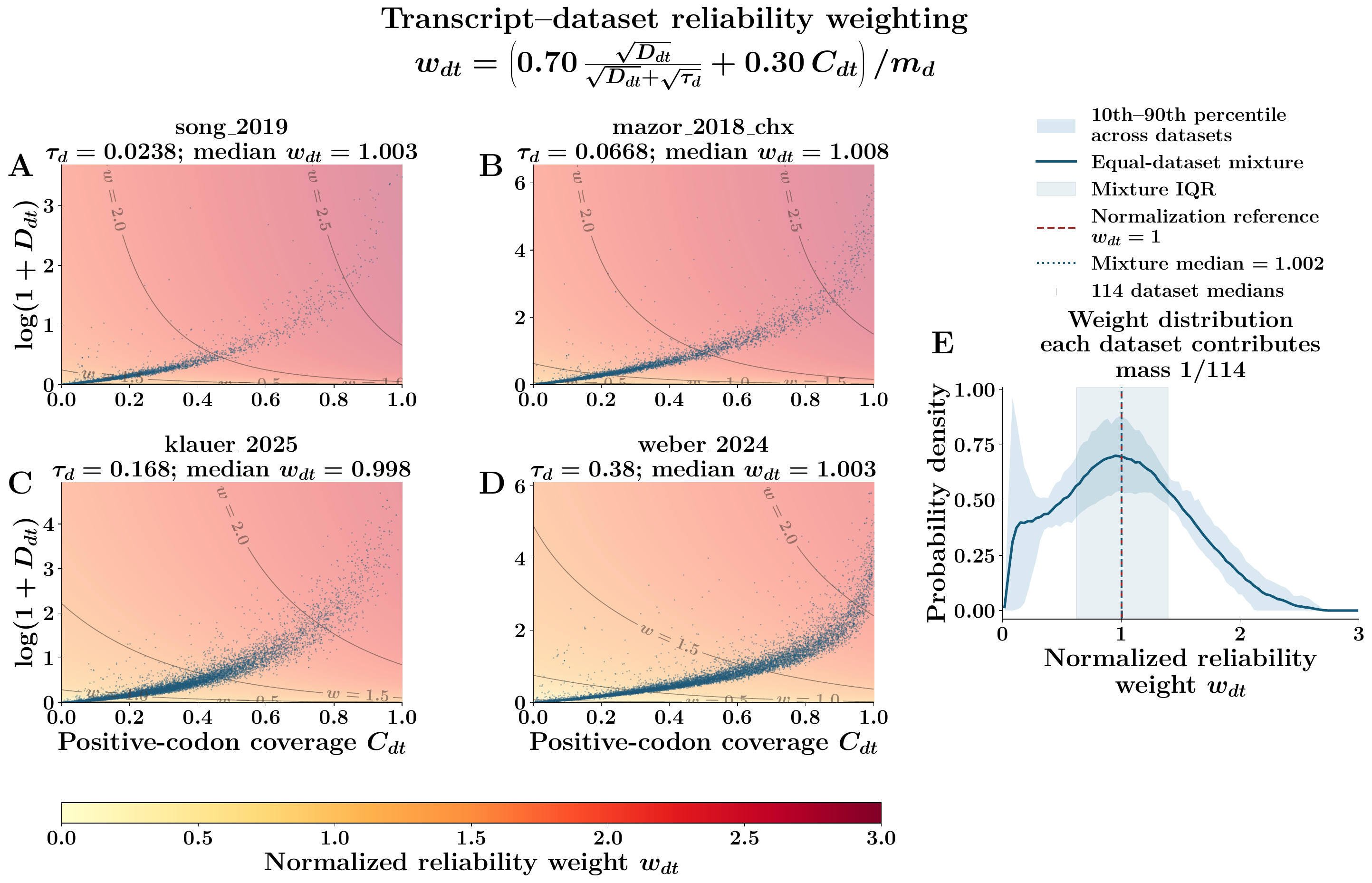}
    \caption{\textbf{Reliability weights across transcripts and datasets.}
    \textbf{(A--D)} Weights as a function of positive-codon coverage and
    log-transformed mean count for four example datasets, using
    normalization statistics calculated from training transcripts.
    \textbf{(E)} Distribution of weights across 114 datasets, with each
    dataset contributing equally. The envelope shows the pointwise
    10th--90th percentile range of dataset-specific densities; the pale
    band marks the interquartile range of the combined distribution.
    The vertical line marks a weight of one.}
    \label{fig:reliability_weight_example}
\end{figure}

Median normalization sets the median training weight within each dataset to one while preserving relative differences between transcripts
(Figure~\ref{fig:reliability_weight_example}). The reliability weights $w_{t,d}$ determine the relative contributions of observed transcript–dataset pairs to the training objective (\cref{app:transcript_reduction}), whereas the reference weights $\pi_d$ determine the centering of dataset-specific corrections

\subsection{HEK multi-dataset training settings}
\label{sec:real_training_settings}

Both sequence branches use two-layer bidirectional GRUs with 256 hidden
units per direction. The dataset-conditioned branch combines dataset
identity, codon, nucleotide, amino-acid, and positional features.
The correction and log-dispersion heads each have a 128-unit hidden
layer with dropout 0.1; predicted log-dispersion is restricted to
$[-5,1]$. All fits use the centering in \cref{eq:gamma_centering}, retain the product \(L_t\odot\gamma_{t,d}\) without final normalization in the decoder, and optimize the objective in Appendix
 \ref{app:training_details}.

Training uses AdamW with weight decay $10^{-2}$ and learning rates
$5\times10^{-4}$ for the shared encoder, $10^{-3}$ for the remaining
mean-model parameters, and $10^{-4}$ for the dispersion head.
Learning rates are reduced when validation loss stops improving.
All fits use random seed 42 and train for at most 200 epochs, with early stopping after 10 consecutive epochs without improvement in validation loss. Evaluation uses the checkpoint with
the lowest validation loss.

\subsection{Evaluation of shared-profile reproducibility}
\label{sec:real_stability_experiments}

Real data provide no independent ground truth for the shared profile.
We therefore evaluate whether models trained on different dataset collections produce similar shared profiles for the same held-out transcripts, and how this agreement changes with dataset selection and reference weighting. For models $A$ and $B$, we compare their mean-one predictions
at matching positions within each transcript:
\begin{equation}
    \operatorname{PCC}_t(A,B)
    =\operatorname{Corr}_i
    \left(\widehat L_{t,i}^{A},\widehat L_{t,i}^{B}\right).
    \label{eq:real_profile_agreement}
\end{equation}
Transcript identities, coding-sequence coordinates, and valid positions
are matched before comparison. This measures reproducibility between
models; biological accuracy requires independent validation.

\subsection{Panels with independent experimental sources}
\label{sec:real_four_panel_design}

We divide the 114 datasets into four panels, assigning all datasets from the same study group to the same panel. Shared source identifiers define these study groups;
when unavailable, datasets are grouped by author and year. Panel
assignment aims to balances median read density, positive-codon coverage,
number of eligible transcripts, and replicate Pearson correlation.
The resulting panels share neither datasets nor study groups
(Table~\ref{tab:real_four_panel_design}).

The evaluation pool contains 15,929 transcripts with usable observations
in at least two datasets from every panel. We select 1,593 validation
and 1,593 test transcripts and exclude both sets from every training
fold. The training populations may differ because observation
availability varies across panels. Each model uses uniform reference
weights $\pi_d=1/N$.

\begin{table}[!htbp]
    \centering
    \small
    \caption{\textbf{Four panels with independent experimental sources.}
    All models share the same 1,593 validation and 1,593 test transcripts.}
    \label{tab:real_four_panel_design}
    \begin{tabular}{@{}lrrr@{}}
        \toprule
        Panel & Datasets & Study groups & Training transcripts \\
        \midrule
        1 & 29 & 20 & 13,261 \\
        2 & 29 & 21 & 13,613 \\
        3 & 28 & 22 & 13,595 \\
        4 & 28 & 22 & 13,340 \\
        \bottomrule
    \end{tabular}
\end{table}

The shared profiles show strong agreement across all six panel pairs
(Figure~\ref{fig:real_data_stability_main}), supporting reproducibility
across experimental sources. These six comparisons reuse four fitted
models and are therefore not six independent replications.

\subsection{Effect of reference weighting on fixed panels}
\label{sec:real_four_panel_directionality}

We next train a separate model for each panel and reference-weighting policy while keeping dataset membership fixed. We use the composite QC score defined in Appendix \ref{app:hek293_quality}, which combines ten measures of sequencing quality, read support, and replicate agreement For each
measure, datasets are ranked from most to least favourable. The score is
\begin{equation}
    s_d=\sum_{m=1}^{10}r_{d,m},
    \label{eq:real_quality_score}
\end{equation}
where $r_{d,m}$ is dataset $d$'s rank for measure $m$.
A smaller score indicates more favourable measured data quality.
This score is not used to construct the four panels above.

The weighting experiment uses their same memberships with one common
split of 5,714 training, 714 validation, and 714 test transcripts.
For each panel $P$, we compare seven reference-weighting policies: uniform weights and weights favouring either better- or worse-scoring datasets at $p\in\{1,3,5\}$.
\begin{equation}
    \pi_{d,P}^{(p,\mathrm{best})}
    =\frac{s_d^{-p}}{\sum_{e\in P}s_e^{-p}},
    \qquad
    \pi_{d,P}^{(p,\mathrm{worst})}
    =\frac{s_d^{p}}{\sum_{e\in P}s_e^{p}},
    \qquad p\in\{1,3,5\}.
    \label{eq:real_four_panel_directional_weights}
\end{equation}
Larger $p$ concentrates the reference on fewer datasets. Across weighting policies within each panel, only the centering weights \(\pi_d\) change. Reliability weights \(w_{t,d}\), training settings, and transcript splits remain fixed.

For each test transcript, we calculate Pearson correlation and root mean
squared error (RMSE) between each of the six panel pairs, then average
across pairs. We report the mean over the 714 transcripts, with 95\%
intervals from 5,000 paired bootstrap resamples of whole transcripts.

\begin{figure}[!htbp]
    \centering
    \includegraphics[width=\linewidth]
        {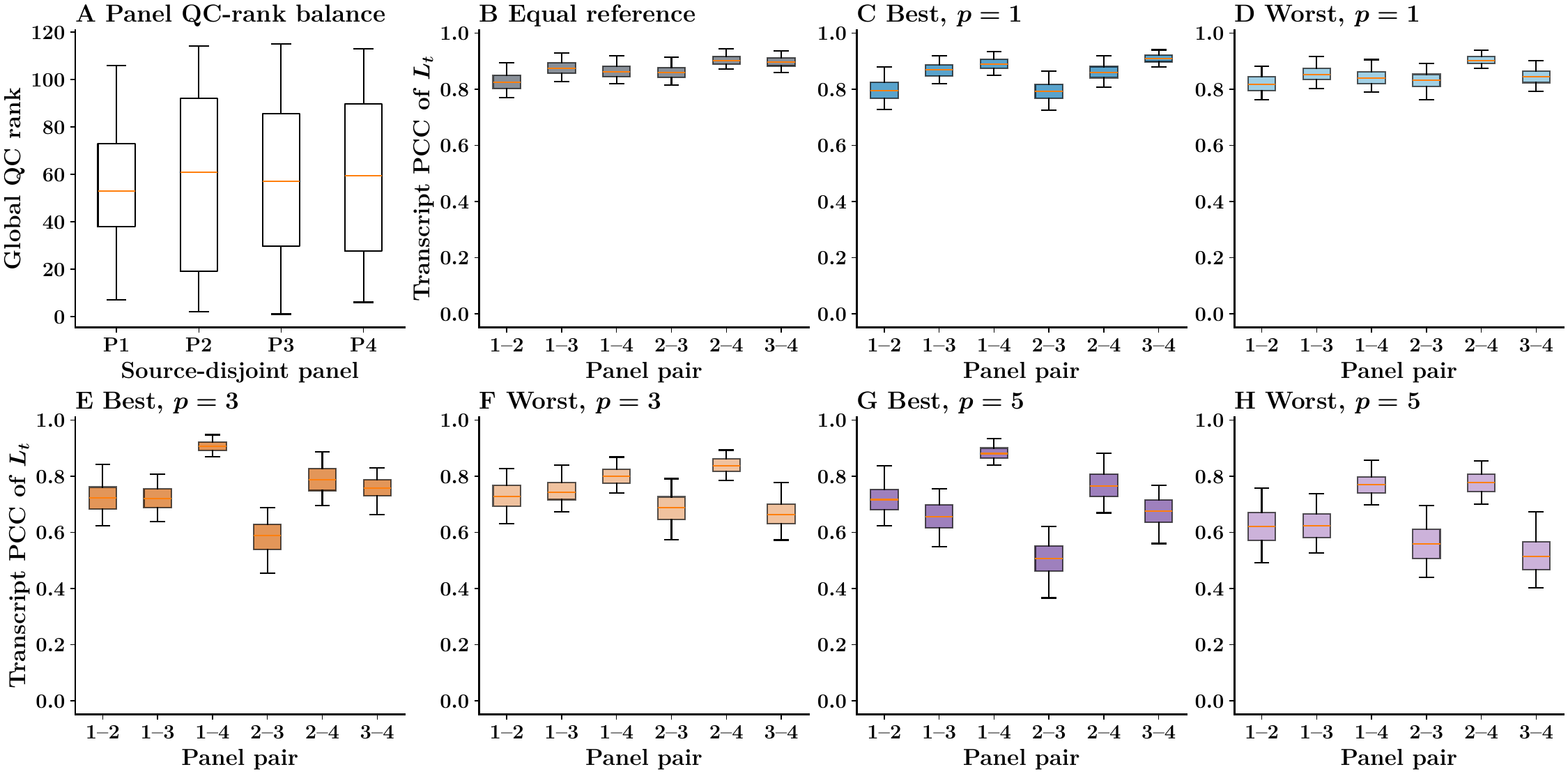}
    \caption{\textbf{Reference weighting and agreement across panels.}
    \textbf{(A)} Distribution of global QC ranks within each panel; lower ranks indicate more favourable measured quality.
    \textbf{(B--H)} Within-transcript Pearson correlations for the six
    panel pairs under uniform weights and weights favouring better- or
    worse-scoring datasets at $p=1,3,5$.
    The 28 fits comprise four panels evaluated under seven reference-weighting policies, all using the same 714 test transcripts.
    Boxes show medians and interquartile ranges; whiskers show the
    5th--95th percentiles.}
    \label{fig:real_four_panel_overview}
    \label{fig:real_four_panel_directionality}
\end{figure}

\begin{table}[!htbp]
    \centering
    \small
    \caption{\textbf{Agreement between shared profiles across panels under different reference-weighting policies.}
    Metrics are averaged over the six panel pairs within each transcript,
    then across transcripts. Brackets give 95\% percentile intervals from 5,000 paired transcript-bootstrap resamples.}
    \label{tab:real_four_panel_directionality}
    \begin{tabular}{@{}lcc@{}}
        \toprule
        Reference weights & Mean PCC ↑ & Mean RMSE ↓ \\
        \midrule
        Uniform & $0.8713\ [0.8695,0.8731]$ & $0.2649\ [0.2638,0.2660]$ \\
        Better-scoring, $p=1$ & $0.8533\ [0.8513,0.8553]$ & $0.3111\ [0.3093,0.3131]$ \\
        Worse-scoring, $p=1$ & $0.8492\ [0.8472,0.8512]$ & $0.2987\ [0.2973,0.3001]$ \\
        Better-scoring, $p=3$ & $0.7463\ [0.7435,0.7490]$ & $0.5622\ [0.5564,0.5682]$ \\
        Worse-scoring, $p=3$ & $0.7453\ [0.7424,0.7483]$ & $0.5295\ [0.5261,0.5328]$ \\
        Better-scoring, $p=5$ & $0.7007\ [0.6977,0.7036]$ & $0.6765\ [0.6701,0.6831]$ \\
        Worse-scoring, $p=5$ & $0.6461\ [0.6424,0.6501]$ & $0.7097\ [0.7051,0.7146]$ \\
        \bottomrule
    \end{tabular}
\end{table}

Uniform weighting gives the highest mean Pearson correlation and lowest mean RMSE across panels among the tested policies
(Table~\ref{tab:real_four_panel_directionality}). Increasing $p$ reduces mean Pearson correlation and increases mean RMSE for both weighting directions.
At $p=5$, favouring better-scoring datasets gives greater agreement
than favouring worse-scoring datasets, but neither improves on uniform
weighting. At $p=1$ and $p=3$, favouring better-scoring datasets gives slightly higher mean Pearson correlation but also higher mean RMSE. Moreover, equal exponents need not produce equal
concentration, so these comparisons do not isolate weighting direction
from concentration.

The worse-scoring $p=5$ models achieve a lower mean composite validation loss than the corresponding uniform-reference models, despite lower agreement across panels. Lower composite validation loss therefore does not necessarily imply greater reproducibility of the learned shared profile. The reference weights determine how positional structure is allocated between the shared profile and dataset-specific corrections; changing these weights therefore changes the reference used to define the shared profile.

\subsection{Stability as datasets are added}
\label{sec:real_cumulative_design}

Using the QC score in Equation~\ref{eq:real_quality_score}, we construct
nested dataset collections of sizes $N\in\{2,5,10,20,40,80,114\}$.
The best-first sequence adds datasets from lowest to highest score;
the worst-first sequence uses the reverse order. Each distinct dataset–weighting configuration is trained from scratch. We use the same 7,142-transcript cohort and 5,714/714/714 training, validation, and test split as in Appendix \ref{sec:real_four_panel_directionality}. This keeps the training, validation, and test transcript sets fixed across collection sizes, while restricting the analysis to transcripts observed in all 114 datasets.
Both the training observations and the centering reference change as datasets are added, so this experiment measures stability under their joint expansion.

\begin{figure}[!htbp]
    \centering
    \includegraphics[width=\linewidth]
        {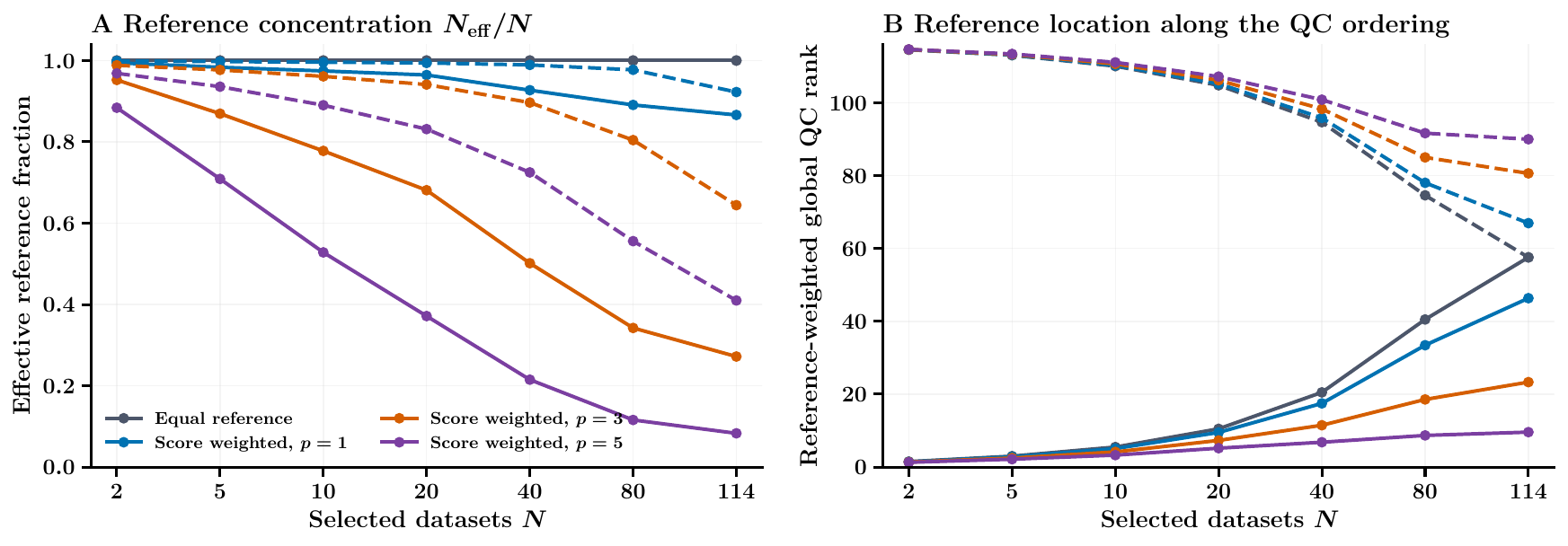}
    \caption{\textbf{Distribution of reference weights as datasets are added.}
    \textbf{(A)} Effective number of reference datasets divided by panel
    size, $N_{\mathrm{eff}}/N$.
    \textbf{(B)} Reference-weighted mean global QC rank, with lower ranks indicating more favourable measured quality.
    Solid lines show best-first collections; dashed lines show worst-first
    collections. These quantities describe the selected datasets and
    weights, rather than prediction performance.}
    \label{fig:real_cumulative_reference_geometry}
\end{figure}

\paragraph{Reference weights.}
\label{app:real_panel_reference_sensitivity}
For each collection \(\mathcal D_N\), we compare uniform reference weights with score-based weights defined by
\begin{equation}
    \pi_d^{(p,\mathrm{best})}
    =\frac{s_d^{-p}}{\sum_{e\in\mathcal D_N}s_e^{-p}},
    \qquad
    \pi_d^{(p,\mathrm{worst})}
    =\frac{s_d^{p}}{\sum_{e\in\mathcal D_N}s_e^{p}},
    \qquad p\in\{1,3,5\}.
    \label{eq:real_score_reference}
\end{equation}
Best-first collections favour better-scoring datasets, and worst-first
collections favour worse-scoring datasets. To describe concentration,
we report the effective number of reference datasets,
\begin{equation}
    N_{\mathrm{eff}}=\frac{1}{\sum_d\pi_d^2}.
    \label{eq:real_reference_diagnostics}
\end{equation}
Uniform weights give $N_{\mathrm{eff}}=N$; concentrating weight on
fewer datasets lowers this value. The same $p$ can give different
concentrations in the two selection directions because their score
distributions differ (Figure~\ref{fig:real_cumulative_reference_geometry}).

\begin{figure}[!htbp]
    \centering
    \includegraphics[width=\linewidth]
        {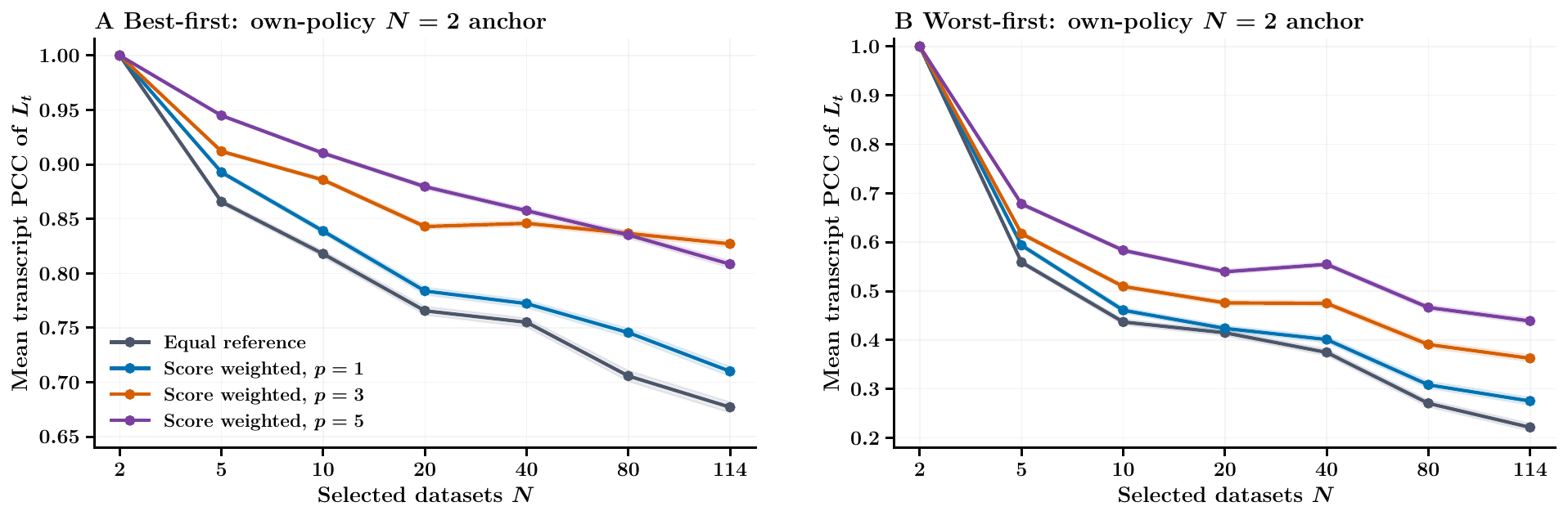}
    \caption{\textbf{Shared-profile stability as datasets are added.}
    \textbf{(A)} Best-first collections. \textbf{(B)} Worst-first collections.
    Each curve reports mean transcripts-level Pearson correlation with
    the corresponding $N=2$ model, using the same selection direction and reference
    weighting. Bands show 95\% transcript-bootstrap intervals over the
    same 714 test transcripts. Every curve starts at one because its first
    point compares a model with itself.}
    \label{fig:real_cumulative_directional}
\end{figure}

\paragraph{Stability relative to the initial fit.}
For each selection direction and weighting policy, we compare the shared profile at every collection size with the profile from that same policy’s $N=2$ fit (Figure~\ref{fig:real_cumulative_directional}). Compared with uniform references, concentrated references yield higher correlation with their respective initial fits along both paths, including the path starting from worse-scoring datasets. Agreement with the initial fit therefore measures stability of the representation, not biological accuracy.

\paragraph{Agreement between opposite selection directions.}
Direct agreement between best-first and worst-first models also increases
with collection size. These comparisons involve collections with no shared datasets through $N=40$, and no shared study groups through $N=20$.
At $N=40$, two study groups occur in both collections despite having
no dataset IDs in common. At $N=80$, the collections share 46 datasets;
at $N=114$, their membership is identical. The full-collection uniform
comparison therefore uses the same model, while differences between
score-weighted fits reflect their different reference weights.

\subsection{Interpretation and scope}

The independent panels support reproducibility across experimental
sources. In the fixed-panel comparison, uniform reference weights yield the highest mean agreement between shared profiles among the tested policies, showing that reproducibility depends on the reference definition.
All experiments use one training seed and fixed panel assignments;
bootstrap intervals describe variation across evaluated transcripts,
not across alternative training runs or panel assignments. 
Agreement across panels can still reflect measurement effects common to those panels or assumptions shared by the fitted models.
Independent biological measurements remain necessary to establish
how faithfully the shared profile reflects ribosome occupancy.

\section{Cross-organism benchmark panel}
\label{app:four_organism_benchmark}
We evaluate RiboUnmix using models trained separately on four organism-specific Ribo-seq benchmarks:
\textit{C. elegans} N2 and \textit{S. cerevisiae} BY4741 from
\citet{Stein2022wurmyeast}, \textit{E. coli} K-12 MG1655 from
\citet{Burkhardt2017ecoli}, and human HEK293T from
\citet{Iwasaki2016human}. All datasets represent control or wild-type conditions.
Dataset details are summarized in Table~\ref{tab:four_benchmark_datasets}.
The accessions are GSE152850 (SRR12055094, SRR12055095 for
\textit{C. elegans}; SRR12055102, SRR12055105 for yeast),
GSE77617 (SRR3147100 for \textit{E. coli}), and GSE70211
(SRR2075925, SRR2075926, SRR2075936, SRR2075937 for human).

\begin{table}[!htb]
\centering
\caption{Ribo-seq benchmark datasets across four organisms. Each dataset was reprocessed through the same core workflow using an organism-specific genome and CDS annotation.}
\label{tab:four_benchmark_datasets}
\small
\resizebox{\linewidth}{!}{%
\begin{tabular}{llllr}
\toprule
Dataset & Organism / system & Assembly / Annotation / CDS model &
RPF lengths (nt) & Profiles \\
\midrule
stein\_2021 &
\textit{C. elegans} N2 &
WBcel235; Ensembl 115, longest CDS/gene &
20--35 & 2 \\
zhang\_2016 &
\textit{E. coli} K-12 MG1655 &
ASM584v2; one CDS/locus tag &
20--40 & 1 \\
stein\_2021\_yeast &
\textit{S. cerevisiae} BY4741 &
R64-1-1; longest-CDS model &
20--35 & 2 \\
iwasaki\_2014 &
Human HEK293T &
GRCh38; GENCODE v46 MANE Select &
26--34 & 4 \\
\bottomrule
\end{tabular}%
}
\end{table}

\subsection{Shared processing of sequencing reads}
\label{app:four_benchmark_processing}

All datasets were processed using the riboseq-flow v1.1.1
\citep{Iosub2024riboseqflow} workflow shown in
Figure~\ref{fig:hek293_processing_pipeline}, with organism-specific references and read-length ranges. Reads are adapter-trimmed and length-filtered with cutadapt \citep{Martin2011cutadapt}, depleted of rRNA and tRNA reads with Bowtie2 \citep{Langmead2012bowtie2} and aligned to organism-specific transcript references with STAR \citep{Dobin2012star}. Alignments are sorted and indexed with samtools
\citep{Li2009samtools}, and coding-sequence reads are quantified with
featureCounts \citep{Liao2013feature}. STAR permits reads to map to at most 20 locations and featureCounts assigns fractional counts to multimapping reads. RiboWaltz estimates P-site offsets separately for each sample and read length, which are used to construct codon-resolution P-site profiles.
\citep{Lauria2018waltz}.

The references are WBcel235 with the Ensembl 115 longest coding sequence
per gene for \textit{C. elegans}; RefSeq GCF\_000005845.2 with one coding
sequence per locus and 50-nucleotide synthetic flanks for \textit{E. coli};
the SGD-derived R64-1-1 longest-coding-sequence model for yeast; and
GRCh38 with GENCODE v46 \citep{Frankish2020gencode} MANE Select
\citep{morales2022mane} transcripts for human. No terminal codons are
removed during this shared processing stage. Model-specific filtering,
normalization, and training masks are applied afterward, as described below.

\subsection{Common test population}
\label{app:four_model_test_set}

We compare RiboUnmix with iXnos \citep{tunney2018accurate}, RiboExp
\citep{hu2021riboexp}, sequence-only Riboformer \citep{shao2024riboformer}, RiboMIMO
\citep{tian2021full}, and Seq2Ribo \citep{kaynar2026seq2ribo}.
For each organism, eligible transcripts are partitioned into training, validation, and test sets using $8:1:1$ ratio
with seed 42. In native settings, model-specific eligibility filters determine which assigned training and validation transcripts are used. These filters do not restrict the assigned test population.

The assigned test populations contain 1,301 \textit{C. elegans}, 414
\textit{E. coli}, 1,499 human, and 506 yeast transcripts.
All settings are evaluated after removing five codons from each end
of the coding sequence. Thus, native filtering rules such as RiboExp's
selection of high-density transcripts or RiboMIMO's coverage threshold
do not restrict the test set.

\subsection{Model-specific preprocessing and training settings}
\label{app:training_set_checkpoint}

Different models use different rules for selecting transcripts,
normalizing counts, and including zero-count or terminal positions
in their training losses. We therefore report three settings for
each baseline:
\begin{itemize}
    \item \emph{Native}: the model's reported implementation of its
    published-style preprocessing and training procedure.
    \item \emph{Matched--U}: all eligible transcripts in the common training split, the matched baseline target preparation described below, and uniform transcript weights.
    \item \emph{Matched--W}: the same preparation as Matched--U, with transcript reliability weights from Equation~\ref{eq:real_reliability}.
\end{itemize}
Matched–U and Matched–W compare training with and without reliability weighting under a shared preprocessing scheme. Model-specific
objectives and checkpoint criteria still differ.

\paragraph{Native preprocessing and checkpoint selection.}
\begin{itemize}
    \item \textbf{iXnos.} Native iXnos requires at least 200 counts and 100 positive-count positions in the CDS window remaining after 20 codons are removed from each end. Zero targets remain in the squared-error
    objective. The selected checkpoint has the lowest validation mean
    squared error (MSE) over 30 epochs.

    \item \textbf{RiboExp.} The 500 transcripts with the highest mean footprint density are selected before intersection with the fixed split. Targets at or below $10^{-6}$ are completely omitted from the dataset (and thus excluded from the regression loss). Checkpoint selection maximizes validation Pearson correlation, with an early-stopping patience of 50 epochs.

    \item \textbf{Riboformer.} The native run uses an unweighted objective and its natively filtered training set, which restricts transcripts to coding sequences $>$200 nt and a mean footprint density in the top quartile of those length-eligible candidates. Checkpoint selection maximizes pooled, unweighted validation Pearson correlation.

    \item \textbf{RiboMIMO.} Training transcripts are retained when greater than 60\% of positions exceed 0.5 counts. Profiles are normalized
    by the mean count over those positions, and five codons at each end
    are masked during training. The coverage filter is not applied to
    the validation cohort. The objective is unweighted, and checkpoint
    selection maximizes validation Pearson correlation.

    \item \textbf{Seq2Ribo.} The native run uses the supplied publication-based training and validation filters, restricting the dataset to transcripts with a valid start codon, no internal stop codons and positive total reads. It then selects the checkpoint with the lowest validation loss.
\end{itemize}

\paragraph{Matched preprocessing.}
Matched baseline runs use all eligible transcripts in the parent training split and divide each observed profile by its mean over positive-count positions. No terminal mask is applied during fitting. For each baseline, Matched–U and Matched–W use identical validation and test transcript lists. 

RiboMIMO, RiboExp and Riboformer runs select checkpoints
by validation Pearson correlation. Seq2Ribo selects based on validation loss and iXnos based on validation MSE.
\cref{tab:four_model_training_provenance} reports cohort sizes and selected epochs.

For human Riboformer Matched–U, validation Pearson correlation was undefined at every epoch, and the implemented fallback selected epoch 1.
Its constant test predictions also give undefined Pearson and Spearman correlations. Undefined correlations are marked with dashes.

\paragraph{Models architecture and optimization.}
RiboMIMO uses 97 one-hot features per codon and a two-layer bidirectional GRU with 256 hidden units and no dropout. iXnos uses 760 one-hot features per codon and a feedforward MLP with a single hidden layer of 200 units and no dropout. Riboexp uses 90 one-hot features per codon (codon, nucleotide, and position) and an actor-critic policy network paired with a single-layer bidirectional GRU with 512 hidden units and 0.4 dropout. Riboformer uses 8-dimensional embeddings for 65 codon tokens and positions, processed by a five-layer CNN tower and a single Transformer block (10 heads) with 0.4 dropout in its dense head. Seq2ribo uses embeddings for 65 codon tokens, a scalar simulation feature, and three geometric features, processed by four Mamba layers with 192 hidden units and 0.1 dropout. 

\begin{table}[!htbp]
\centering
\begingroup
\scriptsize
\setlength{\tabcolsep}{4pt}
\renewcommand{\arraystretch}{1.02}
\begin{tabular}{@{}lccccl@{}}
\toprule
Architecture / setting
& \textit{C. elegans}
& \textit{E. coli}
& Human
& \textit{S. cerevisiae}
& Checkpoint criterion \\
\midrule
\textbf{iXnos}\enspace\textit{native}
& 3,710 (28) & 1,010 (24) & 2,296 (22) & 2,538 (27) & minimum validation MSE \\
\quad\(\hookrightarrow\)\enspace\textit{matched--U}
& 10,412 (30) & 3,307 (30) & 11,989 (14) & 4,043 (29) & minimum validation MSE \\
\quad\(\hookrightarrow\)\enspace\textit{matched--W}
& 10,412 (30) & 3,307 (16) & 11,989 (14) & 4,043 (29) & minimum validation MSE \\
\addlinespace[0.35em]
\textbf{RiboExp}\enspace\textit{native}
& 407 (170) & 409 (43) & 413 (68) & 418 (76) & maximum validation Pearson \\
\quad\(\hookrightarrow\)\enspace\textit{matched--U}
& 10,412 (7) & 3,307 (24) & 11,989 (25) & 4,043 (47) & maximum validation Pearson \\
\quad\(\hookrightarrow\)\enspace\textit{matched--W}
& 10,412 (18) & 3,307 (21) & 11,989 (29) & 4,043 (48) & maximum validation Pearson \\
\addlinespace[0.35em]
\textbf{Riboformer}\enspace\textit{native}
& 2,575 (6) & 785 (6) & 2,982 (11) & 995 (6) & maximum validation Pearson \\
\quad\(\hookrightarrow\)\enspace\textit{matched--U}
& 10,412 (2) & 3,307 (4) & 11,989 (1\textsuperscript{*}) & 4,043 (3) & maximum validation Pearson \\
\quad\(\hookrightarrow\)\enspace\textit{matched--W}
& 10,412 (3) & 3,307 (5) & 11,989 (11) & 4,043 (6) & maximum validation Pearson \\
\addlinespace[0.35em]
\textbf{Seq2Ribo}\enspace\textit{native}
& 10,408 (3) & 3,305 (1) & 11,964 (1) & 4,040 (5) & minimum validation loss \\
\quad\(\hookrightarrow\)\enspace\textit{matched--U}
& 10,412 (1) & 3,307 (1) & 11,989 (1) & 4,043 (1) & minimum validation loss \\
\quad\(\hookrightarrow\)\enspace\textit{matched--W}
& 10,412 (1) & 3,307 (1) & 11,989 (1) & 4,043 (1) & minimum validation loss \\
\addlinespace[0.35em]
\textbf{RiboMIMO}\enspace\textit{native}
& 2,644 (13) & 800 (17) & 875 (30) & 1,796 (21) & maximum validation Pearson \\
\quad\(\hookrightarrow\)\enspace\textit{matched--U}
& 10,412 (6) & 3,307 (8) & 11,989 (26) & 4,043 (10) & maximum validation Pearson \\
\quad\(\hookrightarrow\)\enspace\textit{matched--W}
& 10,412 (6) & 3,307 (13) & 11,989 (21) & 4,043 (10) & maximum validation Pearson \\
\addlinespace[0.35em]
\textbf{RiboUnmix}\enspace\textit{weighted}
& 10,412 (9) & 3,307 (10) & 11,989 (7) & 4,043 (12) & minimum validation loss \\
\bottomrule
\end{tabular}
\endgroup
\caption{\textbf{Training cohorts and validation-selected checkpoints by
architecture and setting.} Each organism cell gives the number of training
transcripts followed by the selected epoch in parentheses.  \emph{Native}
uses the model-specific preprocessing; \emph{matched--U} and
\emph{matched--W} use the full RiboUnmix cohort and and matched baseline target preparation
without or with the reliability weights from \cref{eq:real_reliability}. \textsuperscript{*}Human Riboformer matched--U had undefined validation Pearson at every epoch, so it fell back to epoch 1.}
\label{tab:four_model_training_provenance}
\end{table}

\subsection{Evaluation and results}
\label{app:4organism_results}

Pearson and Spearman correlations are calculated within each transcript using positive-count positions after excluding the first and last five codons, between the expected measured profile $\mu_{t,d}$ and the replicate-mean observation $\bar{Y}_{t,d}$. Normalized root mean squared error (RMSE)
uses the same positions. Before calculating nRMSE, each model’s target normalization is reversed using the corresponding transcript-specific scale factor. For each transcript, predicted and observed counts are divided by the mean observed count across the entire trimmed window, including zero-count positions.
Because predictions are not normalized by their own mean, RMSE retains amplitude errors relative to the supplied observed transcript scale.

Each result is the median transcript-level metric. Undefined correlations,
such as those involving constant predictions or targets, are omitted rather
than replaced by zero. For matched RiboMIMO and RiboUnmix, correlations
are defined for 1,114 of 1,301 \textit{C. elegans}, 337 of 414
\textit{E. coli}, 1,329 of 1,499 human, and 488 of 506 yeast transcripts.
Available-prediction and valid-correlation counts are reported for every setting; human Riboformer Matched–U has no defined correlations.
Consequently, a common intended test population does not guarantee
identical metric-valid subsets for every model.

For each organism and setting, we estimate uncertainty using 10,000 bootstrap resamples of transcripts, recomputing the median for each resample. Tables report each median plus or minus half the width of
its 95\% percentile interval. This compact notation does not preserve
any asymmetry of the original interval. The intervals describe transcript
sampling variation for a fixed trained model; they do not include
variation between training seeds or provide paired tests of model
differences.

RiboUnmix has the highest median Pearson point estimate in all four organisms and the highest median Spearman point estimate in three. In \textit{C. elegans}, native RiboMIMO has a slightly higher median Spearman estimate than RiboUnmix, 0.279 versus 0.277. The lowest normalized RMSE depends on the organism and setting
(Table~\ref{tab:four_model_nonzero_test_performance}).

Compared with Matched–U, Matched–W yields higher median Pearson point estimates for RiboMIMO and RiboExp in all four organisms. For Riboformer, Matched–W yields higher median Pearson estimates in the three organisms with defined correlations in both matched settings. In human, correlations are defined for Matched–W but undefined for Matched–U. Its effect on iXnos and Seq2Ribo depends on the organism.
Native-versus-matched comparisons jointly change transcript selection, normalization, masking, and, for some models, checkpoint selection, so differences cannot be attributed to a single preprocessing choice.

\begin{table}[!htbp]
\centering
\begingroup
\scriptsize
\setlength{\tabcolsep}{3.5pt}
\renewcommand{\arraystretch}{1.12}
\setlength{\aboverulesep}{0.3ex}
\setlength{\belowrulesep}{0.3ex}
\newcommand{\mci}[2]{#1\,{\(\pm\)}\,#2}
\newcommand{\bmci}[2]{\textbf{#1\,{\(\pm\)}\,#2}}
\newcommand{\bestmci}[2]{\underline{\textbf{#1\,{\(\pm\)}\,#2}}}
\newcommand{\naentry}{--}
\resizebox{\textwidth}{!}{%
\begin{tabular}{@{}lcccccc@{}}
\toprule
& \multicolumn{3}{c}{\textit{C. elegans}} & \multicolumn{3}{c}{\textit{E. coli}} \\
\cmidrule(lr){2-4} \cmidrule(lr){5-7}
Architecture / setting
& Pearson & Spearman & RMSE
& Pearson & Spearman & RMSE \\
\midrule
\textbf{iXnos}\enspace\textit{native}
& \mci{0.286}{0.020} & \mci{0.242}{0.020} & \bmci{1.944}{0.271}
& \mci{0.372}{0.030} & \mci{0.367}{0.027} & \mci{3.844}{0.698} \\
\quad\(\hookrightarrow\)\enspace\textit{matched--U}
& \mci{0.241}{0.019} & \mci{0.211}{0.015} & \mci{2.403}{0.367}
& \mci{0.344}{0.024} & \mci{0.350}{0.017} & \mci{4.062}{0.640} \\
\quad\(\hookrightarrow\)\enspace\textit{matched--W}
& \mci{0.271}{0.016} & \mci{0.237}{0.014} & \mci{2.048}{0.301}
& \mci{0.350}{0.028} & \mci{0.353}{0.022} & \mci{3.882}{0.617} \\
\addlinespace[0.3em]
\textbf{RiboExp}\enspace\textit{native}
& \mci{0.269}{0.022} & \mci{0.239}{0.017} & \mci{2.219}{0.371}
& \mci{0.364}{0.026} & \mci{0.359}{0.021} & \mci{5.176}{1.401} \\
\quad\(\hookrightarrow\)\enspace\textit{matched--U}
& \mci{0.204}{0.013} & \mci{0.165}{0.012} & \mci{2.364}{0.347}
& \mci{0.340}{0.024} & \mci{0.323}{0.022} & \mci{3.927}{0.588} \\
\quad\(\hookrightarrow\)\enspace\textit{matched--W}
& \mci{0.253}{0.018} & \mci{0.216}{0.016} & \bmci{1.964}{0.281}
& \mci{0.374}{0.024} & \mci{0.358}{0.027} & \bmci{3.730}{0.590} \\
\addlinespace[0.3em]
\textbf{Riboformer}\enspace\textit{native}
& \mci{0.297}{0.023} & \mci{0.253}{0.016} & \mci{2.000}{0.345}
& \mci{0.479}{0.026} & \mci{0.459}{0.030} & \mci{4.987}{1.222} \\
\quad\(\hookrightarrow\)\enspace\textit{matched--U}
& \mci{0.215}{0.015} & \mci{0.192}{0.011} & \mci{2.421}{0.384}
& \mci{0.461}{0.035} & \mci{0.450}{0.027} & \mci{4.090}{0.737} \\
\quad\(\hookrightarrow\)\enspace\textit{matched--W}
& \mci{0.257}{0.021} & \mci{0.226}{0.016} & \mci{2.154}{0.316}
& \mci{0.466}{0.032} & \mci{0.449}{0.037} & \mci{3.846}{0.753} \\
\addlinespace[0.3em]
\textbf{Seq2Ribo}\enspace\textit{native}
& \mci{0.155}{0.016} & \mci{0.167}{0.013} & \mci{3.135}{0.599}
& \mci{0.201}{0.024} & \mci{0.263}{0.018} & \mci{5.737}{1.334} \\
\quad\(\hookrightarrow\)\enspace\textit{matched--U}
& \mci{0.173}{0.019} & \mci{0.171}{0.014} & \mci{2.672}{0.443}
& \mci{0.224}{0.025} & \mci{0.275}{0.023} & \mci{4.756}{0.744} \\
\quad\(\hookrightarrow\)\enspace\textit{matched--W}
& \mci{0.201}{0.017} & \mci{0.190}{0.016} & \mci{2.545}{0.357}
& \mci{0.236}{0.028} & \mci{0.287}{0.020} & \mci{4.573}{0.754} \\
\addlinespace[0.3em]
\textbf{RiboMIMO}\enspace\textit{native}
& \bmci{0.328}{0.022} & \bestmci{0.279}{0.017} & \bmci{1.958}{0.287}
& \mci{0.519}{0.034} & \mci{0.482}{0.035} & \mci{5.631}{1.448} \\
\quad\(\hookrightarrow\)\enspace\textit{matched--U}
& \mci{0.296}{0.019} & \mci{0.257}{0.016} & \bestmci{1.761}{0.218}
& \mci{0.490}{0.034} & \mci{0.465}{0.027} & \bestmci{3.325}{0.473} \\
\quad\(\hookrightarrow\)\enspace\textit{matched--W}
& \bmci{0.325}{0.024} & \bmci{0.269}{0.019} & \bmci{1.764}{0.240}
& \mci{0.516}{0.028} & \mci{0.487}{0.037} & \bmci{3.668}{0.677} \\
\addlinespace[0.3em]
\textbf{RiboUnmix}\enspace\textit{weighted}
& \bestmci{0.335}{0.026} & \bmci{0.277}{0.019} & \mci{2.394}{0.502}
& \bestmci{0.556}{0.027} & \bestmci{0.521}{0.030} & \mci{3.908}{0.745} \\
\midrule
& \multicolumn{3}{c}{Human (HEK293T)} & \multicolumn{3}{c}{\textit{S. cerevisiae}} \\
\cmidrule(lr){2-4} \cmidrule(lr){5-7}
Architecture / setting
& Pearson & Spearman & RMSE
& Pearson & Spearman & RMSE \\
\midrule
\textbf{iXnos}\enspace\textit{native}
& \mci{0.468}{0.019} & \mci{0.372}{0.015} & \mci{3.872}{0.294}
& \mci{0.442}{0.018} & \mci{0.367}{0.015} & \bmci{1.102}{0.062} \\
\quad\(\hookrightarrow\)\enspace\textit{matched--U}
& \mci{0.477}{0.018} & \mci{0.378}{0.011} & \mci{4.161}{0.270}
& \mci{0.431}{0.019} & \mci{0.361}{0.017} & \mci{1.164}{0.061} \\
\quad\(\hookrightarrow\)\enspace\textit{matched--W}
& \mci{0.472}{0.014} & \mci{0.375}{0.012} & \mci{3.824}{0.257}
& \mci{0.444}{0.019} & \mci{0.365}{0.019} & \bmci{1.107}{0.057} \\
\addlinespace[0.3em]
\textbf{RiboExp}\enspace\textit{native}
& \mci{0.407}{0.018} & \mci{0.352}{0.012} & \mci{5.214}{0.549}
& \mci{0.405}{0.022} & \mci{0.341}{0.019} & \bmci{1.109}{0.059} \\
\quad\(\hookrightarrow\)\enspace\textit{matched--U}
& \mci{0.339}{0.014} & \mci{0.316}{0.009} & \mci{4.400}{0.257}
& \mci{0.435}{0.014} & \mci{0.339}{0.015} & \mci{1.143}{0.055} \\
\quad\(\hookrightarrow\)\enspace\textit{matched--W}
& \mci{0.418}{0.013} & \mci{0.323}{0.009} & \mci{3.651}{0.188}
& \mci{0.456}{0.020} & \mci{0.368}{0.025} & \bmci{1.053}{0.061} \\
\addlinespace[0.3em]
\textbf{Riboformer}\enspace\textit{native}
& \mci{0.458}{0.018} & \mci{0.371}{0.013} & \mci{4.193}{0.371}
& \mci{0.417}{0.019} & \mci{0.354}{0.023} & \mci{1.134}{0.075} \\
\quad\(\hookrightarrow\)\enspace\textit{matched--U}
& \naentry & \naentry & \mci{5.218}{0.363}
& \mci{0.415}{0.017} & \mci{0.350}{0.023} & \mci{1.192}{0.069} \\
\quad\(\hookrightarrow\)\enspace\textit{matched--W}
& \mci{0.465}{0.017} & \mci{0.370}{0.012} & \mci{3.768}{0.212}
& \mci{0.435}{0.016} & \mci{0.362}{0.022} & \mci{1.125}{0.067} \\
\addlinespace[0.3em]
\textbf{Seq2Ribo}\enspace\textit{native}
& \mci{0.194}{0.010} & \mci{0.222}{0.009} & \mci{4.465}{0.317}
& \mci{0.245}{0.014} & \mci{0.242}{0.014} & \mci{1.697}{0.152} \\
\quad\(\hookrightarrow\)\enspace\textit{matched--U}
& \mci{0.223}{0.012} & \mci{0.245}{0.009} & \mci{4.268}{0.287}
& \mci{0.350}{0.018} & \mci{0.314}{0.012} & \mci{1.203}{0.074} \\
\quad\(\hookrightarrow\)\enspace\textit{matched--W}
& \mci{0.222}{0.012} & \mci{0.243}{0.009} & \mci{4.280}{0.338}
& \mci{0.347}{0.014} & \mci{0.311}{0.017} & \mci{1.188}{0.083} \\
\addlinespace[0.3em]
\textbf{RiboMIMO}\enspace\textit{native}
& \mci{0.440}{0.017} & \mci{0.372}{0.013} & \mci{4.692}{0.411}
& \mci{0.492}{0.016} & \bmci{0.403}{0.018} & \bmci{1.061}{0.073} \\
\quad\(\hookrightarrow\)\enspace\textit{matched--U}
& \mci{0.440}{0.017} & \mci{0.345}{0.012} & \bmci{3.322}{0.220}
& \mci{0.489}{0.022} & \bmci{0.401}{0.019} & \bestmci{1.047}{0.069} \\
\quad\(\hookrightarrow\)\enspace\textit{matched--W}
& \mci{0.461}{0.016} & \mci{0.368}{0.011} & \bestmci{3.316}{0.235}
& \bmci{0.502}{0.019} & \bmci{0.405}{0.018} & \bmci{1.053}{0.068} \\
\addlinespace[0.3em]
\textbf{RiboUnmix}\enspace\textit{weighted}
& \bestmci{0.539}{0.019} & \bestmci{0.412}{0.012} & \mci{3.830}{0.282}
& \bestmci{0.517}{0.024} & \bestmci{0.415}{0.019} & \bmci{1.072}{0.078} \\
\bottomrule
\end{tabular}%
}
\endgroup

\caption{\textbf{Benchmark performance across four organisms and training settings.} Entries report median transcript-level metrics $\pm$ half the width of their 95\% percentile bootstrap with 10,000 transcript resamples. All metrics compare predicted and observed profiles at positions with positive observed counts after excluding the first and last five codons of each CDS. For normalized RMSE,
target and prediction are first divided by the target mean over the complete
trimmed window, including zero-count positions For the baselines, native denotes model-specific preprocessing and training, whereas matched–U and matched–W use a common training cohort and target preparation with uniform and reliability-based transcript weights, respectively. Dashes indicate undefined correlations caused by constant test predictions for human Riboformer matched–U.  The best point estimate in each
organism--metric column is bold and underlined (maximum for correlations;
minimum for RMSE).  Bold without underlining marks another point estimate that
falls inside the best entry's reported
\(\widehat m_{\mathrm{best}}\pm\Delta_{95,\mathrm{best}}\) span.  This visual
rule is descriptive and is not a paired significance test.  See
\cref{app:4organism_results} for details.}
\label{tab:four_model_nonzero_test_performance}
\end{table}

The benchmark compares complete prediction pipelines under both native and matched preprocessing. Matching training cohorts reduces one source of variation, while model-specific losses, checkpoint-selection criteria, and the use of a single training seed limit attribution of the remaining differences to architecture alone.

\end{document}